\documentclass[10pt]{article}
\usepackage[latin1]{inputenc}
\usepackage{amsmath}
\usepackage{amsfonts}
\usepackage{amssymb}
\usepackage{mathrsfs}
\usepackage{verbatim}
\usepackage{fancyhdr}
\usepackage{blindtext}
\usepackage{MnSymbol}

\usepackage[cal=boondox,scr=boondoxo]{mathalfa}

\usepackage{indentfirst}

\usepackage{float}
\usepackage{subfig}
\usepackage{fancybox,graphicx}
\usepackage{subfig}
\usepackage{caption}

\usepackage{color}
\usepackage{authblk}

\usepackage{algorithm}
\usepackage{algorithm}  
\usepackage{algpseudocode} 

\usepackage{amsthm}
\usepackage[colorlinks,linkcolor=blue,citecolor=blue]{hyperref}

\usepackage{hyperref}

\newcommand{\doi}[1]{{doi:~\href{https://doi.org/#1}{\nolinkurl{#1}}}\rmFullStop}

\newcommand*{\rmFullStop}{\rmifnextchar{.}{}{}}

\makeatletter
\newcommand{\rmifnextchar}[3]{%
  \begingroup
  \ltx@LocToksA{\endgroup#2}%
  \ltx@LocToksB{\endgroup#3}%
  \ltx@ifnextchar{#1}{%
    \def\next{\the\ltx@LocToksA}%
    \afterassignment\next
    \let\scratch= %
  }{%
    \the\ltx@LocToksB
  }%
}
\makeatother %doi command

\usepackage{accents}
\usepackage[titletoc,title]{appendix}

\usepackage{natbib} %[sort&compress]

\usepackage{multirow}
\usepackage{booktabs}

\usepackage{lipsum}
\usepackage{adjustbox}

\usepackage{mathtools}

\newtheoremstyle{normalfont}
  {}{}
  {\normalfont}
  {0pt}
  {\bfseries}
  {.}
  {.5em}
  {}
\theoremstyle{normalfont}
\newtheorem{assumption}{Assumption}
\newtheorem{lemma}{Lemma}

\usepackage[top=1in, bottom=1in, left=1in, right=1in]{geometry}

\usepackage{stackengine}

\title{Non-asymptotic convergence bound of conditional diffusion models}
\author[1]{Mengze Li}
\affil[1]{School of Mathematics and Statistics, Northeast Normal University, China}%limz770@nenu.edu.cn
\date{}

\begin{document}

\maketitle

\begin{abstract}
\noindent 
Learning and generating various types of data based on conditional diffusion models has been a research hotspot in recent years. Although conditional diffusion models have made considerable progress in improving acceleration algorithms and enhancing generation quality, the lack of non-asymptotic properties has hindered theoretical research.
To address this gap, we focus on a conditional diffusion model within the domains of classification and regression (CARD), which aims to learn the original distribution with given input $x$ (denoted as $Y|X$). It innovatively integrates a pre-trained model $f_{\phi}(x)$ into the original diffusion model framework, allowing it to precisely capture the original conditional distribution given $f$ (expressed as $Y|f_{\phi}(x)$). Remarkably, when $f_{\phi}(x)$ performs satisfactorily, $Y|f_{\phi}(x)$ closely approximates $Y|X$.
Theoretically, we deduce the stochastic differential equations of CARD and establish its generalized form predicated on the Fokker-Planck equation, thereby erecting a firm theoretical foundation for analysis. Mainly under the Lipschitz assumptions, we utilize the second-order Wasserstein distance to demonstrate the upper error bound between the original and the generated conditional distributions. Additionally, by appending assumptions such as light-tailedness to the original distribution, we derive the convergence upper bound between the true value analogous to the score function and the corresponding network-estimated value.
% Meanwhile, in order to verify the theoretical results more quickly, we propose a novel sampler, named the Implicit Classification and Regression Diffusion Model (ICARD), to surmount the efficiency bottleneck in sample generation, augmenting the model's efficiency without compromising on quality. 
Numerical simulations conducted with diverse datasets validate the accuracy of the non-asymptotic convergence upper bound, corroborating the reliability and practicality of the theoretical findings.

\begin{comment}
 \\\\Keywords: Conditional diffusion models, Stochastic differential equations, Wasserstein distance, Taylor polynomials   
\end{comment}
\end{abstract}

\tableofcontents

\section{Introduction}
Diffusion models \citep{sohl2015deep,ho2020denoising} have received extensive attention due to their ability to generate high-quality and diverse samples and have become the foundation of contemporary generative modeling, achieving remarkable success in different fields. Among them, the Score-based Generative Models (SGM) \citep{song2019generative} have demonstrated superior performance in various fields such as computer vision, especially in image generation \citep{dhariwal2021diffusion, rombach2022high, zhang2023adding, zhu2023conditional, chen2023seeing}, natural language processing \citep{austin2021structured, li2022diffusion, li2023diffusion,wu2023ar}, medicine, and biology \citep{liuegdiff, yuan2019deep, he2024ai, zhou2024conditional}. A series of verifications confirm that its performance surpasses earlier models like GANs \citep{goodfellow2020generative} and VAEs \citep{kingma2013auto} in multiple aspects. 

The operation mechanism of diffusion models encompasses two key stages: diffusion and generation process. In the diffusion stage, samples evolve following specific stochastic dynamics, gradually shifting from the target data distribution to a pure noise distribution. The generation stage starts from the pure noise distribution and, based on the pre-learned reverse diffusion process, progressively denoises and reconstructs samples to approximate the original data distribution, fulfilling the core aim of generative modeling.
During this process, the application of score matching methods has become increasingly prominent. SGMs \citep{song2019generative, song2021maximum} skillfully introduce the logarithmic probability density gradient and score matching techniques. This innovation substantially improves the consistency between the generated and actual data distributions. By precisely estimating and utilizing the gradient of the logarithmic probability density function, the model can more sensitively capture the fine-grained features and changing trends of the data distribution, enabling more accurate simulation of real data characteristics during sample generation. This method not only enhances the diversity and authenticity of generated samples but also equips the model with greater adaptability and stability when dealing with complex data distributions, providing more reliable and efficient technical means for the widespread application of diffusion models in various fields and significantly promoting the development and innovation of generative modeling techniques. 

In the evolution of diffusion model research, the primitive model exhibits sluggish iteration due to its stepwise sampling mechanism and the necessity of evaluating the neural network of the score function per iteration, thereby encountering an efficiency constraint in practical applications. In light of this, numerous scholars have delved deeply into accelerating the diffusion model without sacrificing its output quality, yielding a series of pivotal achievements.
Denoising diffusion implicit model (DDIM) \citep{song2020denoising} extends DDPM by combining non-Markov processes with Markov chains, which brings a revolutionary change to the field of denoising, with its sampling methodology akin to a specialized discretization of probabilistic flow ordinary differential equations(ODEs). Generalized DDIM \citep{zhang2022gddim} further expands its application range, enabling deterministic sampling. \cite{nichol2021improved} enhance sampling efficiency while safeguarding sample quality through noise setting optimization.
Regarding sample generation, to precisely govern the generation process, a flexible input "guidance" mechanism has been introduced. \cite{dhariwal2021diffusion} combine the diffusion model with GAN to propose the classifier guidance method. \cite*{ho2022classifier} introduce classifier-free guidance, validating the efficacy of jointly training conditional and unconditional diffusion models, with extensive applications \citep{meng2023distillation, kornblith2023classifier}. \cite{han2022card} augment the generation process by incorporating the conditional mean as a covariate. \cite{chung2022come} introduce a random contraction operation during diffusion, hastening the convergence of samples in the potential space towards the target distribution. Additionally, methods such as DPM-Solver \citep{lu2022dpm1}, DPM-Solver++ \citep{lu2022dpm2}, DPM-Solver-v3 \citep{zheng2023dpm}, and Unipc \citep{zhao2024unipc} leverage pre-trained score functions and modify the sampling process to expedite training. 

In recent years, the convergence theory of diffusion models has emerged as a prominent research focus. However, the current research landscape exhibits an imbalance, with the majority of findings concentrated on unconditional diffusion models \citep{de2021diffusion, kwon2022score, de2022convergence, lee2022convergence, chen2022sampling, zhou2023deep, lee2023convergence, chen2023score, oko2023diffusion, benton2023linear, wibisono2024optimal}, which have investigated the convergence properties of unconditional diffusion models from diverse perspectives, thereby establishing a crucial foundation for the fundamental theory in this domain.
Conversely, while the theoretical research on conditional diffusion models has achieved certain milestones, it frequently hinges on relatively strict prerequisites. For instance, several studies \cite{zhou2023deep, yuan2024reward, baldassari2024conditional, chen2023restoration, gao2024convergence, chen2024convergence, li2023towards, li2024towards, li2024accelerating, li2024sharp, jiao2024model, tang2024conditional, liang2024theory}
necessitate the imposition of additional assumptions regarding the target data distribution or the furnishing of specific assurances for score function estimation. This, to some extent, restricts the scope and generality of their application.
Nevertheless, if precise score estimates are attainable, they can offer robust guarantees for the sampling process without relying on structural assumptions like the logarithmic concavity or smoothness of the target distribution. This has been aptly demonstrated in \citep{chen2022sampling, chen2023improved, benton2023linear, li2024d}. Such findings present novel avenues and potentialities for the application of diffusion models in more intricate and heterogeneous data scenarios. 

In this paper, our focus is on the non-asymptotic properties of the Classification and Regression Diffusion Model (CARD) \citep{han2022card}, which salient characteristic is the integration of pre-trained function \(f_{\phi}(x)\) about input $x$ (where \(\phi\) represents the parameter of network estimation) into the original diffusion model framework \citep{ho2020denoising}, applying to complex real-world situations and augmenting the model's efficacy and adaptability in various tasks.
Futhermore, the pre-trained function \(f_{\phi}(x)\) denoted as the conditional mean function.  From the vantage point of theoretical generality, the function \(f_{\phi}(x)\) exhibits a flexible construction. It can be rendered as an identity mapping to suit specific data distributions and task features, and further extended to a non-identity mapping to address diverse task demands and expand the model's task adaptation range.

We first propose the theoretical framework of the conditional diffusion model with conditional mean function $f_{\phi}(x)$. By leveraging its specific stochastic differential equation as the starting point, we rigorously derive the second-order Wasserstein distance upper bound for the conditional distribution estimation, under the prerequisite that the sample smoothness complies with Lipschitz and other requisite assumptions.
Regarding the logarithmic gradient term and its approximation (analogously defined as the score function estimate), we innovatively establish a non-asymptotic convergence theory regime related to the conditional score function. This is achieved through the application of the Taylor approximation technique of conditional diffusion, under the conditions of controlled data range and light-tailed original conditional distribution. The successful establishment of this theoretical system effectively addresses the existing gaps and deficiencies in this critical theoretical aspect of the current conditional diffusion model. It elucidates the convergence characteristics of the conditional diffusion model within diverse task scenarios, lays a solid theoretical foundation for the optimization and expansion of the model's practical applications, and provides a guiding direction. 

% \noindent \textbf{Main contributions}
\subsection{Main contributions}
Technically, the primary challenge lies in a conditional diffusion model with pre-trained condition, deriving the derivation of error bounds for non-asymptotic convergence of both the conditional distribution and the fractional function respectively. Our main contributions are as follows.\\

\noindent\textbf{Stochastic Differential Equation Form:} 
We derive stochastic differential equation formulations for CARD, and analyze its general expression using the Fokker-Planck equation, providing a robust framework for theoretical investigations.\\

\noindent\textbf{Convergence of the original and generated distribution:}
We pioneer the establishment of the first non-asymptotic convergence bound within the context of CARD, specifically regarding the second-order Wasserstein distance between the original conditional distribution and the generated distribution and giving an theory upper bound for this convergence within the optimal transport theory framework. Our findings are applicable to the generalized form of the score function involving two covariates, significantly expanding the theoretical underpinnings for the development of conditional diffusion models in multivariate task scenarios.\\ %which we validate through experimental results.

\noindent\textbf{Convergence of the score function and network estimator:}
For the purpose of conducting a precise dissection of the extant error bounds and probing deeply into the error between the score function and the network estimate, we anchor our methodology on the presupposition that the original data exhibits H$\ddot{o}$lder continuity and impose constraints on both the value range and the lower limit of the conditional distribution. Within the ambit of the $L_{2}$ norm, by adroitly employing the Taylor approximation technique in the context of conditional diffusion, we meticulously deduce the upper bound of error convergence between the score function and its corresponding estimate. This resultant outcome lucidly delineates the specific manifestation of the upper bound, which is correlated with particular hyperparameters. This forms a pivotal facet of our CARD non-asymptotic theory, furnishing crucial theoretical scaffolding and a precise quantitative analytic foundation for pertinent research endeavours.

\subsection{Related Work}
\noindent \textbf{Convergence theory for diffusion models.}
%\cite{block2020generative, de2021diffusion, de2022convergence, kwon2022score, pidstrigach2022score, liu2022let}
The convergence of early diffusion models mainly centered on unconstrained models. For instance, \citep{block2020generative} estimates the fractional function of the distribution using a denoised autoencoder or denoised fraction matching, and employs Langevin diffusion for sampling. A set of hypotheses are made for the target distribution, such as the Lipschitz property of the score function and the log-Sobolev inequality. The upper bound of the Wasserstein distance between the distribution of the discrete Langevin sampling scheme and the target distribution is derived, and the convergence within finite samples is proved.
\citep{liu2022let} regards diffusion models as potential variable models and applies maximum likelihood estimation to understand and expand diffusion generation models.
\cite{de2022convergence} and \cite{kwon2022score} have contributed to our understanding of DDPM convergence, providing error bounds and demonstrating the minimization of Wasserstein distances between original and generated data under specific conditions.
Most previous works focused on the cases of exact or random gradients. However, in SGM, the score function is only accurate in \(L^{2}\) estimates, which poses a challenge for analysis. The analysis of Langevin Monte Carlo has some issues, such as the curse of dimensionality and the exponential increase of error with time. Nevertheless, \citep{lee2022convergence} proves the first polynomial convergence guarantee for the core mechanism of SGM, although assuming the log-Sobelev inequality on the target distribution and the smoothness of the score estimation. \citep{lee2023convergence} subsequently relaxed the hypothesis by requiring the bounded range of the original data and the estimated error bounds of the score function to improve the convergence result.

Recently there has been extensive exploration of the theoretical understanding of probability flow-based ODE samplers in SGMs. \citep{chen2023restoration} proposes a new operational interpretation of the deterministic sampler (DDIM-type sampler) in the diffusion model. Under certain Lipschitz assumptions like strict high-order smoothness of the data distribution, it proves that the discretization process can approximate the real reverse process to the specified error and gives the corresponding total variation error bounds. \citep{li2023towards} only makes assumptions about the target data distribution and constructs the error bounds between the generated distribution and the true distribution with respect to the total variation distance (TV) under deterministic sampling (ODE) and non-deterministic sampling (SDE) respectively. The non-asymptotic convergence theory of the general probabilistic stream ODE sampler at second-order Wasserstein distance is constructed to provide a theoretical analysis of the iterative complexity of the deterministic ODE sampler under different forward SDE choices. \citep{liang2024non} proposes a new accelerated stochastic DDPM sampler and analysis technique, widening the target distribution to include Gaussian mixture, finite variance (early stop), and distributions with Lipschitz Hessian logarithmic density, providing more relaxed assumptions for non-smooth target distributions. It gives the first accelerated convergence result for the Gaussian mixture distribution and improves the dimensional dependence of the convergence rate of the bounded support distribution. \citep{tang2024contractive} proposes a contraction diffusion probability model (CDPMs), which improves the diffusion model by requiring the contractility of the reverse process. The contractility of the reverse process can prevent the expansion of fractional matching errors and is robust to fractional matching errors and discretization errors. \citep{chen2024probability} 
provide the first convergence guarantees for the probability flow ODE with overdamped Langevin corrector. 
In the realm of fractional diffusion models, the theoretical assurance of model convergence has consistently stood as a pivotal concern. Historically, a substantial portion of prior research has hinged upon relatively stringent data assumptions. This constraint has, to a certain degree, circumscribed the practical application of these models in real - world, intricate data settings. Recently, a significant breakthrough was achieved in \citep{conforti2025kl}, innovatively conducting a deep analysis of the Kullback - Leibler (KL) divergence convergence of fractional diffusion models under minimal data assumptions. Their work offers robust theoretical underpinnings for the model's convergence across a broader spectrum of data distributions. 
\citep{liang2025low} also abandon the assumptions of smoothness or log - concavity of the target distribution, significantly expanding the applicable scope of diffusion models. In low - dimensional structures, it is found that the DDPM sampler has improved the iteration complexity of total - variation convergence from $\tilde{O}(k / \varepsilon^{2})$ to $\tilde{O}(k / \varepsilon)$, while the DDIM sampler, \citep{song2020denoising}, can reach $\tilde{O}(k/\varepsilon)$. These results clarify the efficient sampling capabilities of the two samplers in low - dimensional structures and open up new directions for the research of diffusion models. 
\\

\medskip

\noindent \textbf{Convergence theory for conditional diffusion models.}
Conditional diffusion models are widely applied in various fields, including modern image synthesis. They introduce conditional information to guide sample generation. At present, some problems about the convergence rate of conditional fractional function estimation and conditional distribution estimation have been explored.
\citep{chung2022come} introduces a random contraction operation in the diffusion process, which effectively reduces the distribution range of the sample in the potential space and speeds up the convergence to the target distribution.
\citep{zhou2023deep} develops a deep generation method based on the GAN framework and proves that the conditional generator converges to the underlying conditional distribution under suitable conditions.
\citep{baldassari2024conditional} defines the conditional score for the first time in the infinite-dimensional setting, enabling the use of the conditional score diffusion model for Bayesian inference in the infinite-dimensional case and laying the foundation for subsequent analysis and application.
\citep{fu2024unveil} establishes the theoretical framework of the conditional diffusion model using the classifier-less guidance method and adopts a non-parametric statistical perspective. Assuming that the real conditional distribution has H$\ddot{o}$lder regularity, the sample complexity bound is derived under the condition of limiting the value range of the target data.
\citep{jiao2024model} proposes a new method for deep non-parametric regression statistical inference using the conditional diffusion model. This method solves the problem of missing asymptotic properties in deep non-parametric regression by transforming the depth estimation paradigm into a conditional mean estimation platform.
\citep{wu2024theoretical} shows that diffusion guidance can improve prediction confidence and reduce distribution diversity under certain conditions.
\citep{tang2024conditional} mainly studies the theoretical properties of the conditional diffusion model in conditional distribution estimation. Considering the low-dimensional manifold structure that may exist in data and covariates, it is shown that the convergence rate under the total variation distance is the minimax optimal under certain conditions. The results cover special cases such as unconditional distribution estimation and non-parametric mean regression, and explain the error bounds under the Wasserstein distance in detail, including the source and significance of the two errors and possible phase transition phenomena.
\citep{liang2024theory} For score-mismatched diffusion models and zero-shot conditional samplers, the first performance guarantee and explicit dimensional dependence of the general score-mismatched diffusion sampler are proposed, and the convergence guarantee and related theoretical analysis are provided for these cases, which are common in practical applications but have received little theoretical study.
\citep{li2024provable} To solve the problem of slow sampling speed of the fractional diffusion model, an acceleration scheme without training is proposed. Under the minimum assumption, the scheme can make the sampler reach the \(\epsilon\) precision (total variation) in the iteration of \(\tilde{O}(d^{5/4} / \sqrt{\epsilon})\), which improves the iteration complexity of the standard sampler \(\tilde{O}(d / \epsilon)\). The convergence theory does not rely on strict assumptions about the target distribution or the guarantee of high-order fraction estimation, only \(L^{2}\) accurate fraction estimation and the second moment of the target distribution are limited, which provides theoretical support for efficient sampling and widens the application range of the model.

\subsection{Notation}
We use bold normal font letters to denote vectors, $\mathbf{y}_{0} \in R^d$ represents the ground-truth response variable and its covariates $\mathbf{x}\in R^d_{x}$, $f_{\phi}(\mathbf{x})$ denotes a pre-trained conditional mean estimator, $t\in[0,T]$ denote the time variable ($T > 0$ is the time horizon) and $\mathbf{y}_{1:T}$ represents a sequence of intermediate predictions which made by the diffusion model.
$R^d$ is the $d$-dimensional Euclidean space.
We denote the original distribution and the generated distribution as $q(\mathbf{y}_{0}|\cdot)$ and $p(\mathbf{y}_{0}|\cdot)$.
% $d$ and $\theta$ imply differential and partial differential operators, respectively. 
$\nabla$ means the partial ifferential operator of $\mathbf{y}$.
For any two functions $h(x)$ and $g(x)$, we adopt the notation $h(x) = O(g(x))$ means $h(x) \lesssim g(x)$ and the notation $h(x) = o(g(x))$
means $h(x)/g(x) \rightarrow 0$ as $x$ tend to infinity.
$\|\cdot\|_2$ and $\|\cdot\|_{\infty}$ provide the $L_{2}$ norm of a vector or matrix and $L_\infty$ norm of a scalar function, respectively.
$W_2(\cdot,\cdot)$ represents the Wasserstein distance between two probability distributions. 
%Definition of \(L^p\) spaces
We denote $L^1(X,\mu)$ as the space consisting of all measurable functions $g$ such that \(\int_X|g(x)|d\mu(x)<\infty\), 
% For a measure space \((X,\mu)\), \(L^p(X,\mu)\) (where \(p\geq1\)) is the space consisting of all measurable functions \(f:X\to\mathbb{R}\) (or \(\mathbb{C}\)) such that \(\int_X|f(x)|^p d\mu(x)<\infty\).
%Definition of the function space \(C^{n}\)
and denote $C^{n}$ as the notation for a class of function spaces. For an open set $U\subseteq R^{d}$, $C^{2}(U)$ represents the set of functions that are twice continuously differentiable on $U$.
% That is, if a function \(f\) belongs to \(C^{2}\), then the first - order partial derivatives and second - order partial derivatives of \(f\) exist and are continuous.
% For example, for a binary function \(f(x,y)\), if \(f\in C^{2}\), then \(\frac{\partial f}{\partial x}\), \(\frac{\partial f}{\partial y}\), \(\frac{\partial^{2}f}{\partial x^{2}}\), \(\frac{\partial^{2}f}{\partial y^{2}}\), and the mixed partial derivatives \(\frac{\partial^{2}f}{\partial x\partial y}\) and \(\frac{\partial^{2}f}{\partial y\partial x}\) all exist and are continuous.

% The gradient of a real - valued function $\rho$ with respect to the spatial variable and the time - derivative of $\rho$ are denoted by $\nabla\rho=\left(\frac{\partial\rho}{\partial x_1},\frac{\partial\rho}{\partial x_2},\ldots,\frac{\partial\rho}{\partial x_d}\right)^{\top}$, and $\partial_t\rho$, respectively. The Laplacian of $\rho$ is given as $\Delta\rho=\nabla\cdot(\nabla\rho)$. Here, $\nabla\cdot F=\frac{\partial F_1}{\partial x_1}+\frac{\partial F_2}{\partial x_2}+\cdots+\frac{\partial F_d}{\partial x_d}$ denotes the divergence of
% $F=(F_1,F_2,\ldots,F_d) \text{ with respect to the spatial variable } x.$
% The Hessian of $\rho$ is a $d\times d$ matrix given by $(D^2\rho)_{ij}=\frac{\partial^2\rho}{\partial x_i\partial x_j}$ for $1\leq i,j\leq d$. We indicate Gaussian distribution with mean $\mu$ and covariance $\Sigma$ by $\mathcal{N}(\mu,\Sigma)$, and $L_2$ norm by $\|\cdot\|$. 

\subsection{Organization}
The remainder of this paper is structured as follows. Section \ref{sec2} reviews several theories associated with both the original diffusion model and the score based diffusion model. Notably, we focus on a convergence outcome of the unconditional diffusion model and also revisit the fundamental principles underlying the conditional diffusion model utilized for classification and regression tasks.
Section \ref{sec3} is dedicated to the establishment of the first non-asymptotic theory that pertains to the relationship between the original distribution and the generated distribution. This is achieved within the framework of the conditional diffusion model fraction, specifically designed for classification and regression applications.
Section \ref{sec4} discusses the error bounds of the conditional score function and its corresponding estimates building upon existing theoretical foundations. Moreover, we present the explicit upper bounds for the convergence of the conditional score function.
Section \ref{sec5} primarily validates the proposed theory through simulation experiments.

\section{Background}\label{sec2}%Preliminaries
\noindent\textbf{Diffusion models.} We essentially adopt the notation from \cite{ho2020denoising}. In forward process, $q(\mathbf{y}_{t})$ denote the distribution of $\mathbf{y}_{t}$. Thus, $q(\mathbf{y}_{0})$ represents the original distribution, while $q(\mathbf{y}_{t}|\mathbf{y}_{t-1})$ characterizes the stepwise noise addition process, which is non-parametric. 
In inverse process, we employ a neural network $p_{\theta}(\mathbf{y}_{t-1}|\mathbf{y}_{t})$ to approximate the posterior probability distribution $p(\mathbf{y}_{t-1}|\mathbf{y}_{t})$, where $\theta$ encapsulates all the parameters of the neural network. The generated distribution denotes as $p_{\theta}(\mathbf{y}_{0})$. Above processes are formally defined as follows:
\begin{equation}\label{A1}
\begin{aligned}
&q(\mathbf{y}_{t}|\mathbf{y}_{t-1}) =N(\sqrt{1-\beta_{t}} \mathbf{y}_{t-1},\beta_{t} \mathbf{I}), \ q\left(\mathbf{y}_{1: T} \mid \mathbf{y}_{0}\right)=\prod_{t=1}^{T} q\left(\mathbf{y}_{t} \mid \mathbf{y}_{t-1}\right),\\
&p_{\theta}(\mathbf{y}_{0:T})= p(\mathbf{y}_{T}) \prod\limits_{t=1}^{T} p_{\theta}(\mathbf{y}_{t-1}|\mathbf{y}_{t}),\ p_{\theta}(\mathbf{y}_{t-1}|\mathbf{y}_{t}) = N(\mathbf{\mu}_{\theta}(\mathbf{y}_{t},t),\ \Sigma_{\theta}(\mathbf{y}_{t},t)).
\end{aligned}
\end{equation}
where $\{\beta_{t}\}_{t=1:T} \in (0,1)^{T}$ mean the diffusion schedule.
\cite{kingma2013auto} used the variational inference as its loss function, \cite{ho2020denoising} similarly considered the variational lower bound (VLB) as loss function, which is defined as
\begin{equation*}\label{A2}
\begin{aligned}
% E_{q\left(\mathbf{y}_{0}\right)}\left[-\log p_{\theta}\left(\mathbf{y}_{0}\right)\right] 
%& \leq  E_{q\left(\mathbf{x}_{0: T}\right)}\left[\log \frac{q\left(\mathbf{x}_{1: T} \mid \mathbf{x}_{0}\right)}{p_{\theta}\left(\mathbf{x}_{0: T}\right)}\right]\\
% &\leq 
\textup{E}_{q\left(\mathbf{y}_{0: T}\right)} [ D_{KL}(q(\mathbf{y}_{T}|x_{0})|| p_{\theta}(\mathbf{y}_{T})) + \sum\limits_{t=2}^{T}  D_{KL}(q(\mathbf{y}_{t-1}|\mathbf{y}_{t},\mathbf{y}_{0})|| p_{\theta}(\mathbf{y}_{t-1}|\mathbf{y}_{t})) -\log p_{\theta}(\mathbf{y}_{0}||\mathbf{y}_{1})  ].
\end{aligned}
\end{equation*}

% \textbf{DDIM.} Assuming the forward process is non-Markovian\textcolor{blue}{,} its backward process can rely on more previous samples at each step to predict new samples. The main distributions involved are as follows:
% \begin{equation}\label{B1}
% \begin{aligned}
% &q_{\sigma}(\mathbf{x}_{1},\cdots,\mathbf{x}_{T}|\mathbf{x}_{0}) = \prod_{t=1}^{T} q(\mathbf{x}_{t}|\mathbf{x}_{t-1},\mathbf{x}_{0}),
% %\end{equation}
% %\begin{equation}\label{C3}
% &q_{\sigma}(\mathbf{x}_{t-1}|\mathbf{x}_{t},\mathbf{x}_{0}) = N(\sqrt{\bar{\alpha}_{t-1}}\mathbf{x}_{0}+\sqrt{1-\bar{\alpha}_{t-1}-\sigma_{t}^2} \ \frac{\mathbf{x}_{t}-\sqrt{\bar{\alpha}_{t}}\mathbf{x}_
% {0}}{\sqrt{1-\bar{\alpha}_{t}}},\sigma_{t}^2 \mathbf{I}),
% \end{aligned}
% \end{equation}
% where $\sigma_{t}$ represents a variable parameter whose size controls the randomness of the forward process\citep{song2020denoising}. If we set $\sigma_{t}=\sqrt{\frac{1-\bar{\alpha}_{t-1}}{1-\bar{\alpha}_{t}}} \sqrt{1-\frac{\bar{\alpha}_{t}}{\bar{\alpha}_{t-1}}}$, then DDIM is equal to DDPM for the generating process.\\

\noindent\textbf{CARD.} We follow the notation of \cite{han2022card}, which inject covarite $\mathbf{x}$ and predicted conditional expectation $f_{\phi}(\mathbf{x})$ into all forward and reverse processes, providing an accurate estimate of the conditional distribution $p(\mathbf{y}|\mathbf{x})$. 
The model is trained and inferred the same as the original DDPMs \citep{ho2020denoising} in both classification and regression tasks. It is defined as
\begin{equation}\label{C1}
\begin{aligned}
&q\left(\mathbf{y}_{t} \mid \mathbf{y}_{t-1}, f_{\phi}(\mathbf{x})\right) = N\left( \sqrt{1-\beta_{t}} \mathbf{y}_{t-1}+\left(1-\sqrt{1-\beta_{t}}\right) f_{\phi}(\mathbf{x}), \beta_{t} \mathbf{I}\right),\\
&q\left(\mathbf{y}_{t} \mid \mathbf{y}_{0}, f_{\phi}(\mathbf{x})\right) = N\left( \sqrt{\bar{\alpha}_{t}} \mathbf{y}_{t-1}+\left(1-\sqrt{\bar{\alpha}_{t}}\right) f_{\phi}(\mathbf{x}), \beta_{t} \mathbf{I}\right),\\
&q\left(\mathbf{y}_{t-1} \mid \mathbf{y}_{t},\mathbf{y}_{0}, f_{\phi}(\mathbf{x}))\right) = N\left(\mathbf{\tilde{\mu}}(\mathbf{y}_{t},\mathbf{y}_{0},f_{\phi}(\mathbf{x})), \tilde{\beta}_{t} \mathbf{I} \right),
\end{aligned}
\end{equation}
where $\alpha_{t} = 1-\beta_{t},\  \bar{\alpha}_{t}=\prod_{t} \alpha_{t}$ and 
\begin{equation*}
\begin{aligned}
& \mathbf{\tilde{\mu}}(\mathbf{y}_{t},\mathbf{y}_{0},f_{\phi}(\mathbf{x})) = \frac{\beta_{t} \sqrt{\bar{\alpha}_{t-1}}}{1-\bar{\alpha}_{t}} \mathbf{y}_{0} + \frac{(1-\bar{\alpha}_{t-1}) \sqrt{\alpha_{t}}}{1-\bar{\alpha}_{t}} \mathbf{y}_{t} + \left(  1+\frac{(\sqrt{\bar{\alpha}_{t}}-1)(\sqrt{\alpha_{t}} + \sqrt{\bar{\alpha}_{t-1}})}{1-\bar{\alpha}_{t}} \right) f_{\phi}(\mathbf{x}),
%& \ \ \ \ \ \ \ \ \ \ \ \ \ \ \ \ \ \ \ \ \ \ \ \ \ \ + \left(  1+\frac{(\sqrt{\bar{\alpha}_{t}}-1)(\sqrt{\alpha_{t}} + \sqrt{\bar{\alpha}_{t-1}})}{1-\bar{\alpha}_{t}} \right) f_{\phi}(\boldsymbol{x}),\\
& \tilde{\beta}_{t} = \frac{1-\bar{\alpha}_{t-1}}{1-\bar{\alpha}_{t}} \beta_{t}.
\end{aligned}
\end{equation*}
and assume endpoint of our diffusion process as $p(\mathbf{y}_{T}|\mathbf{x})=N(f_{\phi}(\mathbf{x}),\mathbf{I})$. 
We find that the second mean term in Eq.(\ref{C1}) (i.e. $\sqrt{\bar{\alpha}_{t}} \mathbf{y}_{t-1}+\left(1-\sqrt{\bar{\alpha}_{t}}\right) f_{\phi}(\mathbf{x})$) can be viewed as an
interpolation between true data $\mathbf{y}_{0}$ and the predicted conditional expectation $f_{\phi}(\mathbf{x})$, which gradually
changes from the former to the latter throughout the forward process. 
Meanwhile, loss function is similar to DDPM. 
% which can be described as 
% \begin{equation}\label{C2}
% \log p_{\theta}(\mathbf{y}_{0}|\mathbf{x}) = \log \int p_{\theta}(\mathbf{y}_{0:T}|\mathbf{x}) d\mathbf{y}_{1:T} \geq E_{q(\mathbf{y}_{1:T}|\mathbf{y}_{0},\mathbf{x})} \left[ \log \frac{p_{\theta}(\mathbf{y}_{0:T}|\mathbf{y}_{0},\mathbf{x})} {q(\mathbf{y}_{1:T}|\mathbf{y}_{0},\mathbf{x})} \right].
% \end{equation}

\medskip

\noindent\textbf{Score match.} The score function of a distribution $q(\mathbf{y})$ is the gradient of its log probability density function, $\nabla \log q(\mathbf{y})$, indicating how rapidly the probability density changes. This concept forms the basis for score-based models (SGMs), denoted as $s_{\theta}(\mathbf{y})$, which are proven by \cite{song2020score} and have described by stochastic differential equations (SDEs). These SDEs simplify the calculation by setting the diffusion coefficient to a scalar that does not depend on the variable $\mathbf{y}$. Additionally, DDPM is a specific instance of an SGM, and its corresponding SDEs are elaborated upon in the literature. The SDEs with respect to forward and reverse process of DDPMs are described as
\begin{equation}\label{D1}
\begin{aligned}
& d\mathbf{y} =-\frac{\beta_{t}}{2}\mathbf{y}dt +\sqrt{\beta_{t}}dw,%\ \ dw \sim N(0,\Delta t),
& d\mathbf{y} = \left[ -\frac{\beta_{t}}{2}\mathbf{y} - \beta_{t} \nabla \log q(\mathbf{y}_{t})  \right] dt + \sqrt{\beta_{t}} d\bar{w},%\ \ dw \sim N(0,\Delta t),
\end{aligned}
\end{equation}
% where $dw = w(t+\Delta t)-w(t) \sim N(0,\Delta t)$ refers to the increment in the general Wiener process.\\
where $w,\bar{w}$ are the standard Wiener processes (also known as Brownian motions) corresponding to the time flowing from 0 to T and backwards from T to 0, respectively.

Specially, \cite{kwon2022score} have derived the Fokker-Planck equation of DDPMs with the form
\begin{equation}\label{D2}
\begin{aligned}
 &\partial_{t} q(\mathbf{y}_{t}) - \frac{\beta_{t}}{2} \nabla\left(\mathbf{y}_{t} q(\mathbf{y}_{t}) \right) - \frac{\beta_{t}}{2} \nabla\cdot( \nabla q(\mathbf{y}_{t})) = 0,\\
 &\partial_{t} p(\mathbf{y}_{t})  -\beta_{t} \nabla\left( p(\mathbf{y}_{t})\left(\frac{1}{2} \mathbf{y}_{t} + \nabla \log p(\mathbf{y}_{t})\right) \right) +  \frac{\beta_{t}}{2}  \nabla\cdot( \nabla p(\mathbf{y}_{t})) = 0,
\end{aligned}
\end{equation}
and proved that under appropriate assumptions, there exists a convergent upper bound on the Wasserstein distance between the data distribution and the generated distribution, that is,
\begin{equation}\label{D3}
\begin{aligned}
W_{2}(q(\mathbf{y}_{0}),p(\mathbf{y}_{0})) \leq \int_{0}^{T} g^{2}(t) I(t) \left(\textup{E}_{q(y_{t})}\left[ || \nabla \log q(\mathbf{y}_{t})-s_{\theta}(\mathbf{y},t) ||^{2} \right]\right)^{\frac{1}{2}}dt + I(T)W_{2}(q(\mathbf{y}_{T}),p(\mathbf{y}_{T})),
\end{aligned}
\end{equation}
where the function $I:[0,T] \rightarrow (0,\infty)$ is defined as $I(t) = \exp\left\{\int_{0}^{t} (L_{f}(r) + L_{s}(r)g^{2}(r))dr\right\}.$
If the time step $T$ is large enough with $q(\mathbf{y}_{T})=p(\mathbf{y}_{T})$, then $W(q(\mathbf{y}_{T}),p(\mathbf{y}_{T}))=0$. And if the loss is small enough with $\textup{E}_{q(\mathbf{y}_{t})}\left[ || \nabla \log q(\mathbf{y}_{t})-s_{\theta}(\mathbf{y},t) ||^{2} \right]\rightarrow0$, then $W(q(\mathbf{y}_{0}),p(\mathbf{y}_{0})) \rightarrow0$. This indicates that the error between the original and generated distribution is ideally small.

\section{Upper Bound of Original and Generated Distributions}\label{sec3}
In this section, we first build the SDEs representation of CARD, and generalize it through the application of Fokker-Planck equation.    Under specific assumptions, we provide an upper bound of original and generated distribution by linking the Fokker-Planck equation to the Wasserstein distance.

\subsection{SDEs form}
The noise perturbations employed in CARD can be regarded as discretized approximations of SDEs. We provide the detailed proof later in Appendix \ref{appendixA1} and \ref{appendixA2}. 

Given the conditional distribution between the response variable $\mathbf{y}$, the covariate $\mathbf{x}$, and the conditional mean function $f_{\phi}(\mathbf{x})$ for both forward and reverse processes, as specified by Eq.(\ref{C1}), the distribution can represent through SDEs as follows.
\begin{equation}\label{E1}
d\mathbf{y} =-\frac{\beta_{t}}{2}(\mathbf{y}-f_{\phi}(\mathbf{x}))dt +\sqrt{\beta_{t}}dw,
\end{equation}
\begin{equation}\label{E2}
d\mathbf{y} = - \left[ \frac{1}{2} (\mathbf{y}-f_{\phi}(\mathbf{x}))- \frac{\partial \log p(\mathbf{y}|f_{\phi}(\mathbf{x}))}{\partial \mathbf{y} }\right] \beta_{t} dt + \sqrt{\beta_{t}}d\bar{w},
\end{equation}
where $dw,d\bar{w}$ are all standard Wiener processes, and time is from 0 to $T$ in Eq.(\ref{E1}), and time flows backwards from $T$ to 0 in Eq.(\ref{E2}). 

Similar to \cite{song2020denoising}, we refer to the gradient term $\partial \log q(\mathbf{y}_{t}|f_{\phi}(\mathbf{x})) / \partial \mathbf{y}_{t}$ in Eq.(\ref{E2}) as the conditional score function. The above two SDEs describe the case of continuous variables. When $\Delta t \rightarrow 0$, their discretized representation can be derived. This conclusion stems from the framework of SGMs, where neural networks are employed to accurately estimate the score function, and a numerical solver is utilized for sample generation.

We further employ the Fokker-Planck equation to analysis. Let $q(\mathbf{y}_{t}|f_{\phi}(\mathbf{x}))$ and $p(\mathbf{y}_{t}|f_{\phi}(\mathbf{x}))$ denote the probability distribution of the forward and reverse processes about the random variable $y$ at time $t$, respectively. Then we have
\begin{equation}\label{E3}
\begin{aligned}
&\frac{\partial }{\partial t}q(\mathbf{y}_{t}|f_{\phi}(\mathbf{x})) + \frac{\partial}{\partial \mathbf{y}_{t}} \left[-\frac{1}{2} \beta_{t} (\mathbf{y}_{t}-f_{\phi}(\mathbf{x})) q(\mathbf{y}_{t}|f_{\phi}(\mathbf{x})) \right] - \frac{1}{2} \beta_{t} \frac{\partial^{2}}{\partial \mathbf{y}_{t}^{2}} q(\mathbf{y}_{t}|f_{\phi}(\mathbf{x}))=0,\\
&\frac{\partial }{\partial t}p(\mathbf{y}_{t}|f_{\phi}(\mathbf{x}))+\frac{\partial}{\partial \mathbf{y}_{t}} \left[-\left(\frac{1}{2} (\mathbf{y_{t}}-f_{\phi}(\mathbf{x}))-\frac{\partial }{\partial \mathbf{y}_{t}}\log p(\mathbf{y}_{t}|f_{\phi}(\mathbf{x})) \right)\beta_{t}p(\mathbf{y}_{t}|f_{\phi}(\mathbf{x})) \right]
+\frac{1}{2} \beta_{t} \frac{\partial^{2} }{\partial \mathbf{y}_{t}^{2}}p(\mathbf{y}_{t}|f_{\phi}(\mathbf{x}))=0,
\end{aligned}
\end{equation}
where $q(\mathbf{y}_{0}|f_{\phi}(\mathbf{x}))$ denotes the original distribution and $p(\mathbf{y}_{T}|f_{\phi}(\mathbf{x}))$ denotes the data distribution at the beginning of reverse process. 

Building upon the established conclusions, equivalent ordinary differential equations (ODEs) can be derived. Following the principles of SGM, for all forward processes, there exists a deterministic process whose trajectory shares the same marginal distribution function $\mathbf{y}_{t}$ \citep{song2020score}. This deterministic process satisfies an ODE.
We add the noise variance, denoted as $\beta_{t}$, for generality in Eq.(\ref{E1}). In practical applications, varying levels of noise are often introduced into the data to optimize performance. For the original distribution, adjusting the noise variances $\beta_{t}$ in the forward process must yield the same marginal distribution function $\mathbf{y}_{t}$ to ensure consistency. Assuming there are two different noise variances, denoted as $\beta_{t}^{1},\beta_{t}^{2}$ and $\beta_{t}^{1} \leq \beta_{t}^{2}, t \in [0,T]$, we can easily verify Eq.(\ref{E3}) has the following form
\begin{equation*}\label{E4}
\begin{aligned}
\frac{\partial q(\mathbf{y}_{t}|f_{\phi}(\mathbf{x}))}{\partial t} 
% &=- \frac{\partial}{\partial \mathbf{y}_{t}}\left[-\frac{1}{2} \beta_{t} (\mathbf{y_{t}}-f_{\phi}(\mathbf{x})) q(\mathbf{y}_{t}|f_{\phi}(\mathbf{x}))\right] - \frac{1}{2} \beta_{t} \frac{\partial^{2} q(\mathbf{y}_{t}|f_{\phi}(\mathbf{x}))}{\partial \mathbf{y}_{t}^{2}}\\
% &=- \frac{\partial}{\partial \mathbf{y}_{t}}\left[-\frac{1}{2} \beta_{t} (\mathbf{y_{t}}-f_{\phi}(\mathbf{x})) q(\mathbf{y}_{t}|f_{\phi}(\mathbf{x})) - \frac{1}{2} (\beta_{t}-\sigma_{t}) \frac{\partial q(\mathbf{y}_{t}|f_{\phi}(\mathbf{x}))}{\partial \mathbf{y}_{t}}\right] + \frac{1}{2} \sigma_{t} \frac{\partial^{2} q(\mathbf{y}_{t}|f_{\phi}(\mathbf{x}))}{\partial \mathbf{y}_{t}^{2}}\\
&=- \frac{\partial}{\partial \mathbf{y}_{t}}\left[\left(-\frac{1}{2} \beta_{t}^{2} (\mathbf{y_{t}}-f_{\phi}(\mathbf{x})) -\frac{1}{2} (\beta_{t}^{2}-\beta_{t}^{1}) \frac{\partial \log q(\mathbf{y}_{t}|f_{\phi}(\mathbf{x}))}{\partial \mathbf{y}_{t}} \right) q(\mathbf{y}_{t}|f_{\phi}(\mathbf{x})) \right] + \frac{1}{2} \beta_{t}^{1} \frac{\partial^{2} q(\mathbf{y}_{t}|f_{\phi}(\mathbf{x}))}{\partial \mathbf{y}_{t}^{2}}.
\end{aligned}
\end{equation*}
It is obvious that the marginal distribution is consistent for different noise variances. If we let $\beta_{t}^{1}=0$, then
\begin{equation*}\label{E5}
\frac{\partial q(\mathbf{y}_{t}|f_{\phi}(\mathbf{x}))}{\partial t}  =- \frac{\partial}{\partial \mathbf{y}_{t}}\left[\left(-\frac{1}{2} \beta_{t}^{2} (\mathbf{y_{t}}-f_{\phi}(\mathbf{x})) -\frac{1}{2} \beta_{t}^{2} \frac{\partial \log q(\mathbf{y}_{t}|f_{\phi}(\mathbf{x}))}{\partial \mathbf{y}_{t}} \right) q(\mathbf{y}_{t}|f_{\phi}(\mathbf{x})) \right].
\end{equation*}
This leads to a generalized ODE representation of CARD as follows.
\begin{equation}\label{E6}
d\mathbf{y} =\left[-\frac{1}{2} \beta_{t} (\mathbf{y}-f_{\phi}(\mathbf{x})) -\frac{1}{2} \beta_{t}\frac{\partial \log q(\mathbf{y}|f_{\phi}(\mathbf{x}))}{\partial \mathbf{y}} \right]dt.
\end{equation}

\subsection{Convergence Bound within the Wasserstein metric}
We focus on the Wasserstein distance between original distribution $q(\mathbf{y}_{0}|f_{\phi}(\mathbf{x}))$ and the generated distribution $p(\mathbf{y}_{0}|f_{\phi}(\mathbf{x}))$ by $T$ steps. The detailed proof is given in Appendix \ref{appendixA3} and \ref{appendixA4}.

The Fokker-Planck equation can be interpreted as a gradient flow in the probability space defined by the Wasserstein distance, thereby allowing for the quantification of Wasserstein distance between two distributions using the Fokker-Planck equation. 
This framework offers a precise method for measuring the dissimilarity between probability distributions under the Wasserstein metric. In general, we define the 2-Wasserstein or Monge-Kantorovich distance as
\begin{equation}\label{F1}
    W_2(P,Q) := \inf_{\pi \in \Pi(\mu, \nu)} \left\{ \int_{\mathbb{R}^d\times\mathbb{R}^d} \|x - y\|_{2}^2 d\pi(x,y) \right\}. 
\end{equation}
Here, $\Pi(P,Q)$ is the set of all couplings $\pi$ of $P$ and $Q$ where a Borel probability measure $\pi$ on $R^d\times R^d$ satisfies $\pi(A\times R^d)=P(A)$ and $\pi(R^d\times B)=Q(B)$ for all measurable subsets $A, B \subset R^d$.

Meanwhile, \cite{kwon2022score} have indicated that under appropriate assumptions about drift and diffusion coefficients, the solution of the Fokker-Planck equation satisfies the Wasserstein contraction property, which plays an important role in our analysis.

To begin with, we impose some assumptions on coefficients about SDEs and the conditional score estimates.

\medskip

\begin{assumption}\label{assum1}
\textbf{(Drift coefficient).}
Suppose that the drift coefficient $-\frac{1}{2} \beta_{t} (\mathbf{y}_{t}-f_{\phi}(\mathbf{x}))$ in Eq.(\ref{E1}) is Lipschitz continuous with respect to the variable $\mathbf{y}_{t}$, and linear with respect to $\mathbf{y}_{t},f_{\phi}(x)$ respectively. 
There exists a constant $l_{1}(t) \in (0,\infty)$, for all $\mathbf{y}_{1},\mathbf{y}_{2} \in R^{d}$, we assume
% \begin{equation*}\label{F2}
% ||-\frac{1}{2} \beta_{t} (\mathbf{y}_{1}-f_{\phi}(\mathbf{x}))+\frac{1}{2} \beta_{t} (\mathbf{y}_{2}-f_{\phi}(\mathbf{x}))|| \leq l_{1}(t) ||\mathbf{y}_{1}-\mathbf{y}_{2}||,
% \end{equation*}
% We can simplify it as
\begin{equation}\label{F3}
||-\frac{1}{2} \beta_{t} (\mathbf{y}_{1}-\mathbf{y}_{2})||_{2} \leq l_{1}(t) ||\mathbf{y}_{1}-\mathbf{y}_{2}||_{2}.
\end{equation}

\noindent and there exists constants $C_{1},C_{2},C_{3},C_{4},C_{5},C_{6},C_{7} \in (-\infty,\infty)$, for all $\mathbf{y} \in R^{d}$, we assume
\begin{equation}\label{F4}
\begin{aligned}
& -\frac{1}{2} \beta_{t} (\mathbf{y}-f_{\phi}(\mathbf{x})) \mathbf{y} \leq C_{1} ||\mathbf{y}||_{2}^{2} + C_{2}\mathbf{y}f_{\phi}(\mathbf{x}) + C_{3},\\
& -\frac{1}{2} \beta_{t} (\mathbf{y}-f_{\phi}(\mathbf{x})) (\mathbf{y}-f_{\phi}(\mathbf{x})) \leq C_{4} ||\mathbf{y}||_{2}^{2} + C_{5}\mathbf{y}f_{\phi}(\mathbf{x}) + C_{6}||f_{\phi}(\mathbf{x})||_{2}^{2} + C_{7}. 
\end{aligned}
\end{equation}
\end{assumption}

\begin{assumption}\label{assum2}
\textbf{(Diffusion coefficient).}
Suppose that the diffusion coefficient (i.e., noise variance progressively added by the forward process) is bounded. 
There exists a constant $C_{8} \in (0,\infty)$, for all $t \in [0,T]$, we assume $\frac{1}{C_{8}} < \beta_{t} < C_{8}$.
% \begin{equation}\label{F6}
% \frac{1}{C_{8}} < \beta_{t} < C_{8}.
% \end{equation}
\end{assumption}

\begin{assumption}\label{assum3}
\textbf{(Score function estimator).}
Suppose that the score function estimate $s_{\theta}(\mathbf{x},\mathbf{y}_{t},f_{\phi}(\mathbf{x}),t)$ is one-sided Lipschitz continuous with respect to the variable $\mathbf{y}$. There exists a constant $l_{2}(t) \in R$, for all $\mathbf{y}_{1},\mathbf{y}_{2} \in R^{d}$, we assume
\begin{equation}\label{F5}
[s_{\theta}(\mathbf{x},\mathbf{y}_{1},f_{\phi}(\mathbf{x}),t)-s_{\theta}(\mathbf{x},\mathbf{y}_{2},f_{\phi}(\mathbf{x}),t)](\mathbf{y}_{1}-\mathbf{y}_{2}) \leq l_{2}(t) ||\mathbf{y}_{1}-\mathbf{y}_{2}||_{2}^{2}.
\end{equation}
\end{assumption}

In order to ensure the integrability of the score function. We also  assume that expectations and integrals are finite, i.e., we assume
\begin{equation*}\label{F7}
\begin{aligned}
\textup{E}_{q(\mathbf{y}_{0},\mathbf{x}))} [||\log q(\mathbf{y}_{0}|f_{\phi}(\mathbf{x}))||_{2}],
\textup{E}_{q(\mathbf{y}_{0},\mathbf{x}))} [\log \max(||\mathbf{y}_{0}||_{2},1)] ,
\textup{E}_{p(\mathbf{y}_{T},\mathbf{x}))} [||\log p(\mathbf{y}_{T})||_{2}] ,
\textup{E}_{p(\mathbf{y}_{T},\mathbf{x}))} [\log \max(||\mathbf{y}_{0}||_{2},1)],
\end{aligned}
\end{equation*}
and
\begin{equation*}\label{F8}
\begin{aligned}
\int_{0}^{T} \textup{E}_{q(\mathbf{y}_{t},\mathbf{x}))} \left[ -\frac{1}{2} \beta_{t} (\mathbf{y}-f_{\phi}(\mathbf{x})) \right]^{2}dt,\ \ 
\int_{0}^{T} \textup{E}_{p(\mathbf{y}_{t},\mathbf{x}))} \left[\left(  -\frac{1}{2} \beta_{t} (\mathbf{y}-f_{\phi}(\mathbf{x})) \right)^{2}  -\beta_{t} s_{\theta}(\mathbf{x},\mathbf{y}_{t},f_{\phi}(\mathbf{x}),t) \right]^{2} dt.
\end{aligned}
\end{equation*}
is finite.
Then we obtain the score function is Lebesgue integrable on $R^d$ (i.e., $L^{1}(R^{d})$), which implies that
\begin{equation*}\label{F9}
\int_{0}^{T} \textup{E}_{q(\mathbf{y}_{t}|f_{\phi}(\mathbf{x}))}[||\frac{\partial \log q(\mathbf{y}_{t}|f_{\phi}(\mathbf{x}))} { \partial \mathbf{y}_{t}}||_{2}^{2}]dt < \infty.
\end{equation*}
% i.e. $\partial \log q(\mathbf{y}_{t}|f_{\phi}(\mathbf{x})) / \partial \mathbf{y}_{t} \in L^{1}(R^{d})$

Additionally, we make some constraints on the condition distributions. 
Suppose that the solutions of Eq.(\ref{E1}) and Eq.(\ref{E2}) are $q(\mathbf{y}_{t}|f_{\phi}(\mathbf{x}))$ and $p(\mathbf{y}_{t}|f_{\phi}(\mathbf{x}))$ respectively, and they are positive, second-order continuously differentiable probability density functions (i.e., $C^{2}(\mathbb{R}^{d})$).
With the aim of guaranteeing the precise estimation of the majority of samples, we initially assume the probability distribution function $q(\mathbf{y}_{t}|f_{\phi}(\mathbf{x}))$ and $p(\mathbf{y}_{t}|f_{\phi}(\mathbf{x}))$ exhibit decays rapidly in the $d$ dimensional real space $R^{d}$. Specifically, there exists $m>0$ such that for all $t \in [0,T]$, we have $q(\mathbf{y}_{t}|f_{\phi}(\mathbf{x}))= O(\exp(-\|\mathbf{y}_{t}\|_{2}^{m}))$ and $p(\mathbf{y}_{t}|f_{\phi}(\mathbf{x}))=O(\exp(-\|\mathbf{y}_{t}\|_{2}^{m}))$. The big $O$ notation here represents an upper bound on the growth rate of the function, meaning that the values of these functions approach zero exponentially as $\|\mathbf{y}_{t}\|_{2} \rightarrow \infty$.
Surely, we also assume $q(\mathbf{y}_{0}|f_{\phi}(\mathbf{x})),p(\mathbf{y}_{T}|f_{\phi}(\mathbf{x})) \in C^{2}(R^{d})$, diffusion coefficient $\beta_{t} \in C^{2}([0,T])$, drift coefficient and score function satisfy $-\frac{1}{2} \beta_{t} (\mathbf{y}-f_{\phi}(\mathbf{x})),s_{\theta}(\mathbf{x},\mathbf{y}_{t},f_{\phi}(\mathbf{x}),t) \in C^{2}(R^{d} \times [0,T])$.

When the drift and diffusion terms satisfy the Lipschitz condition in Assumption \ref{assum1} and Assumption \ref{assum2}, their variations in the state space are relatively smooth. Intuitively, the values of terms corresponding to different states will not differ too much. This is because it restricts the sensitivity of terms to changes in the state variable and avoids the system from producing vastly different long-term behaviors due to small initial differences.
Technically, Assumption \ref{assum3} merely restricts the variability within the estimate of the score function and does not pertain to the disparity from the true value.  This aspect will be subjected to a more in-depth analysis in section 4.
These conditions are of great significance for SDEs, which ensure the existence and uniqueness of the solution of the equation under specific initial conditions and related to the stability of the equation which keeps the solution stable within a certain range, preventing unbounded changes or violent fluctuations. 

\begin{assumption}\label{assum4}
\textbf{(Conditional mean)}
Suppose that that the pre-trained network, which estimates the conditional mean, performs well. For any arbitrarily small constant $\delta>0$, there exists a training epoch $a_0$ such that when the actual training epochs $a>a_0$, we have $P(\| f_{\phi}(X) - \textup{E}[Y|X] \|_{2}<\delta) \to 1$ holds. 
\end{assumption}

From the previous definition, we know the predicted conditional mean $f_{\phi}(X)$ is the prior knowledge of the relation between $Y$ and $X$, and pre-trained to approximate $\textup{E}[Y|X]$. 
Assumption \ref{assum4} shows network performs well which means the network possesses adequate capacity, the training algorithm utilized is capable of converging to a point proximate to the global optimal solution, the volume of training data is sufficiently large and exhibit independent and identically distributed characteristics. 
Specifically, with an appropriate network architecture, when $f_{\phi}(X)$ performs well and the model accurately captures the conditional distribution $Y|f_{\phi}(X)$, it closely approximates the real distribution $Y|X$. 

% Suppose predicted conditional mean is satisfied 
% $f_{\phi}(\mathbf{x}) = E[\mathbf{y}|\mathbf{x}] + o(1)$.
% Here, $o(1)$ represents a function that approaches $0$ under certain limiting conditions of $\mathbf{x}$.
% Assumption \ref{assum4} indicates the difference between $f_{\phi}(\mathbf{x})$ and $E[\mathbf{y}|\mathbf{x}]$ is negligible compared to $1$, that is, their error approaches $0$ in the limit case.

According to the above assumptions, we can state the following theorem. For the proof of Theorem \ref{the-1}, please refer to the Appendix \ref{appendixA3}.

\newtheorem{theorem}{Theorem}
\newtheorem{corollary}{Corollary}
\begin{theorem}
Let $q(\mathbf{y}_{0}|f_{\phi}(\mathbf{x})), p(\mathbf{y}_{0}|f_{\phi}(\mathbf{x}))$ denote the original and generated distributions, respectively. Then the 2-Wasserstein distance between the two distributions is upper bounded by
\begin{equation}\label{F10}
W_{2}(q(\mathbf{y}_{0}|f_{\phi}(\mathbf{x})),p(\mathbf{y}_{0}|f_{\phi}(\mathbf{x}))) \leq \int_{0}^{T} \beta_{t} M(t) \sqrt{H(t)}dt + M(T)W_{2}(q(\mathbf{y}_{T}|f_{\phi}(\mathbf{x})),p(\mathbf{y}_{T}|f_{\phi}(\mathbf{x}))),
\end{equation}
where
\begin{equation*}
\begin{aligned}
& M(t) = \exp \left\{\int^{t}_{0} (l_{1}(s)+l_{2}(s)\beta_{s})ds \right\},
& H(t) = \textup{E}_{q(\mathbf{y}_{t}|f_{\phi}(\mathbf{x}))} \left[ \left\|s_{\theta}(\mathbf{y}_{t},f_{\phi}(\mathbf{x}),t) - \frac{\partial \log q(\mathbf{y}_{t}|f_{\phi}(\mathbf{x}))} { \partial \mathbf{y}_{t}}  \right\|_{2}^{2} \right].
\end{aligned}
\end{equation*}
\label{the-1}
\end{theorem}

From the above theorem, $H(t)$ means the average error between the score function and its estimator. To accurately measure the change in $H(t)$, we similarly define $L_{1}(\phi,\theta,\lambda)$ as loss function \citep{song2021maximum}. The specific formula for it is as follows.
\begin{equation}\label{F11}
L_{1}(\phi,\theta,\lambda) = \frac{1}{2} \int^{T}_{0} \lambda(t) \textup{E}_{q(\mathbf{y}_{t}|f_{\phi}(\mathbf{x}))} \left[ ||s_{\theta}(\mathbf{y}_{t},f_{\phi}(\mathbf{x}),t) - \frac{\partial \log q(\mathbf{y}_{t}|f_{\phi}(\mathbf{x}))} { \partial \mathbf{y}_{t}}  ||^{2} \right] dt,
\end{equation}
where $\phi$ denotes the set of all parameters in the pre-trained model, $\theta$ denotes the set of all parameters in the main network of CARD, and $\lambda:[0,T] \rightarrow (0,\infty)$ denotes the weight function. 

We consider the time step $T$ takes a sufficiently large size.   Generally we have $q(\mathbf{y}_{T}|f_{\phi}(\mathbf{x})) \approx p(\mathbf{y}_{T}|f_{\phi}(\mathbf{x}))$, then the 2-Wasserstein distance at time $T$ is closely approaches zero, i.e. $W_{2}(q(\mathbf{y}_{T}|f_{\phi}(\mathbf{x})),p(\mathbf{y}_{T}|f_{\phi}(\mathbf{x}))) \rightarrow 0$. 
Meanwhile, a larger size of $T$ provides ample time for network training, thereby enabling the loss function $L_{1}(\phi,\theta,\lambda)$ to reach its minimum. This implies that  $H(t) \rightarrow 0$ in an ideal case.
In summary, the right-hand side of the inequality in Eq.(\ref{F10}) will gradually approach zero when time step $T$ at a sufficiently large size. And since the distance is non-negative, the 2-Wasserstein distance between the original and the generated distribution also tends towards zero, i.e. $W_{2}(q(\mathbf{y}_{0}|f_{\phi}(\mathbf{x})),p(\mathbf{y}_{0}|f_{\phi}(\mathbf{x}))) \rightarrow 0$. 

This result firmly demonstrates that within the context of the present model framework, the model exhibits favorable convergence capabilities. It is capable of efficiently reducing the disparity between the original distribution and the generated distribution, thereby attaining the desired generation outcome. The numerical simulation results pertaining to Theorem \ref{the-1} have furnished compelling verification and substantiation for this theoretically derived conclusion. The details will be thoroughly presented in Section \ref{sec5}. This will enable us to more lucidly and intuitively comprehend and apprehend the practical verification of this crucial conclusion, as well as the underlying mathematical principles and practical implications.

Furthermore, if we set specific values to the weight function $\lambda(t)$, for instance, setting $\lambda(t) = \beta_{t}$, and suppose $\beta^{2}(t) M^{2}(t) \lambda^{-1}(t)$ is integrable, we can likewise derive an upper bound about the 2-Wasserstein distance between the two distributions. 
\begin{corollary}
If $\lambda(t) = \beta_{t},\forall t \in [0,T]$, then
\begin{equation}\label{F12}
\begin{aligned}
W_{2}(q(\mathbf{y}_{0}|f_{\phi}(\mathbf{x})),p(\mathbf{y}_{0}|f_{\phi}(\mathbf{x}))) & \leq \sqrt{2 \left( \int_{0}^{T} \beta_{t} M^{2}(t)dt \right)L_{1}(\phi,\theta,\lambda)}  + M(T)W_{2}(q(\mathbf{y}_{T}|f_{\phi}(\mathbf{x})),p(\mathbf{y}_{T}|f_{\phi}(\mathbf{x}))).
\end{aligned}
\end{equation}

\noindent If $\lambda(t)$ is any other value which satisfies the conditions, then 
\begin{equation}\label{F13}
\begin{aligned}
W_{2}(q(\mathbf{y}_{0}|f_{\phi}(\mathbf{x})),p(\mathbf{y}_{0}|f_{\phi}(\mathbf{x}))) & \leq \sqrt{2\left( \int^{T}_{0} \beta_{t}^{2} M^{2}(t) \lambda^{-1}(t)  dt \right) L_{1}(\phi,\theta,\lambda)} + M(T)W_{2}(q(\mathbf{y}_{T}|f_{\phi}(\mathbf{x})),p(\mathbf{y}_{T}|f_{\phi}(\mathbf{x}))).
\end{aligned}
\end{equation}
\label{cor-1}
\end{corollary}

According to Corollary \ref{cor-1}, we also consider the time step $T$ takes a sufficiently large size and we similarly have $q(\mathbf{y}_{T}|f_{\phi}(\mathbf{x})) \approx p(\mathbf{y}_{T}|f_{\phi}(\mathbf{x}))$, then the error between the endpoint of forward process and the starting-point of reverse process at time $T$ tends to zero, which means $W_{2}(q(\mathbf{y}_{T}|f_{\phi}(\mathbf{x})),p(\mathbf{y}_{T}|f_{\phi}(\mathbf{x}))) \rightarrow 0$.
As the loss function $L_{1}(\phi,\theta,\lambda)$ decreases and approaches 0, the 2-Wasserstein distance $W_{2}(q(\mathbf{y}_{0}|f_{\phi}(\mathbf{x})),p(\mathbf{y}_{0}|f_{\phi}(\mathbf{x})))$ closely approaches zero. The simulation experiments will confirm this convergence later.

However, there exists an unknown marginal distribution $q(\mathbf{y}_{t}|f_{\phi}(\mathbf{x}))$ at time $t$ in the loss function $L_{1}(\phi,\theta,\lambda)$. We approximate it using $q(\mathbf{y}_{t}|\mathbf{y}_{0},f_{\phi}(\mathbf{x}))$ similarly as DDPMs \citep{ho2020denoising, kwon2022score}, which is consistent with the Markov property of CARD. We redefine the loss function $L_{2}(\phi,\theta,\lambda)$ as
\begin{equation}\label{F14}
L_{2}(\phi,\theta,\lambda) = \frac{1}{2} \int^{T}_{0} \lambda_{t} \textup{E}_{q(\mathbf{y}_{t}|\mathbf{y}_{0},f_{\phi}(\mathbf{x}))q(\mathbf{y}_{0}|f_{\phi}(\mathbf{x}))} \left[ ||s_{\theta}(\mathbf{x},\mathbf{y}_{t},f_{\phi}(\mathbf{x}),t) - \frac{\partial \log q(\mathbf{y}_{t}|\mathbf{y}_{0},f_{\phi}(\mathbf{x}))} { \partial \mathbf{y}_{t}}  ||_{2}^{2} \right].
\end{equation}
In fact, we employ $L_{2}(\phi,\theta,\lambda)$ as the loss function in numerical simulations or real datasets, instead of the theoretically derived $L_{1}(\phi,\theta,\lambda)$. This particular aspect demands special attention to prevent the introduction of additional difficulties.

% it is difficult to accurately determine the form of the marginal distribution $q(\mathbf{x}_{t})$ at time $t$ in DDPMs \citep{ho2020denoising, han2022card}. Therefore we approximate 
% $q(\mathbf{x}_{t})$ using $q(\mathbf{x}_{t}|\mathbf{x}_{0})$, according to $q(\mathbf{x}_{t}) = \int q(\mathbf{x}_{t},\mathbf{x}_{0})d\mathbf{x}_{0}  = \int q(\mathbf{x}_{t}|\mathbf{x}_{0}) q(\mathbf{x}_{0}) d\mathbf{x}_{0}= E_{\mathbf{x}_{0}} [q(\mathbf{x}_{t}|\mathbf{x}_{0})]$. 
% The loss function is finally defined as $E_{\mathbf{x}_{0}} [q(\mathbf{x}_{t}|\mathbf{x}_{0}) ||\nabla \log q(\mathbf{x}_{t}|\mathbf{x}_{0})- s_{\theta}(\mathbf{x}_{t},t)||^{2}]$. 

% \subsection{Overview of Proof Idea}

\section{Upper Bound of Conditional Score Function}\label{sec4}
Within this section, our primary focus lies in the variation of the error between the conditional score function and its estimated value, specifically concentrating on the $H(t)$ term in Theorem \ref{the-1}. By introducing several novel assumptions regarding the original conditional distribution and the data range, we have meticulously derived a convergent upper bound for the conditional score function with the application of Taylor polynomials. The detailed proof is presented in Appendix \ref{appendixB}.

\subsection{New Technical Assumptions}
In generative models including diffusion models, the ideal situation we expect is to have a good estimation form for all the data. However, due to the presence of outliers or the unbalanced nature of the data distribution, it is highly likely to cause deviations in the estimation results in the actual operation. This also implies that it is necessary for us to make assumptions about the distribution range of the data, that is, we hope that our estimation results can be valid for the vast majority of data points, and for the few outliers, we can temporarily ignore the information they reflect in the overall data. 

This is also in line with the assumptions presented in Theorem \ref{the-1}. During the derivation of Theorem \ref{the-1}, we initially postulated that the two conditional distributions at time $t$, namely $q(\mathbf{y}_{t}|f_{\phi}(\mathbf{x})), p(\mathbf{y}_{t}|f_{\phi}(\mathbf{x}))$, would exhibit an exponential decay pattern within the $R^{d}$ space as $\|\mathbf{y}_{t}\|_{2}$ increases. With the aim of more precisely estimating the overall conditional distribution, we herein only impose similar assumptions on the original conditional distribution $q(\mathbf{y}_{0}|f_{\phi}(\mathbf{x}))$. Considering that the overall model adheres to a Gaussian distribution, it is pertinent to take into account the light-tailed nature of the exponential distribution. From this we propose the following related assumptions.

\begin{assumption}\label{assum6}
\textbf{(Light tail condition).}
Suppose that the original distribution $q(\mathbf{y}_{0}|f_{\phi}(\mathbf{x})) \in \mathscr{H}^{\beta}(R^{d} \times R^{d_{x}},B)$ for H$\ddot{o}$lder index $\beta >0$ and constant $B>0$. Meanwhile, there exist a constant $c>0$ such that for all $\mathbf{x} \in R^{d_{x}}$, the predicted conditional mean $\left\| f_{\phi}(\mathbf{x}) \right\|_{2} \leq c$, and then exist constants $c_{1},c_{2}>0$ such that for all $\mathbf{x} \in R^{d_{x}}$, we assume $q(\mathbf{y}_{0}|f_{\phi}(\mathbf{x})) \leq c_{1} \exp\left\{ -c_{2} \left\| \mathbf{y}_{0} \right\|^{2}_{2} /2 \right\}$.
\end{assumption}

\begin{assumption}\label{assum7}
\textbf{(Light tail predicted condition mean).} 
Suppose that there exist constants $c_{f_{1}},c_{f_{2}}>0$ such that for all $\mathbf{x} \in R^{d_{x}}$, $g( f_{\phi}(\mathbf{x})) = c_{f_{1}} \exp\left\{ -c_{f_{2}}\right\| f_{\phi}(\mathbf{x}) \left\|^{2}_{2} /2 \right\}$ holds.
% Suppose that there exist constants $c_{f_{1}},c_{f_{2}}>0$ such that for all $\mathbf{x} \in \mathbb{R}^{d_{x}}$, we construct a truncated distribution $g(\left\| f_{\phi}(\mathbf{x}) \right\|_{2}) = c_{f_{1}} \exp\left\{ -c_{f_{2}}\right\| f_{\phi}(\mathbf{x}) \left\|^{2}_{2} /2 \right\}$ when $0\leq\left\| f_{\phi}(\mathbf{x}) \right\|_{2}\leq c$, and $g(\left\| f_{\phi}(\mathbf{x}) \right\|_{2}) = 0$ when $\left\| f_{\phi}(\mathbf{x}) \right\|_{2}>c$,
% where the normalizing factor denoted as
% $c_{f_{1}} = \int_{0}^{c} \exp\left\{ -c_{f_{2}}\right\| f_{\phi}(\mathbf{x}) \left\|^{2}_{2} /2 \right\} d\left\| f_{\phi}(\mathbf{x}) \right\|_{2}$.
\end{assumption}

\begin{assumption}
\label{assum8}
\textbf{(Data domain truncation).} 
For any given time $t$, we employ a sphere of radius $R$ to conduct a truncation operation on the domain of values of the input $\mathbf{y}_{t}$. To be more specific, we denote it as $D_{1} = \{ \mathbf{y}_{t}: || \mathbf{y}_{t}||_{\infty} \leq R ,\ \forall t \in [0,T], \ \exists R \in B_{\infty}(R)\}$ where $B_{\infty}(R)$ represents the $l_{\infty}$ ball with given radius R. On the complementary region of the set $D_{1}$, written as $\overline{D}_{1}$, we set the corresponding fraction approximation to be uniformly bounded by the constants associated with $R,t,f_{\phi}(\mathbf{x})$.
\end{assumption}

\begin{assumption}\label{assum9}
\textbf{(Conditional density truncation).} 
We establish a threshold for $q(\mathbf{y}_{t}|f_{\phi}(\mathbf{x}))$ at any time $t$ and define $D_2 = \{ \mathbf{y}_{t}:q(\mathbf{y}_{t}|f_{\phi}(\mathbf{x})) \geq \epsilon_{low}, \ \forall t \in [0,T]\}$. Similar to Assumption \ref{assum8}, we additionally set our approximation to be constrained by the constant on the complement of $D_2$, written as $\overline{D}_{2}$.
\end{assumption}

Assumption \ref{assum6} delineates the assumption we make regarding the distribution of the original data to enhance the accuracy of our estimates. First, suppose that the original condition distribution is smooth, so it is easy to estimate. 
Equivalently, let $\beta = s + \gamma, s=[\beta], \gamma \in [0,1)$. The H$\ddot{o}$lder norm of the function $q(\mathbf{y}_{0}|f_{\phi}(\mathbf{x}))$ is defined as 
\begin{equation*}
\| q(\mathbf{y}_{0}|f_{\phi}(\mathbf{x})) \|_{\mathscr{H}^{\beta}(\mathbb{R}^{d} \times \mathbb{R}^{d_{x}},B)}
= \max\limits_{\mathbf{t}:\|\mathbf{t}\|_{1} < s} \sup\limits_{\mathbf{y}_{0}} \partial^{\mathbf{t}} q(\mathbf{y}_{0}|f_{\phi}(\mathbf{x}))
+ \max\limits_{\mathbf{t}:\|\mathbf{t}\|_{1} + s} \sup\limits_{\mathbf{y}_{0} \neq \mathbf{z}} \frac{|\partial^{\mathbf{t}} q(\mathbf{y}_{0}|f_{\phi}(\mathbf{x})) - \partial^{\mathbf{t}} q(\mathbf{z}|f_{\phi}(\mathbf{x}))|}{\|\mathbf{y}_{0}-\mathbf{z}  \|^{\gamma}_{\infty}}.
\end{equation*}
The function $q$ is $\beta$ H$\ddot{o}$lder continuous if and only if $\|q\|_{\mathscr{H}^{\beta}(\mathbb{R}^{d} \times \mathbb{R}^{d_{x}},B)} < \infty$.
Second, since the original conditional distribution is Gaussian, it can be assumed that it has a light tail. 

For simplicity, we use the relu function in the neural network architecture to simplify the calculation to a certain extent, making the model develop in the direction of a linear model \citep{gao2024convergence1, fu2024unveil, jiao2024model}.

% Assumption 6 truncated distribution complent
% When the original input $\mathbf{x}$ is fixed, the predicted conditional mean $f_{\phi}(\mathbf{x})$ is deterministic and lacks the traditional probability distribution in Assumption \ref{assum7}. Thus, we devised a truncated distribution $g(\left\| f_{\phi}(\mathbf{x}) \right\|_{2})$ to mimic light-tailed characteristics. We chose a refined approach without using generalized functions like the Dirac delta function.
% Based on Assumption \ref{assum6} $\left\| f_{\phi}(\mathbf{x}) \right\|_{2} \leq c$, the exponential function is used to make $g(\left\| f_{\phi}(\mathbf{x}) \right\|_{2})$ show an exponential decay similar to a light-tailed distribution within $[0, c]$, and equals to 0 when $\left\| f_{\phi}(\mathbf{x}) \right\|_{2}>c$. Technically, by adjusting $c_{f_{2}}$, we can control the decay rate to simulate different light-tailed features.
% In summary, we use this truncated distribution to describe non-random variables for easier analysis and to utilize probability distribution properties, though it's not a true probability distribution. Notably, in subsequent proofs, its well-defined and limited range simplifies computations, supporting the research. 

Integrating the constraints on the predicted conditional mean in Assumption \ref{assum6} and Assumption \ref{assum7}, the constant c defines a clear boundary for $f_{\phi}(\mathbf{x})$. Within this range, the function $g( f_{\phi}(\mathbf{x}))$  maintains the characteristic of exponential decay as $\left\| f_{\phi}(\mathbf{x}) \right\|_{2}$ increases.
Intuitively, in a light-tailed distribution, the probability mass decreases rapidly when moving away from a central region (analogous to the region where $\left\| f_{\phi}(\mathbf{x}) \right\|_{2}$ is relatively small). The existing constant $c$ ensures that there is no long tail, as all values are limited to the interval $[0, c]$, strengthening the impression of light-tailed.
Unlike the original distribution assumption, which exhibit light-tailed through natural decay, the truncation by setting $c$ here is based on practical requirements, making the characterization of $g(f_{\phi}(\mathbf{x}))$ more in line with reality and facilitating subsequent analysis. 
In the subsequent proof, for computing statistical measures like the expectation of $g(f_{\phi}(\mathbf{x}))$, the integration range narrows to $[0, c]$, simplifying and clarifying the calculation. This precisely captures the average behavior of $[0, c]$ within the bounded range, aligning with the emphasis on its reasonable value range's statistical traits in practical applications.

Based on the assumption of light tail, we focus on specific regions within the original data set. Assumption \ref{assum8} shows that we also consider the range of the variable $\mathbf{y}_{t}$ at any $t$. At the beginning, $\mathbf{y}_{t} \in \mathbb{R}^{d}$, which means the range of values of $\mathbf{y}_{t}$ may be unbounded. However, for different data, we cannot guarantee that the variable $\mathbf{y}_{t}$ is always accurately estimated, which poses difficulties in approximating the conditional density $q(\mathbf{y}_{t}|f_{\phi}(\mathbf{x}))$. Thus, we need to truncate the domain of the variable $\mathbf{y}_{t}$ to make the existence of data within this range meaningful, thereby facilitating the estimation of its conditional density.

The conditional score function can be converted into $\nabla \log q(\mathbf{y}_{t}|f_{\phi}(\mathbf{x})) = \nabla q(\mathbf{y}_{t}|f_{\phi}(\mathbf{x})) / q(\mathbf{y}_{t}|f_{\phi}(\mathbf{x}))$. However, through an analysis of its mathematical expression, a significant problem emerges. If the denominator $q(\mathbf{y}_{t}|f_{\phi}(\mathbf{x}))$ is arbitrarily small, then $1 / q(\mathbf{y}_{t}|f_{\phi}(\mathbf{x}))$ becomes arbitrarily large, and the conditional score function does not yield a good estimate. Therefore, Assumption \ref{assum9} indicates that we need locally truncate the conditional distribution at any $t$ of both forward and reverse processes. That is, we introduces a threshold parameter $\epsilon_{low}>0$ such that $q(\mathbf{y}_{t}|f_{\phi}(\mathbf{x})) \geq \epsilon_{low}$ holds. 

Given the assumptions, it is crucial to explore their significance. Each one is essential in the research, underpinning our later analysis.

\subsection{Convergence guarantee}
The following theorem presents the approximation theory to approximate the conditional score function. The detailed proof shows in Appendix \ref{appendixB}.
\begin{theorem}\label{the-3}
Suppose the above Assumptions hold. For sufficiently small $\varepsilon \in (0,1/e)$, any integer $N>0, t \in [0,T]$ and $R_{*}=\max (2R,2R_{f})$ for constants $B>0,\beta>0,R,R_{f}>1$, it holds that
\begin{equation}\label{F15}
\begin{aligned}
& \int_{R^{d}} \left\| s_{\theta}(\mathbf{y}_{t},f_{\phi}(\mathbf{x}),t)-\nabla \log q(\mathbf{y}_{t}|f_{\phi}(\mathbf{x})) \right\| ^{2}_{2} \ q(\mathbf{y}_{t}|f_{\phi}(\mathbf{x})) \ d \mathbf{y}_{t}
% \\& \lesssim O\left( \frac{N + 1}{\sigma_{t}^{d}}(1+\gamma_{t}^{-d} ) + N^{-\beta} \left[(\log\varepsilon^{-1})^{\frac{1}{2}} +1 \right] \left[\gamma_{t}^{d/2} + \frac{(N^{\beta}+1) N^{-d}}{\sigma_{t}^{d}} \right]  \right).
\lesssim O\left( \left[(\log\varepsilon^{-1})^{\frac{1}{2}} +1 \right] \left[\frac{N^{-\beta}}{\gamma_{t}^{d/2} } + \frac{N^{-d}}{\sigma_{t}^{d}} + \frac{N^{-d-\beta}}{\sigma_{t}^{d}} \right]  \right). 
\end{aligned}
\end{equation}
\label{the-2}
\end{theorem}

Above Eq.(\ref{F15}) measures the integral value of $\mathbf{y}_{t}$ in the $R^{d}$ space with respect to the square of the norm of the difference between the conditional score function and its estimator, and gives an upper bound estimate of this integral value and the exact upper bound form is shown at the end of Appendix \ref{appendixB}. 

We define the parameter $N$ in Appendix \ref{appendixB43} as the number of divisions in each dimension, which means the length of each sub interval is $1/N$. Similar to \cite{fu2024unveil}, it can be used as a measure of network size.
The given conditions $\gamma_{t} = e^{-\frac{\beta}{2} t},\  \sigma^{2}_{t} = 1-e^{- \beta t}$, and we find that as $t$ increases, especially in the limit case when $t \to \infty$, $\sigma_{t} \to 1$ and $\gamma_{t} \to 0$. It is worth noting that the above result is based on the large value of time $t$, due to \cite{song2020improved, vahdat2021score, fu2024unveil} have demonstrated that the conditional score tends to infinity when $t$ approaches zero. the time t enters the conditional score function through the ratio $\gamma_{t}$ and the variance $\sigma_{t}$ of the Gaussian noise added to the clean data distribution. Both $\gamma_{t}$ and $\sigma_{t}$ are super smooth (infinitely differentiable) and therefore, very easy to approximate using neural networks.
For a fixed network size $N$, the approximation error scales as $N^{-d-\beta}$ indicating a faster approximation when the original distribution has a higher order of smoothness and the approximation error increases as time t decreases, so we consider the error upper bound when $t$ is large or when it exceeds a certain value.

\subsection{Overview of Proof Idea}
Now we have a preliminary understanding of the idea of proof in this section. First and foremost, it is essential to clarify our motivation. In practical scenarios, the specific value of the score function cannot be precisely accessed, thereby mandating the use of neural networks for its estimation purposes. To regulate the error between the value estimated by the network and the score function being estimated, a step by step analysis is conducted under the given assumptions.

Based on the existing assumptions, we ultimately derive Theorem \ref{the-3} and elaborate on the principal idea for its proof. The key steps entail the appropriate truncation of the data value range and the conditional density function. Moreover, Taylor polynomials are utilized to approximate the score function.

To begin with, we reformulate the conditional score function as well as a logarithmic gradient term as $\nabla \log q(\mathbf{y}_{t}|f_{\phi}(\mathbf{x})) = \nabla q(\mathbf{y}_{t}|f_{\phi}(\mathbf{x})) / q(\mathbf{y}_{t}|f_{\phi}(\mathbf{x}))$.
Based on the form of the fraction, we can approximate the numerator and denominator separately for the final result. Furthermore, the numerator and denominator are theoretically of similar structures. Hence, we concentrate on approximating $q(\mathbf{y}_{t}|f_{\phi}(\mathbf{x}))$, and the approximate result of its gradient term can be derived accordingly.

From the definition of CARD and the SDE form of its forward process, we write the integral definition of conditional density function $q(\mathbf{y}_{t}|f_{\phi}(\mathbf{x}))$ as
\begin{equation*}
\begin{aligned}
    q(\mathbf{y}_{t}|f_{\phi}(\mathbf{x})) 
    % & = \int_{\mathbb{R}^{d}} q(\mathbf{y}_{t},\mathbf{y}_{0}|f_{\phi}(\mathbf{x}))  d \mathbf{y}_{0}
    % = \int_{\mathbb{R}^{d}} q(\mathbf{y}_{t}|\mathbf{y}_{0},f_{\phi}(\mathbf{x})) q(\mathbf{y}_{0}|f_{\phi}(\mathbf{x})) d \mathbf{y}_{0}\\
    & = \frac{1}{\sigma_{t}^{d} (2 \pi)^{\frac{d}{2}}} \int_{\mathbb{R}^{d}}  
    q(\mathbf{y}_{0}|f_{\phi}(\mathbf{x}))      
    \exp {\left\{-\frac{\left\| \mathbf{y}_{t}-\left[ \gamma_{t} \mathbf{y}_{0} + \left(1-\gamma_{t} \right) f_{\phi}(\mathbf{x}) \right] \right\|^{2}}{2 \sigma^{2}} \right\}} d \mathbf{y}_{0}.
\end{aligned}
\end{equation*}

We pay attention to the integrand constituted by the original conditional density function $q(\mathbf{y}_{0}|f_{\phi}(\mathbf{x}))$ and the Gaussian kernel function $\exp {\left\{-\left\| \mathbf{y}_{t}-\left[ \gamma_{t} \mathbf{y}_{0} + \left(1-\gamma_{t} \right) f_{\phi}(\mathbf{x}) \right] \right\|^{2} / 2 \sigma^{2} \right\}}$. In Assumption \ref{assum6}, it is hypothesized that the original conditional density possesses smoothness property. Currently, when estimating the conditional density at any moment during the diffusion process, a natural thought is to conduct Taylor expansion polynomials for the two terms within the integrand respectively for the purpose of approximation.

%exist two challanges
During actual operation, we find there exist two issues to the fore. Firstly, the unbounded nature of the data domain poses a challenge when uniformly approximating $q(\mathbf{y}_{t}|f_{\phi}(\mathbf{x}))$. Secondly, although the Taylor polynomials can be implemented via a neural network, the integration process therein is extremely arduous. 
% Moreover, from the above integral terms, it is clear that the exponential function depends on the time step $t$, causing trouble for the approximation of Taylor polynomials.

%first challange
For the first issue, we consider the truncation strategy. Based on the concepts in Assumption \ref{assum8} and Assumption \ref{assum9}, in order to ensure a consistent approximation result, validate this result for the vast majority of data points, and avoid the aforementioned fraction from being meaningless, it is necessary to truncate the original data $\mathbf{y}_{0}$ and the conditional density function $q(\mathbf{y}_{t}|f_{\phi}(\mathbf{x}))$ at each step respectively. Meanwhile, make sure that the function error outside the truncated region can be controlled.

%Why use Taylor expansion polynomials to approximate
For the second issue, we use Taylor polynomials to approximate the integrand terms respectively \citep{fu2024unveil, oko2023diffusion}. Regarding the integrand with exponential terms and the function that complies with a specific Gaussian distribution, we can take advantage of the characteristics of both to separately approximate them using Taylor polynomials. In this way, the original function can be approximated as the product of two polynomials, thereby simplifying the calculation of such complex integrals. Simultaneously, the term concerning the time step $t$ exerts an influence on the entire integral through the mean and variance of the Gaussian distribution within the diffusion process. This term is characterized by smoothness, thereby facilitating the approximation process. 

Intuitively, the Taylor polynomial is a typical example in the field of function approximation theory, which approximates a function by means of polynomials. For a function $g(x)$ with $n + 1$ derivatives, it can be expressed as $g(x)=\sum_{k = 0}^{n}\frac{g^{(k)}(a)}{k!}(x - a)^{k}+R_{n}(x)$. When the remainder error $R_{n}(x)$ can be precisely regulated within an acceptable range, the polynomial consisting of the first $n$ terms can provide a relatively accurate approximation of the original function. Meanwhile, higher order terms in Taylor expansion can describe the local characteristics of the function more precisely. During the integration process, the integrand within the integration interval can be regarded as a collection of multiple local approximations. The Taylor polynomial is used to approximate each local segment to estimate the overall integral.

\section{Experiments}\label{sec5}
In order to fully evaluate and analyze the performance of CARD, we conduct simulation experiments on several two-dimensional datasets to verify the conclusions of Theorem \ref{the-1} and Corollary \ref{cor-1} of CARD. See Appendix \ref{appendixD} for all simulation results.\\

\begin{figure}[ht]
  \centering
  \subfloat[Moon]
  {
    \label{fig111:subfig1}\includegraphics[width=0.23\textwidth]{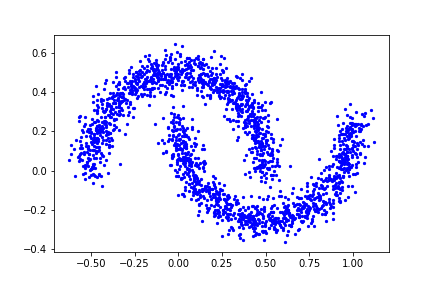}
  }
  \subfloat[Circle]
  {
    \label{fig111:subfig2}\includegraphics[width=0.23\textwidth]{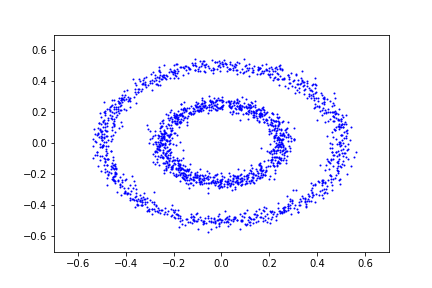}
  }
  \subfloat[$N(\mathbf{0},0.1\mathbf{I})$]
  {
    \label{fig111:subfig3}\includegraphics[width=0.23\textwidth]{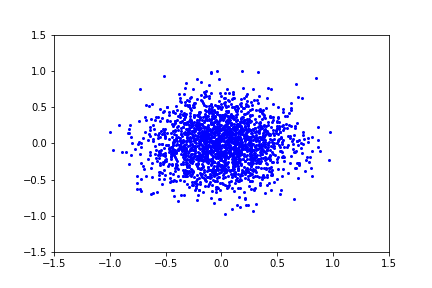}
  }
  \subfloat[$N ((\pm0.5, \pm0.5)^{T}, 0.01\mathbf{I})$]
  {
    \label{fig111:subfig4}\includegraphics[width=0.23\textwidth]{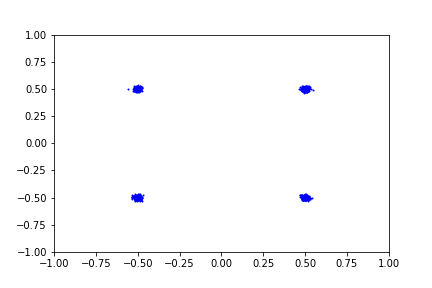}
  }
  \caption{\textbf{Sample distribution of four datasets.} The four subfigures exhibit the sample distributions of various datasets. The leftmost subfigure portrays the Moon dataset, modeling a quintessential nonlinear binary classification challenge. The second subfigure introduces the Circle dataset, where two concentric circles represent two distinct classes and are inherently nonlinearly separable in the feature space. The third subplot displays the zero-mean Gaussian dataset, where all data points are tightly concentrated around the origin, forming a single class. The rightmost subplot depicts a four-class Gaussian mixture dataset,with each class having equal weighting. Data points are generated in four distinct regions, defining independent classes, and ensuring equal probability in sampling from each distribution.}
  \label{fig111}
\end{figure}

We utilized four classic and widely used simulated datasets: the Moon dataset, the Circle dataset, a Gaussian distribution dataset generated with a mean of zero and a covariance matrix scaled by 0.1 times the identity matrix $N(\mathbf{0},0.1\mathbf{I})$, and a Gaussian mixture distribution dataset with equal weights, consisting of two Gaussians centered at $(\pm0.5, \pm0.5)^{T}$ and a covariance matrix scaled by 0.01 times the identity matrix $N((\pm0.5, \pm0.5)^{T}, 0.01\mathbf{I})$. The Moon and Circle datasets were provided by the scikit-learn (sklearn) library in Python. These four datasets represent data distributions with varying degrees of complexity and features, providing an ideal environment for testing and validating the conclusions of our models. The distribution of samples generated for each dataset is shown in Figure \ref{fig111}. 

During the simulation experiments with CARD, we commenced by defining the pre-trained network $f_{\phi}(\mathbf{x})$ as a three-layer fully connected architecture. Specifically, the first layer houses 128 neurons, while the second layer contains 64 neurons. To obtain the output of the final layer, we employed the sigmoid activation function for all layers except the last one. We configured the network with a learning rate of $10^{-4}$ and a batch size of 128 for each training iteration. The mean squared error (MSE) served as the loss function throughout the training, and we leveraged the Adam algorithm for model optimization.
Furthermore, we set the diffusion step size at 20, during which the noise variance $\beta_{t}$ undergoes a sigmoid-shaped growth, gradually increasing from $10^{-5}$ to $10^{-2}$. Additionally, we introduced a four-layer fully connected linear noise prediction network, with each layer comprising 64 neurons. Except for the final layer, we utilized the softplus activation function in all other layers to enhance the network's nonlinear capabilities. \\

To visualize the difference between the distribution of the given and generated data, we take the logarithm of both sides of Eq. (\ref{F12}):
\begin{equation}\label{F16}
\log W_{2}(q(\mathbf{y}_{0}|f_{\phi}(\mathbf{x})),p(\mathbf{y}_{0}|f_{\phi}(\mathbf{x}))) \leq \frac{1}{2} \log \left( 2\int_{0}^{T} \beta_{t} M^{2}(t) dt  \right) + \frac{1}{2} \log L_{1}(\phi,\theta,\lambda).
\end{equation} 
Here, Eq.(\ref{F16}) represents the correlation between two logarithmic variables. During the training of CARD, the integral component $\int_{0}^{T} \beta_{t} M^{2}(t) dt$ remains constant, whereas the loss function $L_{1}(\phi,\theta,\lambda)$ evolves as the training progresses. \\

\begin{figure}[ht]
  \centering
  \subfloat[Moon]
  {
  \label{fig112:subfig1}\includegraphics[width=0.23\textwidth]{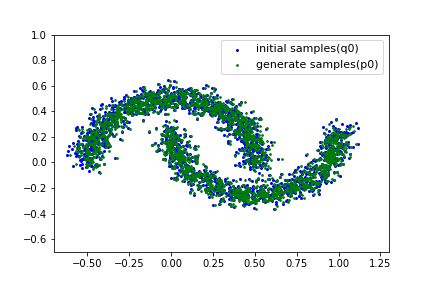}
  }
  \subfloat[Circle]
  {
    \label{fig112:subfig2}\includegraphics[width=0.23\textwidth]{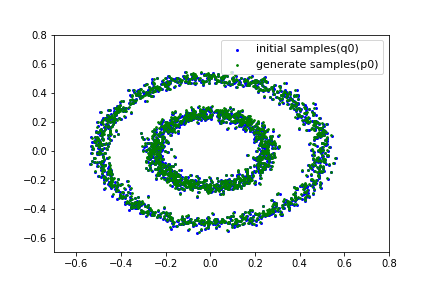}
  }
  \subfloat[$N(\mathbf{0},0.1\mathbf{I})$]
  {
    \label{fig112:subfig3}\includegraphics[width=0.23\textwidth]{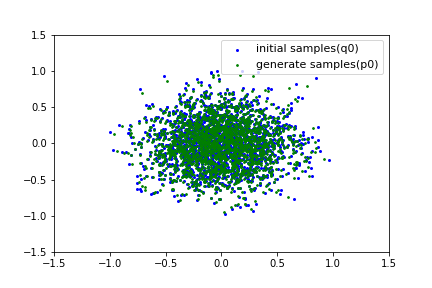}
  }
  \subfloat[$N ((\pm0.5, \pm0.5)^{T}, 0.01\mathbf{I})$]
  {
    \label{fig112:subfig4}\includegraphics[width=0.23\textwidth]{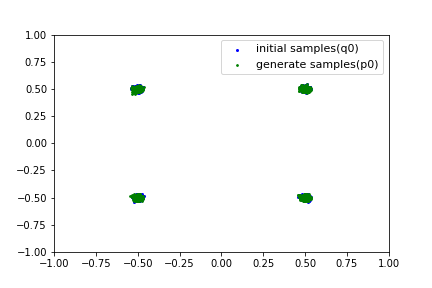}
  }
  \caption{\textbf{Generating sample distributions of four datasets.} In each subfigure, the blue sample points represent the original sample distribution, and the green sample points represent the corresponding sample distribution generated by CARD.}%The subfigures from left to right represent Moon, Circle, zero-mean Gaussian, four-class Gaussian datasets in turn.
  \label{fig112}
\end{figure}

\begin{figure}
  \centering
  %\subfloat[Moon]
  {
      \label{fig113:subfig1}\includegraphics[width=0.23\textwidth]{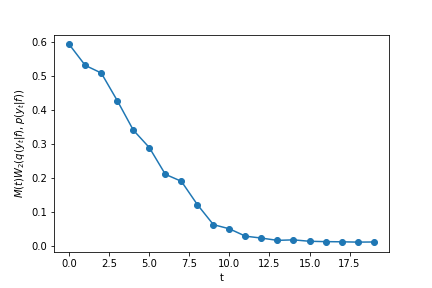}
  }
  %\subfloat[Circle]
  {
      \label{fig113:subfig2}\includegraphics[width=0.23\textwidth]{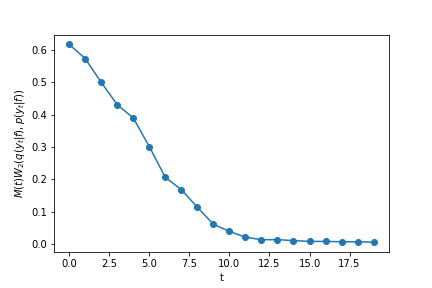}
  }
  %\subfloat[$N(\mathbf{0},0.1\mathbf{I})$]
  {
      \label{fig113:subfig3}\includegraphics[width=0.23\textwidth]{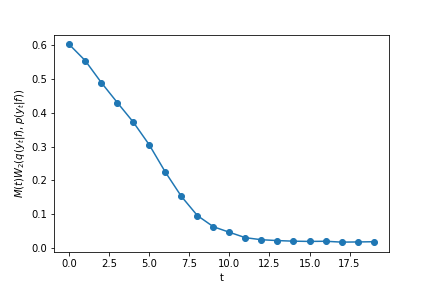}
  }
  %\subfloat[$N ((\pm0.5, \pm0.5)^{T}, 0.01\mathbf{I})$]
  {
      \label{fig113:subfig4}\includegraphics[width=0.23\textwidth]{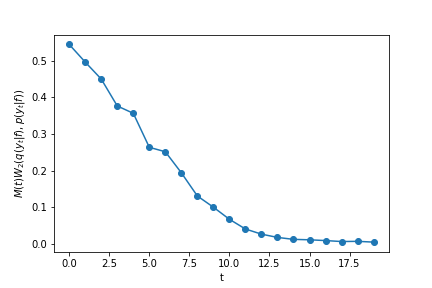}
  }
  \caption{\textbf{Product term of Wasserstein distances for four datasets.} In each subfigure, the horizontal axis represents the diffusion time step, and the vertical axis represents the change in variable $M(t)W_{2}(q(\mathbf{y}_{t}|f_{\phi}(\mathbf{x})),p(\mathbf{y}_{t}|f_{\phi}(\mathbf{x})))$.}
  \label{fig113}
\end{figure}

Clearly, we need to investigate whether the second-order Wasserstein distance between the given and generated data distributions adheres to the upper bound specified in Eq.(\ref{F12}). We conduct simulation experiments on above four datasets. Firstlt, we compare the data generated by CARD with the given data, as exemplified in Figure \ref{fig112}. Visually, the sample points generated under each dataset exhibit minimal deviation from the given data. 
Secondly, in Eq.(\ref{F12}), we assume that two distribution are consistent at $T$ moment, implying $q(\mathbf{y}_{T}|f_{\phi}(\mathbf{x}))=p(\mathbf{y}_{T}|f_{\phi}(\mathbf{x}))$. Here, it has been initially observed in Figure \ref{fig112} that there is a small difference between the generated and the given data distribution. To accurately verify this hypothesis, the difference between the two distributions at each step $t$ is numerically reflected as Figure \ref{fig113}. 
It can be seen that as the diffusion step $t$ increases, the variable $M(t)W_{2}(q(\mathbf{y}_{t}|f_{\phi}(\mathbf{x})),p(\mathbf{y}_{t}|f_{\phi}(\mathbf{x})))$ decreases and eventually tends to zero, indicating that $M(T)W_{2}(q(\mathbf{y}_{T}|f_{\phi}(\mathbf{x})),p(\mathbf{y}_{T}|f_{\phi}(\mathbf{x}))) \rightarrow 0$ at moment $T$. The correctness of the hypothesis is verified.\\

\begin{figure}[H]
  \centering
  \subfloat[Moon]
  {
      \label{fig114:subfig1}\includegraphics[width=0.23\textwidth]{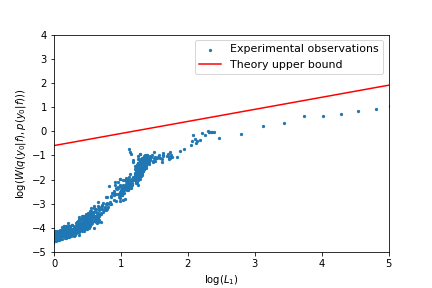}
  }
  \subfloat[Circle]
  {
      \label{fig114:subfig2}\includegraphics[width=0.23\textwidth]{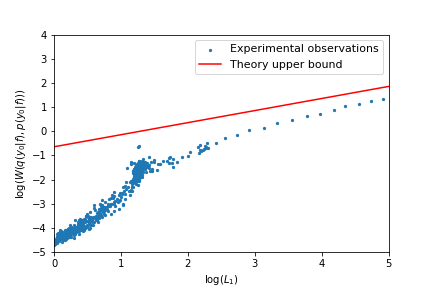}
  }
  \subfloat[$N(\mathbf{0},0.1\mathbf{I})$]
  {
      \label{fig114:subfig3}\includegraphics[width=0.23\textwidth]{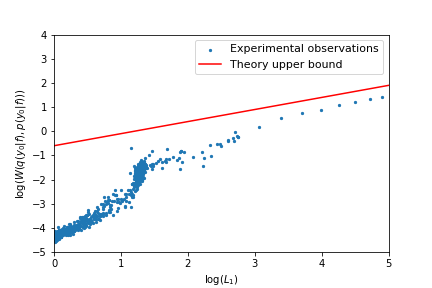}
  }
  \subfloat[$N ((\pm0.5, \pm0.5)^{T}, 0.01\mathbf{I})$]
  {
      \label{fig114:subfig4}\includegraphics[width=0.23\textwidth]{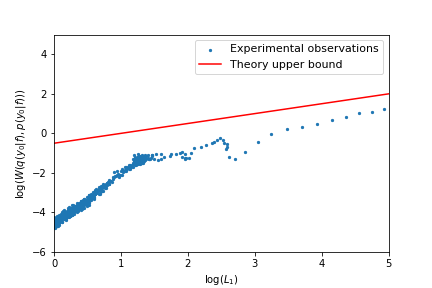}
  }
  \caption{\textbf{Log-plots about Wasserstein distances and loss function of four datasets.} The horizontal axis represents the logarithmic form of the Wasserstein distance between the given sample and the final generated sample, denoted by $\log W_{2}(q(\mathbf{y}_{0}|f_{\phi}(\mathbf{x})),p(\mathbf{y}_{0}|f_{\phi}(\mathbf{x})))$. The vertical axis represents the logarithm of the loss function during training, denoted by $\log L_{1}(\phi,\theta,\lambda)$. The red line represents the theoretical upper bound and the blue dot represents the Wasserstein distance of the samples generated by CARD.}%The subfigures from left to right represent Moon, Circle, zero-mean Gaussian, four-class Gaussian datasets in turn.
  \label{fig114}
\end{figure}

At last, we simulate the theoretical upper bound for the second-order Wasserstein distance in Eq.(\ref{F16}) and concurrently compute the second-order Wasserstein distance between the given data distribution of each dataset and the generated data distribution by CARD during training. Subsequently, using Eq.(\ref{F16}), we plot the relationship between $\log W_{2}(q(\mathbf{y}_{0}|f_{\phi}(\mathbf{x})),p(\mathbf{y}_{0}|f_{\phi}(\mathbf{x})))$ and $\log L_{1}(\phi,\theta,\lambda)$. The specific simulation results are presented in Figure \ref{fig114}. It is evident that regardless of the dataset, the theoretical upper bound (red line) consistently exceeds the observed Wasserstein distance (blue dots), thereby validating the effectiveness of this theoretical upper bound. Furthermore, as the loss function  $L_{1}(\phi,\theta,\lambda)$ decreases, the gap between the theoretical upper bound and the actual values gradually decreases.

\section*{Conclusions}
In the present paper, we have devised a state-of-the-art non-asymptotic theoretical convergence regime specific to conditional diffusion models, thereby inaugurating the theoretical exploration of the diffusion model for classification and regression (CARD). Grounded on the presumption that the drift term and score function within the stochastic differential equation exhibit smoothness, through the utilization of the second-order Wasserstein distance in the domain of optimal transport theory, we have conducted an exhaustive and profound analysis of the error dynamics between the original distribution and the generated distribution, culminating in the derivation of its convergence upper bound.
Thereafter, we proceeded to further relax the underlying assumptions and undertook an elaborate examination of the discrepancies between the score function and its network estimations, yielding a more intricate error upper bound that imposes relatively parsimonious assumptions on the original data distribution. The analytical scaffold proposed herein is likely to offer valuable impetus to the investigation of other variants of score-based conditional models.

% Although the theoretical properties of CARD are thoroughly investigated in this paper, there are still many issues that need to be further explored. 
Looking forward, a multitude of issues are in urgent need of profound theoretical elucidation. For instance, how the CARD model combined with other technologies can play a synergistic effect, and is it feasible for us to curtail the assumptions, such that rapid convergence outcomes in terms of the Wasserstein distance can be attained under the context of a general original data distribution? Alternatively, can other measurement indices be employed to gauge the disparities between two distributions? The pivotal aspect resides in the potential for further expansion of the conditional function embedded within this model. Additionally, what transformations will be manifested in the diffusion model under the non-conditional mean scenario? And how can we expedite the sampling process of this conditional model in practical applications while concurrently guaranteeing its convergence attributes?

\section*{Acknowledgments}

\bibliographystyle{unsrtnat} %alpha, apalike, ieeetr,unsrtnat
\bibliography{bibliography}

\appendix
\newpage
\section{Proof details of error bounds between distributions }
\subsection{Proof of the SDEs form of CARD}\label{appendixA1}
We prove the stochastic differential equation forms for the forward and backward processes in CARD.
First, we derive the SDEs form of the CARD forward process, which is Eq. (\ref{E1}). It's already in CARD \citep{han2022card}
\begin{equation*}
q\left(\mathbf{y}_{t} \mid \mathbf{y}_{t-1}, f_{\phi}(\mathbf{x})\right) = N\left( \sqrt{1-\beta_{t}} \mathbf{y}_{t-1}+\left(1-\sqrt{1-\beta_{t}}\right) f_{\phi}(\mathbf{x}), \beta_{t} \mathbf{I}\right).
\end{equation*}
If we sample from this distribution, we can obtain
\begin{equation}\label{I1}
\mathbf{y}_{t} = \sqrt{1-\beta_{t}} \mathbf{y}_{t-1}+\left(1-\sqrt{1-\beta_{t}}\right) f_{\phi}(\mathbf{x}) + \sqrt{\beta_{t}} \varepsilon_{t-1},\ t=1,2,\cdots,T.
\end{equation}
Now we define a new noise variance sequence such that time $t$ is satisfied with $t \in (0,1), t \in \left\{ 0,\frac{1}{T},\cdots,\frac{T-1}{T} \right\}$ and the new noise variance sequence is $\left\{ \beta(\frac{t}{T})=T\beta_{t}  \right\}|_{t=1}^{T}$. At the same time, since the time $t$ has been changed, the corresponding term with the time $t$ corner label has to be changed, that is, there
\begin{equation*}
\beta(\frac{t}{T})/T=\beta_{t},\ \mathbf{y}(\frac{t}{T})=\mathbf{y}_{t},\ \varepsilon(\frac{t}{T})=\varepsilon_{t},\ \Delta t =\frac{1}{T}.
\end{equation*}
Therefore, Eq.(\ref{I1}) becomes
\begin{equation*}
\mathbf{y}(\frac{t}{T}) = \sqrt{1-\beta(\frac{t}{T})/T} \ \mathbf{y}(\frac{t-1}{T}) + \left(1-\sqrt{1-\beta(\frac{t}{T})/T}\right) f_{\phi}(\mathbf{x}) + \sqrt{\beta(\frac{t}{T})/T} \ \varepsilon(\frac{t-1}{T}),
\end{equation*}
If $\Delta t$ is used to denote the change between two time steps, the above equation becomes equivalently
\begin{equation}\label{I2}
\mathbf{y}(t+\Delta t) = \sqrt{1-\beta(t+\Delta t)\Delta t} \ \mathbf{y}(t) + \left(1-\sqrt{1-\beta(t+\Delta t)\Delta t}\right) f_{\phi}(\mathbf{x}) + \sqrt{\beta(t+\Delta t)\Delta t} \ \varepsilon(t),
\end{equation}
To simplify Eq.(\ref{I2}), the first two terms on the right side of the equal sign are formulated separately
\begin{equation*}
\begin{aligned}
& \sqrt{1-\beta(t+\Delta t)\Delta t} \ \mathbf{y}(t) = \sqrt{\left(1-\frac{1}{2}\beta(t+\Delta t)\Delta t \right)^{2}-\frac{1}{4}\beta^{2}(t+\Delta t)\Delta t} \ \mathbf{y}(t),\\
& \left(1-\sqrt{1-\beta(t+\Delta t)\Delta t}\right) f_{\phi}(\mathbf{x}) = \left(1- \sqrt{\left(1-\frac{1}{2}\beta(t+\Delta t)\Delta t\right)^{2}-\frac{1}{4}\beta^{2}(t+\Delta t)\Delta t} \right) f_{\phi}(\mathbf{x}),
\end{aligned}
\end{equation*}
When $\Delta t \rightarrow 0$, reduces to
% \begin{equation*}
% \begin{aligned}
% & \sqrt{1-\beta(t+\Delta t)\Delta t} \ \mathbf{y}(t) \approx \left(1-\frac{1}{2}\beta(t+\Delta t)\Delta t \right)\ \mathbf{y}(t),\\
% & \left(1-\sqrt{1-\beta(t+\Delta t)\Delta t}\right) f_{\phi}(\mathbf{x}) \approx \frac{1}{2}\beta(t+\Delta t)\Delta t  f_{\phi}(\mathbf{x}).
% \end{aligned}
% \end{equation*}
% Therefore, Eq.(\ref{I2}) becomes
% \begin{equation*}
% \mathbf{y}(t+\Delta t) \approx \left(1-\frac{1}{2}\beta(t+\Delta t)\Delta t \right)\ \mathbf{y}(t) + \frac{1}{2}\beta(t+\Delta t)\Delta t  f_{\phi}(\mathbf{x}) + \sqrt{\beta(t+\Delta t)\Delta t} \ \varepsilon(t),
% \end{equation*}
% then we take $\Delta t \rightarrow 0$,
\begin{equation*}
\begin{aligned}
\mathbf{y}(t+\Delta t) & \approx \left(1-\frac{1}{2}\beta(t)\Delta t \right)\ \mathbf{y}(t) + \frac{1}{2}\beta(t)\Delta t  f_{\phi}(\mathbf{x}) + \sqrt{\beta(t)\Delta t} \ \varepsilon(t),\\
& \approx \mathbf{y}(t)-\frac{1}{2}\beta(t)\Delta t (\mathbf{y}(t)-f_{\phi}(\mathbf{x})) + \sqrt{\beta(t)\Delta t} \ \varepsilon(t).
\end{aligned}
\end{equation*}
To put this in a more general form,
\begin{equation}\label{I3}
\mathbf{y}(t+\Delta t) -  \mathbf{y}(t) \approx-\frac{1}{2}\beta(t)\Delta t (\mathbf{y}(t)-f_{\phi}(\mathbf{x})) + \sqrt{\beta(t)\Delta t} \ \varepsilon(t).
\end{equation}
Eq.(\ref{I3}) is the discretized representation at $\Delta t \rightarrow 0$, if in the continuous case using stochastic differential equation can be expressed as
\begin{equation}\label{I4}
d \mathbf{y} = -\frac{1}{2}\beta(t)(\mathbf{y}-f_{\phi}(\mathbf{x})) dt + \sqrt{\beta(t)} dw,\ dw \sim N(0,\Delta t).
\end{equation}
In summary, Eq.(\ref{I4}) is the stochastic differential equation form of the forward process in CARD.

Second, the SDEs form of the CARD reverse process is derived, that is, Eq.(\ref{E2}). It is known that in CARD, the conditional distribution of the backward process is
\begin{equation*}
q\left(\mathbf{y}_{t-1} \mid \mathbf{y}_{t},\mathbf{y}_{0}, f_{\phi}(\mathbf{x}))\right) = N\left(\mathbf{\tilde{\mu}}(\mathbf{y}_{t},\mathbf{y}_{0},f_{\phi}(\mathbf{x})), \tilde{\beta}_{t} \mathbf{I} \right),
\end{equation*}
where
\begin{equation*}
\begin{aligned}
& \mathbf{\tilde{\mu}}(\mathbf{y}_{t},\mathbf{y}_{0},f_{\phi}(\mathbf{x})) = \frac{\beta_{t} \sqrt{\bar{\alpha}_{t-1}}}{1-\bar{\alpha}_{t}} \mathbf{y}_{0} + \frac{(1-\bar{\alpha}_{t-1}) \sqrt{\alpha_{t}}}{1-\bar{\alpha}_{t}} \mathbf{y}_{t} + \left(  1+\frac{(\sqrt{\bar{\alpha}_{t}}-1)(\sqrt{\alpha_{t}} + \sqrt{\bar{\alpha}_{t-1}})}{1-\bar{\alpha}_{t}} \right) f_{\phi}(\mathbf{x}),
%& \ \ \ \ \ \ \ \ \ \ \ \ \ \ \ \ \ \ \ \ \ \ \ \ \ \ + \left(  1+\frac{(\sqrt{\bar{\alpha}_{t}}-1)(\sqrt{\alpha_{t}} + \sqrt{\bar{\alpha}_{t-1}})}{1-\bar{\alpha}_{t}} \right) f_{\phi}(\boldsymbol{x}),\\
&\tilde{\beta}_{t} = \frac{1-\bar{\alpha}_{t-1}}{1-\bar{\alpha}_{t}} \beta_{t}.
\end{aligned}
\end{equation*}

\noindent Now that we have derived the SDEs form of the CARD forward process  which is Eq.(\ref{I3}), we can write the conditional distribution in terms of its discretization at $\Delta t \rightarrow 0$.
\begin{equation*}
q(\mathbf{y}(t+\Delta t)|\mathbf{y}(t),f_{\phi}(\mathbf{x})) =N\left( \mathbf{y}(t) -\frac{1}{2}\beta(t)(\mathbf{y}(t)-f_{\phi}(\mathbf{x}))\Delta t , \beta(t)\Delta t \mathbf{I} \right),
\end{equation*}
Using Bayes theorem, the posterior probability distribution is derived $q(\mathbf{y}(t)|\mathbf{y}(t+\Delta t),f_{\phi}(\mathbf{x}))$, which is consistent with the solution idea of reverse process in DDPM.
\begin{equation*}
\begin{aligned}
q(\mathbf{y}(t)|\mathbf{y}(t+\Delta t),f_{\phi}(\mathbf{x})) & = \frac{q(\mathbf{y}(t),\mathbf{y}(t+\Delta t),f_{\phi}(\mathbf{x}))}{q(\mathbf{y}(t+\Delta t),f_{\phi}(\mathbf{x}))}
%& = \frac{q(\mathbf{y}(t),\mathbf{y}(t+\Delta t),f_{\phi}(\mathbf{x}))}{q(\mathbf{y}(t),f_{\phi}(\mathbf{x}))} \frac{q(\mathbf{y}(t),f_{\phi}(\mathbf{x}))}{q(\mathbf{y}(t+\Delta t),f_{\phi}(\mathbf{x}))}\\
=q(\mathbf{y}(t+\Delta t)|\mathbf{y}(t),f_{\phi}(\mathbf{x}))  \frac{q(\mathbf{y}(t)|f_{\phi}(\mathbf{x}))}{q(\mathbf{y}(t+\Delta t)|f_{\phi}(\mathbf{x}))}\\
& =q(\mathbf{y}(t+\Delta t)|\mathbf{y}(t),f_{\phi}(\mathbf{x})) \cdot \exp \left\{ \log q(\mathbf{y}(t)|f_{\phi}(\mathbf{x}))-\log q(\mathbf{y}(t+\Delta t)|f_{\phi}(\mathbf{x})) \right\},\\
\end{aligned}
\end{equation*}
Since the specific distribution form of $q(\mathbf{y}(t+\Delta t)|\mathbf{y}(t),f_{\phi}(\mathbf{x}))$ is known, therefore
\begin{equation}\label{I5}
\begin{aligned}
q(\mathbf{y}(t)|\mathbf{y}(t+\Delta t),f_{\phi}(\mathbf{x})) &\propto \exp \left\{ \frac{|| \mathbf{y}(t+\Delta t)-\mathbf{y}(t)+ \frac{1}{2}\beta(t)(\mathbf{y}(t)-f_{\phi}(\mathbf{x}))\Delta t||^{2}}{2 \beta(t) \Delta t} \right\}\\
&\ \ \ +\exp\left\{ \log q(\mathbf{y}(t)|f_{\phi}(\mathbf{x}))-\log q(\mathbf{y}(t+\Delta t)|f_{\phi}(\mathbf{x}))  \right\}\\
&\propto HH_1+HH_2+HH_3.
\end{aligned}
\end{equation}
when $\Delta t \rightarrow 0$, we have 
\begin{equation*}
HH_1 = \lim\limits_{\Delta t \rightarrow 0} \exp\left\{  \frac{|| \mathbf{y}(t+\Delta t)-\mathbf{y}(t) ||^{2}}{2 \beta(t) \Delta t}  \right\}.
\end{equation*}
This fraction is obviously not 0 if and only if $\mathbf{y}(t+\Delta t),\mathbf{y}(t)$ is infinitely close, so $\Delta t \rightarrow 0$ is required for the posterior probability distribution to be meaningful.

The third term on the right $HH_3$ takes the Taylor expansion at $\mathbf{y}(t)$. For each input data in CARD, a model is pre-trained to obtain the conditional mean $f_{\phi}(\mathbf{x})$, which is used as the condition in the generative model. Thus, the only variables in the generative model are $\mathbf{y}(t),t$. When you take the Taylor expansion of a function with two variables, you take the partial derivative with respect to each variable, i.e.,
\begin{equation*}
\begin{aligned}
HH_3 &= \log q(\mathbf{y}(t)|f_{\phi}(\mathbf{x})) + \frac{\partial \log q(\mathbf{y}(t)|f_{\phi}(\mathbf{x}))}{\partial \mathbf{y}(t)} (\mathbf{y}(t+\Delta t)-\mathbf{y}(t)) +\frac{\partial \log q(\mathbf{y}(t)|f_{\phi}(\mathbf{x}))}{\partial t} (t+\Delta t-t) + o(\Delta t),
\end{aligned}
\end{equation*}
when $\Delta t \rightarrow 0$, we have $o(\Delta t)\rightarrow 0$ and this term is negligible. Bring back Eq.(\ref{I5}) to obtain
\begin{equation}\label{I6}
\begin{aligned}
q(\mathbf{y}(t)|\mathbf{y}(t+\Delta t),f_{\phi}(\mathbf{x})) &\propto \exp \left\{ \frac{|| \mathbf{y}(t+\Delta t)-\mathbf{y}(t)+ \frac{1}{2}\beta(t)(\mathbf{y}(t)-f_{\phi}(\mathbf{x}))\Delta t||^{2}}{2 \beta(t) \Delta t} \right\}\\
&\ \ \ -\exp\left\{ \frac{2 \beta(t) \Delta t \frac{\partial \log q(\mathbf{y}(t)|f_{\phi}(\mathbf{x}))}{\partial \mathbf{y}(t)} (\mathbf{y}(t+\Delta t)-\mathbf{y}(t))}{2 \beta(t) \Delta t}   \right\}\\
&\ \ \ -\exp\left\{ \frac{2 \beta(t) (\Delta t)^{2}  \frac{\partial \log q(\mathbf{y}(t)|f_{\phi}(\mathbf{x}))}{\partial t} }{2 \beta(t) \Delta t} \right\}
\end{aligned}
\end{equation}
Only the exponential part is proportional to 
\begin{equation*}
\frac{||\mathbf{y}(t+\Delta t)-\mathbf{y}(t)+  \left[ \frac{1}{2} (\mathbf{y}(t)-f_{\phi}(\mathbf{x}))- \frac{\partial \log q(\mathbf{y}(t)|f_{\phi}(\mathbf{x}))}{\partial \mathbf{y}(t)}   \right]\beta(t) \Delta t ||^{2}} {2 \beta(t) \Delta t} .
% \end{aligned}
\end{equation*}
The penultimate equality is that the last two terms of the cause tend to zero at $\Delta t \rightarrow 0$. Due to the conditional distribution $q(\mathbf{y}(t)|\mathbf{y}(t+\Delta t),f_{\phi}(\mathbf{x}))$, the distribution at time $t+\Delta t$ is known to $q(\mathbf{y}(t+\Delta t))$. So equivalently, results conditions becomes $q(\mathbf{y}(t+\Delta t)),f_{\phi}(\mathbf{x})$, namely the part into the index
\begin{equation*}
\frac{||\mathbf{y}(t)-\mathbf{y}(t+\Delta t)+  \left[ \frac{1}{2} (\mathbf{y}(t+\Delta t)-f_{\phi}(\mathbf{x}))- \frac{\partial \log q(\mathbf{y}(t+\Delta t)|f_{\phi}(\mathbf{x}))}{\partial \mathbf{y}(t+\Delta t)}   \right]\beta(t+\Delta t) \Delta t ||^{2}} {2 \beta(t+\Delta t) \Delta t} .
\end{equation*}
If we denote the backward distribution as $p$, then
\begin{equation}\label{I7}
p(\mathbf{y}(t)|\mathbf{y}(t+\Delta t),f_{\phi}(\mathbf{x})) = N(\mu_{1}, \beta_{1}),
\end{equation}
where
\begin{equation*}
\begin{aligned}
& \mu_{1} = \mathbf{y}(t+\Delta t)- \left[ \frac{1}{2} (\mathbf{y}(t+\Delta t)-f_{\phi}(\mathbf{x}))- \frac{\partial \log q(\mathbf{y}(t+\Delta t)|f_{\phi}(\mathbf{x}))}{\partial \mathbf{y}(t+\Delta t)}   \right]\beta(t+\Delta t) \Delta t,
& \beta_{1} = 2 \beta(t+\Delta t)\Delta t \mathbf(I) .
\end{aligned}
\end{equation*}
The corresponding SDEs are denoted by
\begin{equation}\label{I8}
dy = -\left[\frac{1}{2} (\mathbf{y}-f_{\phi}(\mathbf{x}))- \frac{\partial \log q(\mathbf{y}|f_{\phi}(\mathbf{x}))}{\partial \mathbf{y}} \right]\beta(t) dt + \sqrt{\beta(t)}dw,\ dw \sim N(0,I).
\end{equation}

Finally, Eq.(\ref{I4}) and Eq.(\ref{I8}) are the forms of stochastic differential equations for the forward and backward processes of the classification and regression diffusion models.

\subsection{Proof of the Fokker-Planck equation for CARD}\label{appendixA2}
Eq.(\ref{E3}) describe the Fokker-Planck equation (FP for short) corresponding to the model. The proof idea of the two processes is roughly the same, but the differences are that the forward process changes from time 0 to time $T$, and the reverse process starts from time $T$ and gradually pushes back to time 0. Suppose that the forward process starts at time 0 and arrives at time t after time interval $\Delta t$, and the reverse process arrives at time $t'$ after the same time interval. Then it is clear that $t'=T-t$. Thus the two are slightly different in the proof of FP function. The proof mainly uses Ito's lemma \citep{oksendal2013stochastic} and involves distributional integration and Fubini's lemma. 

First, We prove the FP equation of the forward process. Suppose there exists a function $F(\mathbf{y}_{t},t)$, which satisfy:
(1) It is second-order continuously derivable for $\mathbf{y}_{t}$.
(2) It is first derivable with respect to $t$.
(3) For all $t_{1}<t_{2}$, $F(\mathbf{y}_{t_{1}},t_{1})=F(\mathbf{y}_{t_{2}},t_{2})=0$.
(4) When $\mathbf{y}_{t} \rightarrow \pm\infty$, $ F(\mathbf{y}_{t},t) \rightarrow 0,\frac{\partial F(\mathbf{y}_{t},t)}{\partial \mathbf{y}_{t}}=F_{\mathbf{y}_{t}} \rightarrow 0$.
By Ito's lemma, differentiating the function $F(\mathbf{y},t)$ follows
\begin{equation}\label{J3}
\begin{aligned}
d F(\mathbf{y}_{t},t) &= \frac{\partial F}{\partial t}dt + \frac{\partial F}{\partial \mathbf{y}_{t}}d\mathbf{y}_{t} + \frac{1}{2}\frac{\partial^{2} F}{\partial \mathbf{y}_{t}^{2}} (d\mathbf{y}_{t})^{2}=F_{t} dt+F_{\mathbf{y}_{t}} d\mathbf{y}_{t} + \frac{1}{2}F^{2}_{\mathbf{y}_{t}} (d\mathbf{y}_{t})^{2}.
\end{aligned}
\end{equation}
Combined with differential equations for the CARD forward process Eq.(\ref{E1}), 
\begin{equation*}
\begin{aligned}
d F(\mathbf{y}_{t},t) 
% &=F_{t} dt+F_{\mathbf{y}_{t}} \left(-\frac{1}{2}\beta_{t}(\mathbf{y}_{t}-f_{\phi}(\mathbf{x})) dt + \sqrt{\beta_{t}} dw\right) + \frac{1}{2}F^{2}_{\mathbf{y}_{t}} \left(-\frac{1}{2}\beta_{t}(\mathbf{y}_{t}-f_{\phi}(\mathbf{x})) dt + \sqrt{\beta_{t}} dw\right)^{2}\\
&= F_{t}dt + F_{\mathbf{y}_{t}} \left(-\frac{1}{2}\beta_{t}(\mathbf{y}-f_{\phi}(\mathbf{x}))\right) dt + F_{\mathbf{y}_{t}}\sqrt{\beta_{t}} dw + \frac{1}{2}F^{2}_{\mathbf{y}_{t}}\left( \beta_{t}(dw)^{2} \right)\\
&\ \ \ + \frac{1}{2}F^{2}_{\mathbf{y}_{t}}\left(\left(-\frac{1}{2}\beta_{t}(\mathbf{y}_{t}-f_{\phi}(\mathbf{x}))\right)^{2} (dt)^{2} \right) + F^{2}_{\mathbf{y}_{t}}\left( -\frac{1}{2}(\beta_{t})^{\frac{3}{2}}(\mathbf{y}_{t}-f_{\phi}(\mathbf{x}))  dtdw \right),
\end{aligned}
\end{equation*}
We know that $dw =\sqrt{dt}$ in SDEs, and we only consider the terms of order one or lower in $dt$, ignoring terms of order higher than it, such as $(dt)^{2},(dt)^{\frac{3}{2}}$, etc. The final simplification is
\begin{equation}\label{J4}
d F(\mathbf{y}_{t},t) = \left(F_{t}+ F_{\mathbf{y}_{t}} \left(-\frac{1}{2}\beta_{t}(\mathbf{y}_{t}-f_{\phi}(\mathbf{x}))\right) + \frac{1}{2} F^{2}_{\mathbf{y}_{t}}\beta_{t}  \right) dt + F_{\mathbf{y}_{t}}\sqrt{\beta_{t}} dw.
\end{equation}
We use the integral of continuous functions to solve, with the help of mathematical expectation idea. Let's take the expectation with respect to t on both sides of Eq.(\ref{J4}),
\begin{equation*}
\begin{aligned}
E[d F(\mathbf{y}_{t},t)] 
% &= E\left[  \left(F_{t}+ F_{\mathbf{y}_{t}} \left(-\frac{1}{2}\beta_{t}(\mathbf{y}_{t}-f_{\phi}(\mathbf{x}))\right) + \frac{1}{2} F^{2}_{\mathbf{y}_{t}}\beta_{t}  \right) dt + F_{\mathbf{y}_{t}}\sqrt{\beta_{t}} dw \right]\\
&=E\left[  \left(F_{t}+ F_{\mathbf{y}_{t}} \left(-\frac{1}{2}\beta_{t}(\mathbf{y}_{t}-f_{\phi}(\mathbf{x}))\right) + \frac{1}{2} F^{2}_{\mathbf{y}_{t}}\beta_{t}  \right) dt\right],
\end{aligned}
\end{equation*}
The second equal sign above is because $E[dw]=0$, which reduces the above equation to integral form by definition,
\begin{equation*}
\begin{aligned}
\int_{t_{1}}^{t_{2}}d F(\mathbf{y}_{t},t) &= \int_{t_{1}}^{t_{2}}  \left(F_{t}+ F_{\mathbf{y}_{t}} \left(-\frac{1}{2}\beta_{t}(\mathbf{y}_{t}-f_{\phi}(\mathbf{x}))\right) + \frac{1}{2} F^{2}_{\mathbf{y}_{t}}\beta_{t}  \right) dt ,
\end{aligned}
\end{equation*}
According to the assumption of the function $F$,
\begin{equation*}
\int_{t_{1}}^{t_{2}}d F(\mathbf{y}_{t},t) =F(\mathbf{y}_{t},t_{2})-F(\mathbf{y}_{t},t_{1})=0,
\end{equation*}
To both sides to take about $\mathbf{y}_{t}$ expectations and integral form, 
\begin{equation*}
\begin{aligned}
0&= \int_{-\infty}^{+\infty} \int_{t_{1}}^{t_{2}}  \left(F_{t}+ F_{\mathbf{y}_{t}} \left(-\frac{1}{2}\beta_{t}(\mathbf{y}_{t}-f_{\phi}(\mathbf{x}))\right) + \frac{1}{2} F^{2}_{\mathbf{y}_{t}}\beta_{t}  \right) q(\mathbf{y}_{t}|f_{\phi}(\mathbf{x})) dt d\mathbf{y}_{t}\\
 & = \int_{-\infty}^{+\infty} \int_{t_{1}}^{t_{2}} F_{t}  q(\mathbf{y}_{t}|f_{\phi}(\mathbf{x})) dtdy + F_{\mathbf{y}_{t}} \left(-\frac{1}{2}\beta_{t}(\mathbf{y}_{t}-f_{\phi}(\mathbf{x}))\right) q(\mathbf{y}_{t}|f_{\phi}(\mathbf{x})) dt d\mathbf{y}_{t} 
+ \frac{1}{2} \int_{-\infty}^{+\infty}  \int_{t_{1}}^{t_{2}} F^{2}_{\mathbf{y}_{t}}\beta_{t} q(\mathbf{y}_{t}|f_{\phi}(\mathbf{x}))dt d\mathbf{y}_{t}\\
 &= I_{1}+I_{2}+I_{3}.
\end{aligned}
\end{equation*}
Simplification as follows separately.
\begin{equation*}
\begin{aligned}
I_{1}
=\int_{-\infty}^{+\infty} \left( F(\mathbf{y}_{t},t)q(\mathbf{y}_{t}|f_{\phi}(\mathbf{x}))|^{t_{2}}_{t_{1}}-  \int_{t_{1}}^{t_{2}}F(\mathbf{y}_{t},t) \frac{\partial}{\partial t}q(\mathbf{y}_{t}|f_{\phi}(\mathbf{x})) dt  \right) d\mathbf{y}_{t}
% &=- \int_{-\infty}^{+\infty} \int_{t_{1}}^{t_{2}} F(\mathbf{y}_{t},t) \frac{\partial}{\partial t}q(\mathbf{y}_{t}|f_{\phi}(\mathbf{x})) dt  d\mathbf{y}_{t} \\
=-\int_{t_{1}}^{t_{2}} \int_{-\infty}^{+\infty} F(\mathbf{y}_{t},t) \frac{\partial}{\partial t}q(\mathbf{y}_{t}|f_{\phi}(\mathbf{x}))  d\mathbf{y}_{t} dt.
\end{aligned}
\end{equation*}
and
\begin{equation*}
\begin{aligned}
I_{2}
&=\int_{t_{1}}^{t_{2}} \left[ F(\mathbf{y}_{t},t)\left(-\frac{1}{2}\beta_{t}(\mathbf{y}_{t}-f_{\phi}(\mathbf{x}))\right) q(\mathbf{y}_{t}|f_{\phi}(\mathbf{x}))|_{-\infty}^{+\infty} \right. - \left. \int_{-\infty}^{+\infty} F(\mathbf{y}_{t},t) \frac{\partial}{\partial \mathbf{y}_{t}}\left(\left(-\frac{1}{2}\beta_{t}(\mathbf{y}_{t}-f_{\phi}(\mathbf{x}))\right)q(\mathbf{y}_{t}|f_{\phi}(\mathbf{x})) \right) d\mathbf{y}_{t}  \right]dt\\
&\approx -\int_{t_{1}}^{t_{2}} \int_{-\infty}^{+\infty} F(\mathbf{y}_{t},t) \frac{\partial}{\partial \mathbf{y}_{t}}\left(\left(-\frac{1}{2}\beta_{t}(\mathbf{y}_{t}-f_{\phi}(\mathbf{x}))\right)q(\mathbf{y}_{t}|f_{\phi}(\mathbf{x})) \right) d\mathbf{y}_{t}  dt.
\end{aligned}
\end{equation*}
In the same way, we have
\begin{equation*}
\begin{aligned}
I_{3}
&=\frac{1}{2} \int_{t_{1}}^{t_{2}} \left[ F_{\mathbf{y}_{t}} \beta_{t} q(\mathbf{y}_{t}|f_{\phi}(\mathbf{x}))|_{-\infty}^{+\infty} -\int_{-\infty}^{+\infty} F_{\mathbf{y}_{t}} \beta_{t}  \frac{\partial}{\partial \mathbf{y}_{t}}\left( q(\mathbf{y}_{t}|f_{\phi}(\mathbf{x})) \right) d\mathbf{y}_{t}\right] dt\\
&\approx -\frac{1}{2} \int_{t_{1}}^{t_{2}} \int_{-\infty}^{+\infty} F_{\mathbf{y}_{t}} \beta_{t}  \frac{\partial}{\partial \mathbf{y}_{t}}\left( q(\mathbf{y}_{t}|f_{\phi}(\mathbf{x})) \right) d\mathbf{y}_{t} dt\\
% &=-\frac{1}{2} \int_{t_{1}}^{t_{2}} \left[ F(\mathbf{y}_{t},t) \beta_{t}\frac{\partial}{\partial \mathbf{y}_{t}}\left( q(\mathbf{y}_{t}|f_{\phi}(\mathbf{x})) \right)|_{-\infty}^{+\infty} - \int_{-\infty}^{+\infty} F(\mathbf{y}_{t},t)  \beta_{t}  \frac{\partial^{2}}{\partial \mathbf{y}_{t}^{2}}\left( q(\mathbf{y}_{t}|f_{\phi}(\mathbf{x})) \right) d\mathbf{y}_{t} \right] dt\\
&\approx \frac{1}{2} \int_{t_{1}}^{t_{2}}\int_{-\infty}^{+\infty} F(\mathbf{y}_{t},t)  \beta_{t}  \frac{\partial^{2}}{\partial \mathbf{y}_{t}^{2}}\left( q(\mathbf{y}_{t}|f_{\phi}(\mathbf{x})) \right) d\mathbf{y}_{t} dt.
\end{aligned}
\end{equation*}
Because of $I_{1}+I_{2}+I_{3}=0$, then 
\begin{equation*}\label{J5}
\begin{aligned}
&\int_{t_{1}}^{t_{2}}\left[ \int_{-\infty}^{+\infty} F(\mathbf{y}_{t},t) \left( - \frac{\partial}{\partial t}q(\mathbf{y}_{t}|f_{\phi}(\mathbf{x})) - \frac{\partial}{\partial \mathbf{y}_{t}}\left(\left(-\frac{1}{2}\beta_{t}(\mathbf{y}_{t}-f_{\phi}(\mathbf{x}))\right)q(\mathbf{y}_{t}|f_{\phi}(\mathbf{x})) \right) \right. \right.
%&\ \ \ \ \ \ \ \ \ \ \ \ \ \ \ \ \ \ \ \ \ \ \ \ \ \ \ 
\left.\left.+\frac{1}{2}  \beta_{t}  \frac{\partial^{2}}{\partial \mathbf{y}_{t}^{2}}\left( q(\mathbf{y}_{t}|f_{\phi}(\mathbf{x})) \right) \right) d\mathbf{y}_{t} \right]dt = 0.
\end{aligned}
\end{equation*}
Since the function $F$ is arbitrarily chosen to satisfy the above assumptions, the above equation only needs to be true if the polynomial multiplied by the function $F$ is always 0.
\begin{equation*}
-\frac{\partial}{\partial t}q(\mathbf{y}_{t}|f_{\phi}(\mathbf{x}))
-\frac{\partial}{\partial \mathbf{y}}\left(\left(-\frac{1}{2}\beta_{t}(\mathbf{y}_{t}-f_{\phi}(\mathbf{x}))\right)q(\mathbf{y}_{t}|f_{\phi}(\mathbf{x})) \right)
+\frac{1}{2} \beta_{t}  \frac{\partial^{2}}{\partial \mathbf{y}_{t}^{2}}\left( q(\mathbf{y}_{t}|f_{\phi}(\mathbf{x})) \right) =0.
\end{equation*}
We obtain the FP equation for the CARD forward process.
\begin{equation*}
\frac{\partial }{\partial t}q(\mathbf{y}_{t}|f_{\phi}(\mathbf{x})) + \frac{\partial}{\partial \mathbf{y}_{t}} \left[-\frac{1}{2} \beta_{t} (\mathbf{y}_{t}-f_{\phi}(\mathbf{x})) q(\mathbf{y}_{t}|f_{\phi}(\mathbf{x})) \right] - \frac{1}{2} \beta_{t} \frac{\partial^{2}}{\partial \mathbf{y}_{t}^{2}} q(\mathbf{y}_{t}|f_{\phi}(\mathbf{x}))=0.
\end{equation*}

Second, we prove the FP equation of CARD backward process.
Combined with the differential equations of the CARD backward process Eq.(\ref{E2}), we have
\begin{equation*}
d F(\mathbf{y}_{t},t')
=F_{t'} dt'+F_{\mathbf{y}_{t'}} d\mathbf{y}_{t'} + \frac{1}{2}F^{2}_{\mathbf{y}_{t'}} (d\mathbf{y}_{t'})^{2}.
\end{equation*}
Since the backward process starts at time $t$, it follows that $t'=T-t$, i.e
\begin{equation*}
\begin{aligned}
d F(\mathbf{y}_{T-t},T-t)
&=F_{T-t} d(T-t)+F_{\mathbf{y}_{T-t}} d\mathbf{y}_{T-t} + \frac{1}{2}F^{2}_{\mathbf{y}_{T-t}} (d\mathbf{y}_{T-t})^{2}\\
&=-F_{T-t} dt + F_{\mathbf{y}_{T-t}} \left(-\frac{1}{2} (\mathbf{y}_{T-t}-f_{\phi}(\mathbf{x}))+\frac{\partial }{\partial \mathbf{y}_{T-t}}\log p(\mathbf{y}_{t}|f_{\phi}(\mathbf{x})) \right)\\
&\ \ \ \cdot \beta_{T-t} d(T-t) + F_{\mathbf{y}_{T-t}} \sqrt{\beta_{T-t}} d\bar{w} + \frac{1}{2}  F^{2}_{\mathbf{y}_{T-t}} \beta_{T-t} (d\bar{w})^{2}\\
&\ \ \ + \frac{1}{2} F^{2}_{\mathbf{y}_{T-t}} \left(-\frac{1}{2} (\mathbf{y}_{T-t}-f_{\phi}(\mathbf{x}))+\frac{\partial }{\partial \mathbf{y}_{T-t}}\log p(\mathbf{y}_{T-t}|f_{\phi}(\mathbf{x})) \right)^{2} \\
&\ \ \ \cdot (\beta_{T-t})^{2} (d(T-t))^{2} + F^{2}_{\mathbf{y}_{T-t}} \sqrt{\beta_{T-t}} \ d\bar{w}\ d(T-t) \\
&=\left[ -F_{T-t} dt - F_{\mathbf{y}_{T-t}}  \left(-\frac{1}{2} (\mathbf{y}_{T-t}-f_{\phi}(\mathbf{x}))+\frac{\partial }{\partial \mathbf{y}_{T-t}}\log p(\mathbf{y}_{t}|f_{\phi}(\mathbf{x})) \right)\right.\\
&\left.\ \ \ \cdot \beta_{T-t} + \frac{1}{2}  F^{2}_{\mathbf{y}_{T-t}} \beta_{T-t} \right] dt + F_{\mathbf{y}_{T-t}} \sqrt{\beta_{T-t}} d\bar{w}.
\end{aligned}
\end{equation*}
For convenience, 
\begin{equation*}
\mathbf{m} = -\frac{1}{2} (\mathbf{y}_{T-t}-f_{\phi}(\mathbf{x}))+\frac{\partial }{\partial \mathbf{y}_{T-t}}\log p(\mathbf{y}_{T-t}|f_{\phi}(\mathbf{x})).
\end{equation*}
then
\begin{equation}\label{J6}
d F(\mathbf{y}_{T-t},T-t)=\left( -F_{T-t}  - F_{\mathbf{y}_{T-t}}  \mathbf{m} \beta_{T-t} + \frac{1}{2}  F^{2}_{\mathbf{y}_{T-t}} \beta_{T-t} \right) dt + F_{\mathbf{y}_{T-t}} \sqrt{\beta_{T-t}} d\bar{w}.
\end{equation}
Taking the expectation with respect to $t$ on both sides of Eq.(\ref{J6}) and converting it to integral form.
\begin{equation*}
\begin{aligned}
\textup{E}[d F(\mathbf{y}_{T-t},T-t)] 
&= \textup{E}\left[  \left( -F_{T-t} - F_{\mathbf{y}_{T-t}}  \mathbf{m} \beta_{T-t} + \frac{1}{2}  F^{2}_{\mathbf{y}_{T-t}} \beta_{T-t} \right) dt + F_{\mathbf{y}_{T-t}} \sqrt{\beta_{T-t}} d\bar{w} \right]\\
&=\textup{E}\left[  \left( -F_{T-t}  - F_{\mathbf{y}_{T-t}}  \mathbf{m} \beta_{T-t} + \frac{1}{2}  F^{2}_{\mathbf{y}_{T-t}} \beta_{T-t} \right) dt \right],
\end{aligned}
\end{equation*}
According to the assumption $\forall t_{1}<t_{1},F(\mathbf{y}_{t_{1}},t_{1})=F(\mathbf{y}_{t_{2}},t_{2})=0$, we detrive that $F(\mathbf{y}_{T-t_{1}},T-t_{1})=F(\mathbf{y}_{T-t_{2}},T-t_{2})=0$.
\begin{equation*}
\begin{aligned}
0 &= \int_{t_{1}}^{t_{2}}  \left( -F_{T-t}  - F_{\mathbf{y}_{T-t}}  \mathbf{m} \beta_{T-t} + \frac{1}{2}  F^{2}_{\mathbf{y}_{T-t}} \beta_{T-t} \right) dt,
\end{aligned}
\end{equation*}
At the same time, taking the expectation of $\mathbf{y}_{T-t}$ and converting it into integral form, we can obtain
\begin{equation*}
\begin{aligned}
0&= \int_{-\infty}^{+\infty} \int_{t_{1}}^{t_{2}}  \left( -F_{T-t} - F_{\mathbf{y}_{T-t}}  \mathbf{m} \beta_{T-t} + \frac{1}{2}  F^{2}_{\mathbf{y}_{T-t}} \beta_{T-t} \right) p(\mathbf{y}_{T-t}|f_{\phi}(\mathbf{x})) dt d\mathbf{y}_{T-t}\\
 & = -\int_{-\infty}^{+\infty} \int_{t_{1}}^{t_{2}} F_{T-t} p(\mathbf{y}_{T-t}|f_{\phi}(\mathbf{x})) dt d\mathbf{y}_{T-t} 
 - \int_{-\infty}^{+\infty} \int_{t_{1}}^{t_{2}} F_{\mathbf{y}_{T-t}}  \mathbf{m} \beta_{T-t} p(\mathbf{y}_{T-t}|f_{\phi}(\mathbf{x})) dt d\mathbf{y}_{T-t}\\
 &\ \ \ + \frac{1}{2} \int_{-\infty}^{+\infty} \int_{t_{1}}^{t_{2}} F^{2}_{\mathbf{y}_{T-t}} \beta_{T-t} p(\mathbf{y}_{T-t}|f_{\phi}(\mathbf{x})) dt d\mathbf{y}_{T-t}\\
 &= I_{4}+I_{5}+I_{6}.
\end{aligned}
\end{equation*}
The derivation process following this section is essentially identical to the forward process, with the exception that the reverse process alters the time range, resulting in a change of sign for certain terms and consequently leading to different conclusions. Additionally, both distributed integral and Fubini's theorem are employed. The results are calculated separately as follows: 
\begin{equation*}
\begin{aligned}
I_{4}
&=-\int_{-\infty}^{+\infty} \left[ F(\mathbf{y}_{T-t},T-t)q(\mathbf{y}_{T-t}|f_{\phi}(\mathbf{x}))|^{t_{2}}_{t_{1}} \right.
\left.-\int_{t_{1}}^{t_{2}} F(\mathbf{y}_{T-t},T-t) \frac{\partial}{\partial (T-t)}p(\mathbf{y}_{T-t}|f_{\phi}(\mathbf{x})) dt  \right] d\mathbf{y}_{T-t}\\
% &=\int_{-\infty}^{+\infty} \int_{t_{1}}^{t_{2}} F(\mathbf{y}_{T-t},T-t) \frac{\partial}{\partial (T-t)}p(\mathbf{y}_{T-t}|f_{\phi}(\mathbf{x})) dt  d\mathbf{y}_{T-t}\\
&=\int_{t_{1}}^{t_{2}} \int_{-\infty}^{+\infty} F(\mathbf{y}_{T-t},T-t) \frac{\partial}{\partial (T-t)}p(\mathbf{y}_{T-t}|f_{\phi}(\mathbf{x})) dt  d\mathbf{y}_{T-t}.
\end{aligned}
\end{equation*}
and
\begin{equation*}
\begin{aligned}
I_{5}
&=-\int_{t_{1}}^{t_{2}} \left[ F(\mathbf{y}_{T-t},T-t) \mathbf{m}  \beta_{T-t} p(\mathbf{y}_{T-t}|f_{\phi}(\mathbf{x}))|_{-\infty}^{+\infty} \right.
- \left. \int_{-\infty}^{+\infty} F(\mathbf{y}_{T-t},T-t) \frac{\partial}{\partial \mathbf{y}_{T-t}}\left(\mathbf{m} \beta_{T-t} p(\mathbf{y}_{T-t}|f_{\phi}(\mathbf{x})) \right) d\mathbf{y}_{T-t}  \right]dt\\
&\approx \int_{t_{1}}^{t_{2}} \int_{-\infty}^{+\infty} F(\mathbf{y}_{T-t},T-t) \frac{\partial}{\partial \mathbf{y}_{T-t}} \left(\mathbf{m} \beta_{T-t} p(\mathbf{y}_{T-t}|f_{\phi}(\mathbf{x}))\right) d\mathbf{y}_{T-t}  dt.
\end{aligned}
\end{equation*}
Similarly, we have
\begin{equation*}
\begin{aligned}
I_{6}
&=\frac{1}{2} \int_{t_{1}}^{t_{2}} \left[ F_{\mathbf{y}_{T-t}} \beta_{T-t} p(\mathbf{y}_{T-t}|f_{\phi}(\mathbf{x}))|_{-\infty}^{+\infty} -\int_{-\infty}^{+\infty} F_{\mathbf{y}_{T-t}} \beta_{T-t}  \frac{\partial}{\partial \mathbf{y}_{T-t}} p(\mathbf{y}_{T-t}|f_{\phi}(\mathbf{x}))  d\mathbf{y}_{T-t}\right] dt\\
% &\approx -\frac{1}{2} \int_{t_{1}}^{t_{2}} \int_{-\infty}^{+\infty} F_{\mathbf{y}_{T-t}} \beta_{T-t}  \frac{\partial}{\partial \mathbf{y}_{T-t}} p(\mathbf{y}_{t}|f_{\phi}(\mathbf{x}))  d\mathbf{y}_{t} dt\\
%&=-\frac{1}{2} \int_{t_{1}}^{t_{2}} \left[ F(\mathbf{y}_{T-t},T-t) \beta_{T-t} \frac{\partial}{\partial \mathbf{y}_{T-t}} q(\mathbf{y}_{t}|f_{\phi}(\mathbf{x}))|_{-\infty}^{+\infty} \right.\\
%&\ \ \ \ \left.- \int_{-\infty}^{+\infty} F(\mathbf{y}_{T-t},T-t)  \beta_{T-t}  \frac{\partial^{2}}{\partial \mathbf{y}_{T-t}^{2}} p(\mathbf{y}_{T-t}|f_{\phi}(\mathbf{x}))  d\mathbf{y}_{T-t} \right] dt\\
&\approx \frac{1}{2} \int_{t_{1}}^{t_{2}} \int_{-\infty}^{+\infty} F(\mathbf{y}_{T-t},T-t)  \beta_{T-t}  \frac{\partial^{2}}{\partial \mathbf{y}_{T-t}^{2}} p(\mathbf{y}_{T-t}|f_{\phi}(\mathbf{x}))  d\mathbf{y}_{T-t} dt.
\end{aligned}
\end{equation*}
Since $I_{4}+I_{5}+I_{6}=0$, then 
%\begin{equation}\label{J7}
%\begin{aligned}
%0&=I_{4}+I_{5}+I_{6}\\
%&\approx \int_{t_{1}}^{t_{2}} \int_{-\infty}^{+\infty} \left[ F(\mathbf{y}_{T-t},T-t) \left( \frac{\partial}{\partial (T-t)}p(\mathbf{y}_{T-t}|f_{\phi}(\mathbf{x}))+\frac{\partial}{\partial \mathbf{y}_{T-t}} \left(\mathbf{m} \beta_{T-t} p(\mathbf{y}_{T-t}|f_{\phi}(\mathbf{x}))\right) \right.\right.\\
%&\left.\left.\ \ \ \ \ \ \ \ \ \ \ \ \ \ \ \ \ \ \ \ \ \ \ \ + \frac{1}{2}\beta_{T-t}  \frac{\partial^{2}}{\partial \mathbf{y}_{T-t}^{2}} p(\mathbf{y}_{T-t}|f_{\phi}(\mathbf{x})) \right) \right]dt  d\mathbf{y}_{T-t},
%\end{aligned}
%\end{equation}
\begin{equation*}
\frac{\partial}{\partial (T-t)}p(\mathbf{y}_{T-t}|f_{\phi}(\mathbf{x}))+\frac{\partial}{\partial \mathbf{y}_{T-t}} \left(\mathbf{m} \beta_{T-t} p(\mathbf{y}_{T-t}|f_{\phi}(\mathbf{x}))\right)+ \frac{1}{2}  \frac{\partial^{2}}{\partial \mathbf{y}_{T-t}^{2}} (\beta_{T-t} p(\mathbf{y}_{T-t}|f_{\phi}(\mathbf{x})) )=0,
\end{equation*}
Using $t$ instead of $T - t$,
\begin{equation*}
\begin{aligned}
\frac{\partial}{\partial t}p(\mathbf{y}_{t}|f_{\phi}(\mathbf{x}))& +\frac{\partial}{\partial \mathbf{y}_{t}} \left[ \left(-\frac{1}{2} (\mathbf{y}_{t}-f_{\phi}(\mathbf{x}))+\frac{\partial }{\partial \mathbf{y}_{t}}\log p(\mathbf{y}_{t}|f_{\phi}(\mathbf{x}))\right) \beta_{t} p(\mathbf{y}_{t}|f_{\phi}(\mathbf{x}))\right] + \frac{1}{2} \beta_{t} \frac{\partial^{2}}{\partial \mathbf{y}_{t}^{2}} p(\mathbf{y}_{t}|f_{\phi}(\mathbf{x}))=0.
\end{aligned}
\end{equation*}
The above equation is the FP equation of the CARD reverse process, and the proof is complete.

\subsection{Proof of Theorem \ref{the-1}}\label{appendixA3}
We establish the claim between the  FP equation and the Wasserstein distance of order n on the probability space necessitates.

We have derived the SDE form of CARD, and in reverse, using the idea of the score function, with $s_{\theta}(\mathbf{y},f_{\phi}(\mathbf{x}),t)$ says unknown gradient $\partial \log p(\mathbf{y}|f_{\phi}(\mathbf{x})) / \partial \mathbf{y} $ for the solution. That is, the SDE of the reverse process Eq.(\ref{E2}) becomes
\begin{equation}\label{K1}
d\mathbf{y} = - \left[ \frac{1}{2} (\mathbf{y}-f_{\phi}(\mathbf{x}))- \frac{\partial \log p(\mathbf{y}|f_{\phi}(\mathbf{x}))}{\partial \mathbf{y} }\right] \beta_{t} dt + \sqrt{\beta_{t}}d\bar{w},\ \ d\bar{w} \sim N(0,\Delta t).
\end{equation}
The corresponding FP equation is given by
\begin{equation}\label{K2}
\begin{aligned}
\frac{\partial }{\partial t}p(\mathbf{y}_{t}|f_{\phi}(\mathbf{x}))  &+\frac{\partial}{\partial \mathbf{y}_{t}} \left[-\left(\frac{1}{2} (\mathbf{y_{t}}-f_{\phi}(\mathbf{x}))-s_{\theta}(\mathbf{y}_{t},f_{\phi}(\mathbf{x}),t) \right)\beta_{t}p(\mathbf{y}_{t}|f_{\phi}(\mathbf{x})) \right] + \frac{1}{2} \beta_{t} \frac{\partial^{2} }{\partial \mathbf{y}_{t}^{2}}p(\mathbf{y}_{t}|f_{\phi}(\mathbf{x}))=0.
\end{aligned}
\end{equation}
With the help of the proof idea in \cite{kwon2022score}, we first express the two FP equations by continuity equations, and then connect them by the optimal transport theory.

Assuming $q(\mathbf{y}_{t}|f_{\phi}(\mathbf{x})),p(\mathbf{z}_{t}|f_{\phi}(\mathbf{x}))$ is the Solution of Eq.(\ref{E3}) and Eq.(\ref{K2}). It is simply written as $q_{t},p_{t}$. Let $\mathbf{v}$ denote the velocity field vector, whose velocity field with respect to $q_{t},p_{t}$ is defined as
\begin{equation}\label{K3}
\begin{aligned}
& \mathbf{v}[q_{t}](\mathbf{y}_{t}) = -\frac{1}{2}\beta_{t} (\mathbf{y}_{t}-f_{\phi}(\mathbf{x})) -\frac{1}{2} \beta_{t} \frac{\partial}{\partial \mathbf{y}_{t} }\log q_{t},\\
& \mathbf{v}[p_{t}](\mathbf{z}_{t}) = \left(-\frac{1}{2} (\mathbf{z_{t}}-f_{\phi}(\mathbf{x}))+s_{\theta}(\mathbf{z}_{t},f_{\phi}(\mathbf{x}),t) \right) \beta_{t} + \frac{1}{2} \beta_{t} \frac{\partial }{\partial \mathbf{z}_{t}} \log p_{t}.
\end{aligned}
\end{equation}
Then Eq.(\ref{E3}) and Eq.(\ref{K2}) can be written
\begin{equation*}
\begin{aligned}
& \frac{\partial }{\partial t}q_{t} + \frac{\partial}{\partial \mathbf{y}_{t}} (q_{t} \mathbf{v}[q_{t}])=0,
& \frac{\partial }{\partial t}p_{t} + \frac{\partial}{\partial \mathbf{z}_{t}} (p_{t} \mathbf{v}[p_{t}])=0,
\end{aligned}
\end{equation*}
% Theorem 8.4.7   Corollary 5.25
Both \cite{ambrosio2005gradient} and \cite{santambrogio2015optimal} describe the optimal transport theory with respect to the Wasserstein distance of order n. So if we assume that $\pi_{t}$ is the optimal transport from $q_{t}$ to $p_{t}$, we have the conclusion
\begin{equation}\label{K4}
\frac{1}{2} \frac{\partial }{\partial t} W^{2}_{2}(q_{t},p_{t}) = E_{\pi_{t}} \left[ ( \mathbf{y}_{t}- \mathbf{z}_{t}) (\mathbf{v}[p_{t}](\mathbf{y}_{t}) - \mathbf{v}[p_{t}](\mathbf{z}_{t}) )  \right].
\end{equation}
According to the assumptions in Section \ref{sec3}, especially the Lipschitz continuity assumption, the left hand side in Eq. (\ref{K4}) simplifies as 
\begin{equation}\label{K5}
-\frac{1}{2} \frac{\partial }{\partial t} W^{2}_{2}(q_{t},p_{t}) = -W_{2}(q_{t},p_{t}) \left( \frac{\partial }{\partial t} W_{2}(q_{t},p_{t}) \right),
\end{equation}
and the right hand side in Eq.(\ref{K4}) simplifies as 
\begin{equation*}
\begin{aligned}
-E_{\pi_{t}} \left[ ( \mathbf{y}_{t}- \mathbf{z}_{t}) (\mathbf{v}[p_{t}](\mathbf{y}_{t}) - \mathbf{v}[p_{t}](\mathbf{z}_{t}) )  \right] 
&=-E_{\pi_{t}} \left[( \mathbf{y}_{t}- \mathbf{z}_{t}) \frac{1}{2}\beta_{t} ( \mathbf{z}_{t}- \mathbf{y}_{t}) \right]\\
&\ \ \ +E_{\pi_{t}} \left[( \mathbf{y}_{t}- \mathbf{z}_{t}) \beta_{t} \left( s_{\theta}(\mathbf{z}_{t},f_{\phi}(\mathbf{x}),t)-\frac{\partial }{\partial \mathbf{y}_{t}} \log q_{t} \right) \right]\\
&\ \ \ +\frac{1}{2}E_{\pi_{t}} \left[ ( \mathbf{y}_{t}- \mathbf{z}_{t}) \beta_{t} \left( \frac{\partial }{\partial \mathbf{z}_{t}} \log p_{t}-\frac{\partial }{\partial \mathbf{y}_{t}} \log q_{t} \right) \right]\\
&=D_{1}+D_{2}+D_{3}.
\end{aligned}
\end{equation*}
Calculate the three terms $D_{1},D_{2},D_{3}$ as below
\begin{equation*}
\begin{aligned}
D_{1}
&=E_{\pi_{t}} \left[( \mathbf{y}_{t}- \mathbf{z}_{t}) \frac{1}{2}\beta_{t} ( \mathbf{y}_{t}- \mathbf{z}_{t}) \right]\\
&\leq E_{\pi_{t}} \left[( \mathbf{y}_{t}- \mathbf{z}_{t}) L_{1}(t) ( \mathbf{y}_{t}- \mathbf{z}_{t}) \right]\\
&=E_{\pi_{t}} \left[( \mathbf{y}_{t}- \mathbf{z}_{t})^{2}\right] L_{1}(t)
=L_{1}(t)  W^{2}_{2}(q_{t},p_{t}).
\end{aligned}
\end{equation*}
This inequality reduction uses the drifting term assumption. 
\begin{equation*}
\begin{aligned}
D_{2}
&= \beta_{t} E_{\pi_{t}} \left[( \mathbf{y}_{t}- \mathbf{z}_{t}) \left( s_{\theta}(\mathbf{z}_{t},f_{\phi}(\mathbf{x}),t)-s_{\theta}(\mathbf{y}_{t},f_{\phi}(\mathbf{x}),t)+s_{\theta}(\mathbf{y}_{t},f_{\phi}(\mathbf{x}),t)-\frac{\partial }{\partial \mathbf{y}_{t}} \log q_{t} \right) \right]\\
&=\beta_{t} E_{\pi_{t}} \left[( \mathbf{y}_{t}- \mathbf{z}_{t}) \left( s_{\theta}(\mathbf{z}_{t},f_{\phi}(\mathbf{x}),t)-s_{\theta}(\mathbf{y}_{t},f_{\phi}(\mathbf{x}),t) \right)\right]
+\beta_{t} E_{\pi_{t}} \left[( \mathbf{y}_{t}- \mathbf{z}_{t}) \left( s_{\theta}(\mathbf{y}_{t},f_{\phi}(\mathbf{x}),t)-\frac{\partial }{\partial \mathbf{y}_{t}} \log q_{t} \right) \right]\\
& \leq \beta_{t} \left\{ L_{2}(t) E_{\pi_{t}} \left[ ( \mathbf{y}_{t}- \mathbf{z}_{t})^{2}\right]+ \left[ E_{\pi_{t}} \left[ ( \mathbf{y}_{t}- \mathbf{z}_{t})^{2} \right] \right]^{\frac{1}{2}}  \cdot \left[ E_{\pi_{t}} \left( s_{\theta}(\mathbf{y}_{t},f_{\phi}(\mathbf{x}),t)-\frac{\partial }{\partial \mathbf{y}_{t}} \log q_{t} \right)^{2} \right]^{\frac{1}{2}}    \right\}\\
&=\beta_{t} \left\{ L_{2}(t)W^{2}_{2}(q_{t},p_{t})+W_{2}(q_{t},p_{t}) \left[ E_{\pi_{t}} \left( s_{\theta}(\mathbf{y}_{t},f_{\phi}(\mathbf{x}),t)-\frac{\partial }{\partial \mathbf{y}_{t}} \log q_{t} \right)^{2} \right]^{\frac{1}{2}}    \right\}\\
&=W_{2}(q_{t},p_{t}) \beta_{t} \left\{ L_{2}(t)W_{2}(q_{t},p_{t}) + \left[ E_{\pi_{t}} \left( s_{\theta}(\mathbf{y}_{t},f_{\phi}(\mathbf{x}),t)-\frac{\partial }{\partial \mathbf{y}_{t}} \log q_{t} \right)^{2} \right]^{\frac{1}{2}}  \right\},
\end{aligned}
\end{equation*}
Let
\begin{equation}\label{K6}
H(t)=E_{\pi_{t}} \left( s_{\theta}(\mathbf{y}_{t},f_{\phi}(\mathbf{x}),t)-\frac{\partial }{\partial \mathbf{y}_{t}} \log q_{t} \right)^{2},
\end{equation}
then
\begin{equation*}
D_{2} \leq W_{2}(q_{t},p_{t}) \beta_{t} \left[ L_{2}(t)W_{2}(q_{t},p_{t}) + \sqrt{H(t)}  \right].
\end{equation*}
Here, the first inequality with respect to the reduction $D_{2}$ is obtained with score function estimator assumption and the Cauchy-Schwarz inequality. In Lemma 2 of \cite{kwon2022score}, we have shown $$E_{\pi_{t}} \left[ ( \mathbf{y}- \mathbf{z}) \left( \frac{\partial }{\partial \mathbf{z}} \log p_{t}(\mathbf{z})-\frac{\partial }{\partial \mathbf{y}} \log q_{t}(\mathbf{y}) \right)\right],$$ is nonpositive, which $q_{t}(\mathbf{y}),p_{t}(\mathbf{z})$ denote the Density function of $f_{\phi}(\mathbf{x})$ without conditions. We similarly have
\begin{equation}\label{K7}
D_{3}=E_{\pi_{t}} \left[ ( \mathbf{y}_{t}- \mathbf{z}_{t}) \left( \frac{\partial }{\partial \mathbf{z}_{t}} \log p_{t}-\frac{\partial }{\partial \mathbf{y}_{t}} \log q_{t} \right)\right]<0.
\end{equation}

\noindent According to the results calculated separately above, it can be obtained
\begin{equation}\label{K8}
\begin{aligned}
D_{1}+D_{2}+D_{3} & \leq L_{1}(t)  W^{2}_{2}(q_{t},p_{t}) + W_{2}(q_{t},p_{t}) \beta_{t} \left[ L_{2}(t)W_{2}(q_{t},p_{t}) + \sqrt{H(t)}  \right] + D_{3}\\
& \leq L_{1}(t)  W^{2}_{2}(q_{t},p_{t}) + W_{2}(q_{t},p_{t}) \beta_{t} \left[ L_{2}(t)W_{2}(q_{t},p_{t}) + \sqrt{H(t)} \right].
\end{aligned}
\end{equation}
Eq.(\ref{K4}) eventually becomes of the form Eq.(\ref{K5}) and Eq.(\ref{K8}), that is, 
\begin{equation*}
-W_{2}(q_{t},p_{t}) \left( \frac{\partial }{\partial t} W_{2}(q_{t},p_{t}) \right)
\leq L_{1}(t)  W^{2}_{2}(q_{t},p_{t}) + W_{2}(q_{t},p_{t}) \beta_{t} \left[ L_{2}(t)W_{2}(q_{t},p_{t}) + \sqrt{H(t)}  \right],
\end{equation*}
By the definition of the second-order Wasserstein distance Eq.(\ref{F1}), we know that $W_{2}(q_{t},p_{t})>0$, then
\begin{equation}\label{K9}
- \left( \frac{\partial }{\partial t} W_{2}(q_{t},p_{t}) \right)
\leq (L_{1}(t)+\beta_{t} L_{2}(t)) W_{2}(q_{t},p_{t}) \beta_{t} \sqrt{H(t)}.
\end{equation}

The result of Eq.(\ref{K9}) has been obtained by combining the optimal transport theory of Wasserstein distance of order n Eq.(\ref{K4}) with the corresponding theory of CARD. Next, we will analyze Eq.(\ref{K9}), and an upper bound on the convergence of the discrepancy between the original data distribution and the generated distribution is already anticipated.

The left side of the equation in Eq.(\ref{K9}) represents the partial derivative form of the 2-Wasserstein distance with respect to the variable $t$, while the first term on the right side represents the multiple form of this distance. We should consider whether it is feasible to determine the original function from a function that contains the 2-Wasserstein distance. If we have
\begin{equation}\label{K10}
M(t)=\exp \left\{  \int^{t}_{0} (L_{1}(s)+\beta_{s} L_{2}(s))ds \right\},
\end{equation}
then
\begin{equation*}
\frac{\partial}{\partial t}M(t)=\frac{d}{d t}M(t)=(L_{1}(t)+\beta_{t} L_{2}(t))M(t).
\end{equation*}
Therefore, by multiplying both sides of Eq.(\ref{K9}) with $M(t)$ and shifting the term, we can derive Eq.(\ref{K10}) which can be further simplified along with its derivative
%\begin{equation*}
%-M(t)\frac{\partial}{\partial t}W_{2}(q_{t},p_{t})-(L_{1}%(t)+\beta_{t} L_{2}(t))M(t)W_{2}(q_{t},p_{t}) \leq \beta_{t} %\sqrt{H(t)} M(t),
%\end{equation*}
\begin{equation*}
-M(t)\frac{\partial}{\partial t}W_{2}(q_{t},p_{t})-W_{2}(q_{t},p_{t})\frac{\partial}{\partial t}M(t) \leq \beta_{t} \sqrt{H(t)} M(t),
\end{equation*}
that is
\begin{equation*}
-\frac{\partial}{\partial t} (M(t) W _{2}(q_{t},p_{t})) \leq \beta_{t} \sqrt{H(t)} M(t).
\end{equation*}
We integrate both sides of the inequality with respect to $t$ at $0,T$, and obtain
\begin{equation*}
 \int^{T}_{0}  -\frac{\partial}{\partial t}(M(t) W _{2}(q_{t},p_{t})) dt \leq \int^{T}_{0} \beta_{t} \sqrt{H(t)} M(t) dt.
\end{equation*}
From the definition of $M(t)$ in Eq.(\ref{K10}), we know that $M(0)=1$, so the above integral calculation result is
\begin{equation*}
 W_{2}(q_{0},p_{0})  \leq \int^{T}_{0} \beta_{t} \sqrt{H(t)} M(t) dt + M(T) W_{2}(q_{T},p_{T}),
\end{equation*}
Completely written as
\begin{equation*}
\begin{aligned}
W_{2}(q(\mathbf{y}_{0}|f_{\phi}(\mathbf{x})),p(\mathbf{z}_{0}|f_{\phi}(\mathbf{x})))  &\leq \int^{T}_{0} \beta_{t} \sqrt{H(t)} M(t) dt
+ M(T) W_{2}(q(\mathbf{y}_{T}|f_{\phi}(\mathbf{x})),p(\mathbf{z}_{T}|f_{\phi}(\mathbf{x}))) ,
\end{aligned}
\end{equation*}
More generally, 
\begin{equation*}
\begin{aligned}
W_{2}(q(\mathbf{y}_{0}|f_{\phi}(\mathbf{x})),p(\mathbf{y}_{0}|f_{\phi}(\mathbf{x})))  &\leq \int^{T}_{0} \beta_{t} \sqrt{H(t)} M(t) dt
+ M(T) W_{2}(q(\mathbf{y}_{T}|f_{\phi}(\mathbf{x})),p(\mathbf{y}_{T}|f_{\phi}(\mathbf{x}))).
\end{aligned}
\end{equation*}
The proof is complete. 

\subsection{Proof of Corollary \ref{cor-1}}\label{appendixA4}
Based on Theorem \ref{the-1}, we introduce the loss function for the model and derive a novel convergence upper bound. Following the principles of score-based generative models (SGMs), the loss of CARD is defined as the squared difference between the score function and model output. Consequently, the loss function for CARD, denoted as $L_{1}(\phi,\theta,\lambda)$, is specified in Eq.(\ref{F11}).
\begin{equation*}
\begin{aligned}
L_{1}(\phi,\theta,\lambda) &= \frac{1}{2} \int^{T}_{0} \lambda_{t} \textup{E}_{q(\mathbf{y}_{t}|f_{\phi}(\mathbf{x}))} \left[ ||s_{\theta}(\mathbf{y}_{t},f_{\phi}(\mathbf{x}),t) - \frac{\partial \log q(\mathbf{y}_{t}|f_{\phi}(\mathbf{x}))} { \partial \mathbf{y}_{t}}  ||^{2} \right] dt
=\frac{1}{2}\int^{T}_{0}  H(t) \lambda(t) dt.
\end{aligned}
\end{equation*}

According to the conclusion given in Eq.(\ref{F10}), the loss function $L_{1}(\phi,\theta,\lambda)$ is defined in Eq.(\ref{K10}) within the context of $H(t)$. Due to the presence of a common the integral term across these expressions, the first term on the right-hand side of Eq.(\ref{F10}) requires simplification. Utilizing the Cauchy-Schwarz inequality, we can proceed with the simplification as follows.
\begin{equation*}
\begin{aligned}
\int^{T}_{0} \beta_{t} \sqrt{H(t)} M(t) dt &\leq \left( \int^{T}_{0} 2\beta_{t}^{2} M^{2}(t) \lambda^{-1}(t)  dt \right)^{\frac{1}{2}}  \left( \int^{T}_{0} \frac{1}{2} H(t) \lambda(t) dt \right)^{\frac{1}{2}}\\
& = \left( \int^{T}_{0} 2\beta_{t}^{2} M^{2}(t) \lambda^{-1}(t)  dt \right)^{\frac{1}{2}} \sqrt{J_{1}(\phi,\theta,\lambda)}\\
& = \sqrt{2\left( \int^{T}_{0} \beta_{t}^{2} M^{2}(t) \lambda^{-1}(t)  dt \right) L_{1}(\phi,\theta,\lambda)}.
\end{aligned}
\end{equation*}
If for all $t \in (0,T)$, $\lambda(t)=\beta_{t}$, then 
\begin{equation}\label{K11}
\begin{aligned}
\int^{T}_{0} \beta_{t} \sqrt{H(t)} M(t) dt &\leq \sqrt{2\left( \int^{T}_{0} \beta_{t} M^{2}(t)  dt \right) J_{1}(\phi,\theta,\lambda)}.
\end{aligned}
\end{equation}
If $\lambda(t)$ is set in another way, 
\begin{equation}\label{K12}
\begin{aligned}
\int^{T}_{0} \beta_{t} \sqrt{H(t)} M(t) dt &\leq  \sqrt{2\left( \int^{T}_{0} \beta_{t}^{2} M^{2}(t) \lambda^{-1}(t)  dt \right) J_{1}(\phi,\theta,\lambda)}.
\end{aligned}
\end{equation}

Therefore under the assumption of $\lambda(t)=\beta_{t},\forall t \in (0,T)$, according to type Eq.(\ref{K11}) as a result, bringing it to type Eq.(\ref{F10}) of the available Eq.(\ref{F12}).
If we make no special assumptions on $\lambda(t)$, then Eq.(\ref{K12}) can be obtained from Eq.(\ref{F13}). 

The proof is complete.

\newpage
\section{Proof of Theorem \ref{the-2}}\label{appendixB}
The following will present a detailed proof of the above conclusion for reference.
Our aim is to ascertain the error between the true value and the network estimate, considering the simplification of this term $H(t)$ in light of the truncation strategy within the above-mentioned assumptions. The preliminary simplification results are presented as follows. 
\begin{equation*}
\begin{aligned}
    H(t) 
    &= \textup{E}_{q(\mathbf{y}_{t}|f_{\phi}(\mathbf{x}))} \left[ \left\|s_{\theta}(\mathbf{y}_{t},f_{\phi}(\mathbf{x}),t) - \frac{\partial \log q(\mathbf{y}_{t}|f_{\phi}(\mathbf{x}))} { \partial \mathbf{y}_{t}}  \right\|^{2}_{2} \right]\\
    &= \int_{\mathbb{R}^{d}} \left\| s_{\theta}(\mathbf{y}_{t},f_{\phi}(\mathbf{x}),t)-\nabla \log q(\mathbf{y}_{t}|f_{\phi}(\mathbf{x})) \right\| ^{2}_{2} \ q(\mathbf{y}_{t}|f_{\phi}(\mathbf{x})) \ d \mathbf{y}_{t}\\
    &= \int_{\left\| \mathbf{y}_{t} \right\|_{\infty} >R} \left\| s_{\theta}(\mathbf{y}_{t},f_{\phi}(\mathbf{x}),t)-\nabla \log q(\mathbf{y}_{t}|f_{\phi}(\mathbf{x})) \right\| ^{2}_{2} \ q(\mathbf{y}_{t}|f_{\phi}(\mathbf{x})) \ d \mathbf{y}_{t}\\
    &\ + \int_{\left\| \mathbf{y}_{t} \right\|_{\infty} \leq R} \left\| s_{\theta}(\mathbf{y}_{t},f_{\phi}(\mathbf{x}),t)-\nabla \log q(\mathbf{y}_{t}|f_{\phi}(\mathbf{x})) \right\| ^{2}_{2} \ q(\mathbf{y}_{t}|f_{\phi}(\mathbf{x})) \  \mathbf{1}_{\left\{ \left|q(\mathbf{y}_{t}|f_{\phi}(\mathbf{x}))\right| < \varepsilon_{low} \right\} } \ d \mathbf{y}_{t}\\
    &\ + \int_{\left\| \mathbf{y}_{t} \right\|_{\infty} \leq R} \left\| s_{\theta}(\mathbf{y}_{t},f_{\phi}(\mathbf{x}),t)-\nabla \log q(\mathbf{y}_{t}|f_{\phi}(\mathbf{x})) \right\| ^{2}_{2} \ q(\mathbf{y}_{t}|f_{\phi}(\mathbf{x})) \  \mathbf{1}_{\left\{ \left|q(\mathbf{y}_{t}|f_{\phi}(\mathbf{x}))\right| \geq \varepsilon_{low} \right\} } \ d \mathbf{y}_{t}\\
    &= H_{1} + H_{2} +H_{3}.
\end{aligned}
\end{equation*}

Based on the above preliminary simplification results, when we intend to determine the error bounds between the true value and the network estimate, these error bounds can be estimated respectively from the three terms in the truncation process. Firstly, the truncation error ($H_{1}$) is caused by the unbounded range of $\mathbf{y}_{t}$. Secondly, $H_{2}$ is induced by the overly small conditional density function $q(\mathbf{y}_{t}|f_{\phi}(\mathbf{x}))$. Finally, the approximation error ($H_{3}$) under the above two truncation strategies.

Through a series of algebraic operations and the upper bound forms within the aforementioned assumptions, we initially achieved a preliminary treatment of the problem. 
\begin{equation*}
    \begin{aligned}
        H_{1} & \leq 2\int_{\left\| \mathbf{y}_{t} \right\|_{\infty} >R} \left\| s_{\theta}(\mathbf{y}_{t},f_{\phi}(\mathbf{x}),t) \right\| ^{2}_{2} \ q(\mathbf{y}_{t}|f_{\phi}(\mathbf{x})) \ d \mathbf{y}_{t} + 2\int_{\left\| \mathbf{y}_{t} \right\|_{\infty} >R} \left\|\nabla \log q(\mathbf{y}_{t}|f_{\phi}(\mathbf{x})) \right\| ^{2}_{2} \ q(\mathbf{y}_{t}|f_{\phi}(\mathbf{x})) \ d \mathbf{y}_{t}\\
        & \leq 2d \int_{\left\| \mathbf{y}_{t} \right\|_{\infty} >R} \left\| s_{\theta}(\mathbf{y}_{t},f_{\phi}(\mathbf{x}),t) \right\| ^{2}_{\infty} \ q(\mathbf{y}_{t}|f_{\phi}(\mathbf{x})) \ d \mathbf{y}_{t} + 2\int_{\left\| \mathbf{y}_{t} \right\|_{\infty} >R} \left\|\nabla \log q(\mathbf{y}_{t}|f_{\phi}(\mathbf{x})) \right\| ^{2}_{2} \ q(\mathbf{y}_{t}|f_{\phi}(\mathbf{x})) \ d \mathbf{y}_{t}
    \end{aligned}
\end{equation*}
where the first inequality involves the Cauchy-Schwarz inequality, and the second inequality applies the relationship between the $L_2$ norm and $L_{\infty}$ norm, that is, $\left\| s_{\theta}(\mathbf{y}_{t},f_{\phi}(\mathbf{x}),t) \right\| ^{2}_{2} \leq d \left\| s_{\theta}(\mathbf{y}_{t},f_{\phi}(\mathbf{x}),t) \right\| ^{2}_{\infty}$.
Analogously, we have
\begin{equation*}
    \begin{aligned}
        H_{2} & \leq \int_{\left\| \mathbf{y}_{t} \right\|_{\infty} \leq R} 2\left( d \left\| s_{\theta}(\mathbf{y}_{t},f_{\phi}(\mathbf{x}),t) \right\| ^{2}_{\infty} + \left\| \nabla \log q(\mathbf{y}_{t}|f_{\phi}(\mathbf{x})) \right\| ^{2}_{2}\right) \ q(\mathbf{y}_{t}|f_{\phi}(\mathbf{x})) \  \mathbf{1}_{\left\{ \left|q(\mathbf{y}_{t}|f_{\phi}(\mathbf{x}))\right| < \varepsilon_{low} \right\} } \ d \mathbf{y}_{t}.
    \end{aligned}
\end{equation*}
\begin{equation*}
    \begin{aligned}
        H_{3} & \leq d \int_{\left\| \mathbf{y}_{t} \right\|_{\infty} \leq R} \frac{\left\| s_{\theta}(\mathbf{y}_{t},f_{\phi}(\mathbf{x}),t)-\nabla \log q(\mathbf{y}_{t}|f_{\phi}(\mathbf{x})) \right\| ^{2}_{\infty} \ q^{2}(\mathbf{y}_{t}|f_{\phi}(\mathbf{x}))}{q(\mathbf{y}_{t}|f_{\phi}(\mathbf{x}))} \  \mathbf{1}_{\left\{ \left|q(\mathbf{y}_{t}|f_{\phi}(\mathbf{x}))\right| \geq \varepsilon_{low} \right\} } \ d \mathbf{y}_{t}.
    \end{aligned}
\end{equation*}
% Among them, $H_{1}$ is primarily dedicated to determining the upper bounds corresponding to the conditional density $q(\mathbf{y}_{t}|f_{\phi}(\mathbf{x}))$ and the gradient of the logarithmic density $\nabla \log q(\mathbf{y}_{t}|f_{\phi}(\mathbf{x}))$ respectively when $\mathbf{y}_{t} \in \overline{D}_{1}$. Then, analogously, $H_{2}$ aims to ascertain the upper bounds corresponding to the conditional density and the gradient of the logarithmic density respectively when $\mathbf{y}_{t} \in D_{1} \cap \overline{D}_{2}$. Subsequently, regarding $H_{3}$, it focuses on the error bound between the true value and the estimated value under the condition that two truncation conditions are satisfied, that is $\mathbf{y}_{t} \in D_{1} \cap D_{2}$. During this process, inspired by XXX, we then consider employing local polynomials to conduct approximate expansion and estimation. Details of the proof are shown later. 

It should be emphasized that, in the context of the results of the preliminary simplification, if we postulate that $\left\| s_{\theta}(\mathbf{y}_{t},f_{\phi}(\mathbf{x}),t) \right\| ^{2}_{\infty}$ has an upper bound $K_{t}$, then the solution of $H_{1}$ and $H_{2}$ is transmuted into determining the bounds of $ q(\mathbf{y}_{t}|f_{\phi}(\mathbf{x}))$ and $\nabla \log q(\mathbf{y}_{t}|f_{\phi}(\mathbf{x}))$, respectively, under the circumstances of $D_1$ and $D_{1} \cap \overline{D}_{2}$. $H_{3}$ is to determine the upper bound of $\left\| s_{\theta}(\mathbf{y}_{t},f_{\phi}(\mathbf{x}),t)-\nabla \log q(\mathbf{y}_{t}|f_{\phi}(\mathbf{x})) \right\| _{\infty} \ q(\mathbf{y}_{t}|f_{\phi}(\mathbf{x}))$ under the condition of complying with the two truncation strategies $D_{1} \cap D_{2}$. Inspired by \cite{fu2024unveil}, we will employ the local diffusion polynomial for approximate expansion and estimation.

After rigorous calculation, $H_{1},H_{2},H_{3}$ can be respectively 
derived under the corollary of Lemma \ref{lem-4}, Lemma \ref{lem-5} and Lemma \ref{lem-6}, and the specific form are presented as follows
\begin{equation*}
\begin{aligned}
H_{1}  &\leq \frac{2 d c_1 R \left[K_t \sigma_{t}^{6} +c_4^{2} \left( R^{2} + 6(1-\gamma_{t})^{2} R_{f}^{2}  \right)\right]}{(\gamma_{t}^{2}+c_2 \sigma_{t}^{2})^{\frac{d}{2}} \sigma_{t}^{6} },\\
% \noindent
H_{2}  
% \leq  2^{d+1} d R^{d} \varepsilon_{low} K_t
% + \frac{1}{\sigma_{t}^{6}} \left[ 2^{d+1} d c_4^{2} R^{d} \varepsilon_{low} \left( R^{2} + 6(1-\gamma_{t})^{2} R_{f}^{2}  \right) \right].
&\leq 2^{d+1} d R^{d} \varepsilon_{low} \left[ K_t + \frac{c_4^{2}}{\sigma_{t}^{6}} \left( R^{2} + 6(1-\gamma_{t})^{2} R_{f}^{2}  \right) \right],\\
% \noindent
H_{3}  
& \leq  d R^{d} \left[   \frac{1}{\sigma_{t}}\left( \varepsilon 
+ \frac{c BR^{s}_{*} (d+d_x)^{s} \left(\log \varepsilon^{-1}\right)^{\frac{1}{2}}}{N^{\beta} s! \gamma_{t}^{\frac{d}{2}}} 
+ \frac{c B R_{*}^{s+d} \varepsilon^{\frac{3}{e}} \left(\log \varepsilon^{-1}\right)^{\frac{1}{2}} }{N^d \sigma_{t}^{d} (2 \pi)^{\frac{d}{2}}} \left(1+\frac{ (d+d_x)^{s}}{N^{\beta} s!}  \right)  \right)\right.\\
&\ \ \ +\left.\left(\frac{c_4}{\sigma_{t}^{3}} \left( R^{2} + 2(1-\gamma_{t})^{2} R_{f}^{2}  \right)^{\frac{1}{2}} + c_5 \right)
\left( \varepsilon + \frac{BR^{s}_{*} (d+d_x)^{s}}{N^{\beta} s! \gamma_{t}^{\frac{d}{2}}} + \frac{d B R_{*}^{s+d} \varepsilon^{\frac{3}{e}}}{N^d \sigma_{t}^{d} (2 \pi)^{\frac{d}{2}}} \left(1+\frac{ (d+d_x)^{s}}{N^{\beta} s!} \right) \right) \right].
\end{aligned}
\end{equation*}
Therefore we have
\begin{equation*}
\begin{aligned}
&\ \ \ \int_{\mathbb{R}^{d}} \left\| s_{\theta}(\mathbf{y}_{t},f_{\phi}(\mathbf{x}),t)-\nabla \log q(\mathbf{y}_{t}|f_{\phi}(\mathbf{x})) \right\| ^{2}_{2} \ q(\mathbf{y}_{t}|f_{\phi}(\mathbf{x})) \ d \mathbf{y}_{t}\\
&\lesssim \frac{K_t \sigma_{t}^{6} +c_4^{2} \left( R^{2} + 6(1-\gamma_{t})^{2} R_{f}^{2}  \right)}{(\gamma_{t}^{2}+c_2 \sigma_{t}^{2})^{\frac{d}{2}} \sigma_{t}^{6} }
+ \varepsilon_{low} \left[ K_t + \frac{c_4^{2}}{\sigma_{t}^{6}} \left( R^{2} + 6(1-\gamma_{t})^{2} R_{f}^{2}  \right) \right]\\
&+\left[  \frac{1}{\sigma_{t}} \left(\varepsilon 
+ \frac{B \left(\log \varepsilon^{-1}\right)^{\frac{1}{2}}}{N^{\beta} \gamma_{t}^{\frac{d}{2}}} 
+ \frac{ B \varepsilon^{\frac{3}{e}} \left(\log \varepsilon^{-1}\right)^{\frac{1}{2}} (N^{\beta}+1)}{N^{d+\beta} \sigma_{t}^{d}} \right) 
+ \left(\frac{1}{\sigma_{t}^{3}} \left( R^{2} + 2(1-\gamma_{t})^{2} R_{f}^{2}  \right)^{\frac{1}{2}} + c_5 \right)   
\left( \varepsilon + \frac{B}{N^{\beta}  \gamma_{t}^{\frac{d}{2}}} + \frac{B \varepsilon^{\frac{3}{e}} (N^{\beta}+1)}{N^{d+\beta} \sigma_{t}^{d}}\right)
\right].    
\end{aligned}    
\end{equation*}
The details of the proof will be presented hereafter.

\subsection{Gaussian representation}
We know that $q\left(\mathbf{y} \mid \mathbf{y}_0,f_\phi(\boldsymbol{x})\right)=N\left(\sqrt{\bar{\alpha}_t} \mathbf{y}_0+\left(1-\sqrt{\bar{\alpha}_t}\right) f_\phi(\mathbf{x}),\left(1-\bar{\alpha}_t\right) \mathbf{I}\right),\ \bar{\alpha}_t= \prod_s^t \alpha_s= \prod_s^t (1-\beta_s)$ is true in the original diffusion process of CARD. If we commence with the known stochastic differential equations to conduct a backward derivation of the corresponding Gaussian distribution form,  this will facilitate the subsequent analysis. Actually, this constitutes a process of performing a reverse reconstruction of the probability distribution. The detailed derivation process is presented below.

According to Eq.(\ref{E1}), we have the SDE form of the forward process, 
\begin{equation*}
d\mathbf{y} = -\frac{\beta}{2}(\mathbf{y}-f_{\phi}(\mathbf{x}))dt +\sqrt{\beta}dw,\ \ dw \sim N(0,\Delta t).
\end{equation*}
where the variable $\mathbf{y}_{t}$ at each of these steps is simplified to be represented by $\mathbf{y}$. If there is a function $\mu(t)$ satisfying $\mu'(t) = \frac{\beta}{2} \mu(t)$,
then we can deduce that $\mu(t) = e^{\int \frac{\beta}{2} dt} = e^{\frac{\beta}{2} t}$.
Multiply both sides of Eq. (\ref{E1}) by $\mu(t)$ and simplified as
\begin{equation*}
d \left( e^{\frac{\beta}{2} t} (\mathbf{y}-f_{\phi}(\mathbf{x})) \right) = e^{\frac{\beta}{2} t} \sqrt{\beta_{t}}dw,
\end{equation*}
Integrating both sides of $t$, we end up with a specific expression for the random variable $\mathbf{y}$,
\begin{equation*}
    \mathbf{y} = e^{-\frac{\beta}{2} t} (\mathbf{y}_{0}-f_{\phi}(\mathbf{x})) + \sqrt{\beta_{t}} \int_{0}^{t} e^{-\frac{\beta}{2} (t-s)} dw + f_{\phi}(\mathbf{x}).
\end{equation*}

Obviously, when $\mathbf{y}$ follows a Gaussian distribution, its corresponding mean and variance are respectively
\begin{equation}\label{BB1}
\textup{E}(\mathbf{y}) = e^{-\frac{\beta}{2} t} (\mathbf{y}_{0}-f_{\phi}(\mathbf{x})) + \sqrt{\beta_{t}} \int_{0}^{t} e^{-\frac{\beta}{2} (t-s)} \textup{E}(d w) + f_{\phi}(\mathbf{x})
= e^{-\frac{\beta}{2} t} \mathbf{y}_{0} + \left(1-e^{-\frac{\beta}{2}} \right) f_{\phi}(\mathbf{x}).
\end{equation}
In addition, another concise method also could calculate the mean. If we denote $u$ as the mean value of $\mathbf{y}$, that is $u = \textup{E}(\mathbf{y})$, then
$d u =\textup{E}\left( -\beta(u - f_{\phi}(\mathbf{x}))d t/2 \right)$ and $du/dt = -\beta(u - f_{\phi}(\mathbf{x}))/2$. After calculation, Eq.(\ref{BB1}) is also established. 

When calculating the variance, given that both the initial input $\mathbf{y}_{0}$ and its transformation $f_{\phi}(\mathbf{x})$ are known, we only concentrate on the random terms.
\begin{equation}\label{BB2}
    \begin{aligned}
        Var(\mathbf{y}) &= Var\left(\sqrt{\beta_{t}} \int_{0}^{t} e^{-\frac{\beta}{2} (t-s)} dw \right)
        = \beta \textup{E}\left[ \int_{0}^{t} e^{-\frac{\beta}{2} (t-s)} dw - \textup{E}\left( \int_{0}^{t} e^{-\frac{\beta}{2} (t-s)} dw  \right)  \right]^{2}\\
        & = \beta \textup{E}\left( \int_{0}^{t} e^{-\frac{\beta}{2} (t-s)} dw \right)^{2}
        = \beta  \int_{0}^{t} e^{-\beta (t-s)} ds\\
        & = (1-e^{- \beta t})\mathbf{I}.
    \end{aligned}
\end{equation}

Therefore, we can write the conditional distribution in the diffusion process 
\begin{equation*}
    q(\mathbf{y}|\mathbf{y}_{0},f_{\phi}(\mathbf{x})) = N\left(e^{-\frac{\beta}{2} t} \mathbf{y}_{0} + \left(1-e^{-\frac{\beta}{2}} \right) f_{\phi}(\mathbf{x}),\ (1-e^{- \beta t}) \mathbf{I}\right),
\end{equation*} 
For shortness, we define $\gamma_{t} = e^{-\frac{\beta}{2} t},\  \sigma^{2}_{t} = 1-e^{- \beta t}$, that is 
\begin{equation}\label{BB3}
    \mathbf{y} \sim q(\mathbf{y}|\mathbf{y}_{0},f_{\phi}(\mathbf{x})) = N(\gamma_{t} \mathbf{y}_{0} + \left(1-\gamma_{t} \right) f_{\phi}(\mathbf{x}),\ \sigma^{2}_{t}\mathbf{I}).
\end{equation}

\subsection{Understanding of truncation radius R}
According to the setting in Assumption \ref{assum8}, we select the radius $R$ of the $l_{\infty}$ ball to truncate the variable $\mathbf{y}_{t}$. A more general definition is that for given center $\mathbf{a}=(a_{1},a_{2},\cdots,a_{n})$ and radius R, $l_{\infty}$ ball means $B_{\infty}(\mathbf{a},R)= \{ \mathbf{y}_{t}=\{\mathbf{y}_{t}^{1},\mathbf{y}_{t}^{2},\cdots,\mathbf{y}_{t}^{n}\}:\left\|\mathbf{y}_{t}- \mathbf{a}\right\|_{\infty} \leq R \}$.
This implies that within the truncation region, the $L_{\infty}$ distance between any variable and the origin does not exceed $R$.

With particular focus on the characteristics of Gaussian functions within the light tail hypothesis above, we consider the value range of $R$. First, when we calculate the conditional density at any time, we have we have the following integral formula for the original variable $\mathbf{y}_{0}$.
\begin{equation*}
\begin{aligned}
    q(\mathbf{y}_{t}|f_{\phi}(\mathbf{x})) & = \int_{\mathbb{R}^{d}} q(\mathbf{y}_{t},\mathbf{y}_{0}|f_{\phi}(\mathbf{x}))  d \mathbf{y}_{0}
    = \int_{\mathbb{R}^{d}} q(\mathbf{y}_{t}|\mathbf{y}_{0},f_{\phi}(\mathbf{x})) q(\mathbf{y}_{0}|f_{\phi}(\mathbf{x})) d \mathbf{y}_{0}\\
    & = \frac{1}{\sigma_{t}^{d} (2 \pi)^{\frac{d}{2}}} \int_{\mathbb{R}^{d}}   
    q(\mathbf{y}_{0}|f_{\phi}(\mathbf{x}))      
    \exp {\left\{-\frac{\left\| \mathbf{y}_{t}-\left[ \gamma_{t} \mathbf{y}_{0} + \left(1-\gamma_{t} \right) f_{\phi}(\mathbf{x}) \right] \right\|^{2}}{2 \sigma_{t}^{2}} \right\}} d \mathbf{y}_{0}.
\end{aligned}
\end{equation*}

Based on the light tail assumption of the original conditional density in Assumption \ref{assum6} and the truncation strategy for variables in Assumption \ref{assum8}, we anticipate that the truncation error of the conditional density can be effectively controlled outside the truncation region.

In other words, given that the original conditional density $q(\mathbf{y}_{0}|f_{\phi}(\mathbf{x}))$ is manageable, our focus shifts to how the exponential terms of the Gaussian function vary. 
Since assuming $\mathbf{y}_{0}$ is like the tail probability in Assumption \ref{assum6}, consider the sub Gaussian tail probability bound. There exists constant $c$ depending on the dimension $d$ and $\varepsilon \in (0,\frac{1}{e})$, so that $P(\left\| \mathbf{y}_{0} \right\|_{\infty} \geq R) \leq 2 \exp{\left( -\frac{R^{2}}{c^{2}} \right)}$. We obtain $R \geq c\sqrt{\log \varepsilon^{-1}}$ and denote 
\begin{equation}\label{D01}
D_{01}=\left[-c\sqrt{\log \varepsilon^{-1}}, c\sqrt{\log \varepsilon^{-1}}  \right].
\end{equation}

Meanwhile, for the exponential terms in the above integral, if there exists a sufficiently large constant $R$ and a sufficiently small constant $\varepsilon$ such that for $\left\| \mathbf{y}_{t}-\left[ \gamma_{t} \mathbf{y}_{0} + \left(1-\gamma_{t} \right) f_{\phi}(\mathbf{x}) \right] \right\|^{2}/2 \sigma_{t}^{2} > R$, we have $\exp {\left\{-\left\| \mathbf{y}_{t}-\left[ \gamma_{t} \mathbf{y}_{0} + \left(1-\gamma_{t} \right) f_{\phi}(\mathbf{x}) \right] \right\|^{2}/2 \sigma_{t}^{2} \right\}} \leq \varepsilon$, this implies that the integral error outside the truncation region is small and controllable. 
Through simplification, we obtain and denote 
\begin{equation}\label{D02}
D_{02}=\left[ \frac{\mathbf{y}_{t}- \left(1-\gamma_{t} \right) f_{\phi}(\mathbf{x})-\sigma_{t} c \sqrt{\log \varepsilon^{-1}}}{\gamma_{t}},\ \frac{\mathbf{y}_{t}- \left(1-\gamma_{t} \right) f_{\phi}(\mathbf{x})+\sigma_{t} c \sqrt{\log \varepsilon^{-1}}}{\gamma_{t}}  \right].
\end{equation}
if and only if the equal sign is established. 
Ultimately, we redefine the truncation domain as 
\begin{equation*}
\begin{aligned}
 &\mathbf{y}_{0} \in D_{0}=D_{01}\cap D_{02}\\
&=\left[-c\sqrt{\log \varepsilon^{-1}}, c\sqrt{\log \varepsilon^{-1}}  \right]
\cap \left[ \frac{\mathbf{y}_{t}- \left(1-\gamma_{t} \right) f_{\phi}(\mathbf{x})-\sigma_{t} c\sqrt{\log \varepsilon^{-1}}}{\gamma_{t}},\ \frac{\mathbf{y}_{t}- \left(1-\gamma_{t} \right) f_{\phi}(\mathbf{x})+\sigma_{t} c\sqrt{\log \varepsilon^{-1}}}{\gamma_{t}}  \right].
\end{aligned}
\end{equation*}

According to the above analysis, we also could control the error of conditional density $q(\mathbf{y}_{t}|f_{\phi}(\mathbf{x}))$ outside the truncation domain.

\begin{assumption}\label{assum10}
\textbf{(Clip the integral.)} Under above Assumptions, for any $\mathbf{y}_{0} \in \overline{D}_{0}$ and $0 \leq \varepsilon \leq \frac{1}{e}$, condition density $q(\mathbf{y}_{t}|f_{\phi}(\mathbf{x}))$ holds that
\begin{equation}\label{BB4}
    \begin{aligned}
       &\int_{\overline{D}_{0}} q(\mathbf{y}_{0}|f_{\phi}(\mathbf{x})) \phi(\mathbf{y}_{t}|\mathbf{y}_{0},f_{\phi}(\mathbf{x})) d \mathbf{y}_{0} \leq \varepsilon.
    \end{aligned}
\end{equation}
Meanwhile, for any $\mathbf{v} \in \mathbb{Z}_{+}^{d}$, with $\left\| \mathbf{v} \right\|_{1} \leq n$, the gradient term $\nabla q(\mathbf{y}_{t}|f_{\phi}(\mathbf{x}))$ is similar as   %the first coordinate of
\begin{equation}\label{BB5}
\begin{aligned}
       &\int_{\overline{D}_{0}} \left| \left(\frac{\mathbf{y}_{t}-\left( \gamma_{t} \mathbf{y}_{0}+(1-\gamma_{t}) f_{\phi}(\mathbf{x}) \right)}{\gamma_{t}^{2}}  \right)^{\mathbf{v}} \right| 
       q(\mathbf{y}_{0}|f_{\phi}(\mathbf{x})) \phi(\mathbf{y}_{t}|\mathbf{y}_{0},f_{\phi}(\mathbf{x})) d \mathbf{y}_{0} \leq \varepsilon.
\end{aligned}
\end{equation}
where $\phi$ means the known Gaussian function and
\begin{equation*}
    \left(\frac{\mathbf{y}_{t}-\left( \gamma_{t} \mathbf{y}_{0}+(1-\gamma_{t}) f_{\phi}(\mathbf{x}) \right)}{\gamma_{t}^{2}}  \right)^{\mathbf{v}}  
    = \prod\limits_{i=1}^{d}  \left(\frac{\mathbf{y}_{ti}-\left( \gamma_{t} \mathbf{y}_{0i}+(1-\gamma_{t}) f_{\phi}(\mathbf{x}) \right)}{\gamma_{t}^{2}}  \right)^{\mathbf{v}_{i}}.
\end{equation*}    
\end{assumption}

For any $d$ dimensional non negative integer vector $\mathbf{v}$ with the sum of its components being less than or equal to the sample size $n$\ $(\left\| \mathbf{v} \right\|_{1} = \sum_{i=1}^{d} \left| v_{i} \right| \leq n)$, it is introduced for the purpose of taking into account the partial derivatives of the conditional density $q(\mathbf{y}_{0}|f_{\phi}(\mathbf{x}))$ when solving the integral, thereby making the formula more general.

\subsection{Upper Bounds}
\subsubsection{Conditional density upper and lower bounds}
\begin{lemma}\label{lem-1}
Under Assumption \ref{assum6} with respect to the light tailed property of the original conditional density $q(\mathbf{y}_{0}|f_{\phi}(\mathbf{x}))$ and truncation strategy of variable $variable$, there exists a constant $c_3>0$ such that the diffused density function $q(\mathbf{y}_{t}|f_{\phi}(\mathbf{x}))$ can be bounded as:
\begin{equation}\label{BB6}
    \frac{c_3}{\sigma_{t}^{d}} 
    \exp {\left\{-\frac{\left\| \mathbf{y}_{t}\right\|^{2} 
    + 2 \left(1-\gamma_{t} \right)^{2} \left\|f_{\phi}(\mathbf{x})  \right\|^{2} + 2\gamma_{t}^{2} R^{2}}{\sigma_{t}^{2}} \right\}} 
    \leq q(\mathbf{y}_{t}|f_{\phi}(\mathbf{x})) 
    \leq \frac{c_1}{\left( \gamma_{t}^{2}+c_2 \sigma_{t}^{2} \right)^{\frac{d}{2}}} \exp{\left\{- \frac{c_2 \left\|\mathbf{y}_{t}-(1-\gamma_{t})f_{\phi}(\mathbf{x}))   \right\|^{2}}{2 (\gamma_{t}^{2}+c_2 \sigma_{t}^{2})}    \right\}}.
\end{equation}
\end{lemma}

\noindent \textbf{Proof.} Based on the above mentioned assumptions, especially considering the light tailed property of the original conditional distribution in Assumption \ref{assum6} and the known Gaussian transfer kernel representation, we are able to gradually derive the upper and lower bounds of the conditional density $q(\mathbf{y}_{t}|f_{\phi}(\mathbf{x}))$ during each step within the diffusion process.

First, we derive the upper bound form of the conditional density function.
\begin{equation*}
\begin{aligned}
    q(\mathbf{y}_{t}|f_{\phi}(\mathbf{x})) 
    & = \int_{\mathbb{R}^{d}} q(\mathbf{y}_{t}|\mathbf{y}_{0},f_{\phi}(\mathbf{x})) q(\mathbf{y}_{0}|f_{\phi}(\mathbf{x})) d \mathbf{y}_{0}\\
    & = \frac{1}{\sigma_{t}^{d} (2 \pi)^{\frac{d}{2}}} \int_{\mathbb{R}^{d}}   
    q(\mathbf{y}_{0}|f_{\phi}(\mathbf{x}))      
    \exp {\left\{-\frac{\left\| \mathbf{y}_{t}-\left[ \gamma_{t} \mathbf{y}_{0} + \left(1-\gamma_{t} \right) f_{\phi}(\mathbf{x}) \right] \right\|^{2}}{2 \sigma_{t}^{2}} \right\}} d \mathbf{y}_{0}\\
    & \leq \frac{c_{1}}{\sigma_{t}^{d} (2 \pi)^{\frac{d}{2}}} \int_{\mathbb{R}^{d}}   
    \exp{\left\{- \frac{c_2 \left\| \mathbf{y}_{0} \right\|^{2}}{2} \right\} }   
    \exp {\left\{-\frac{\left\| \mathbf{y}_{t}-\left[ \gamma_{t} \mathbf{y}_{0} + \left(1-\gamma_{t} \right) f_{\phi}(\mathbf{x}) \right] \right\|^{2}}{2 \sigma_{t}^{2}} \right\}} d \mathbf{y}_{0}\\
    % & = \frac{c_{1}}{\sigma_{t}^{d} (2 \pi)^{\frac{d}{2}}} \int_{\mathbb{R}^{d}}
    % \exp{\left\{-\frac{1}{2 \sigma^{2}} \left[ c_2 \sigma^{2} \mathbf{y}_{0}^{2} + \mathbf{y}_{t}^{2} + \gamma_{t}^{2}\mathbf{y}_{0}^{2} + (1-\gamma_{t})^{2} f_{\phi}^{2}(\mathbf{x}) - 2\mathbf{y}_{t} (\gamma_{t} \mathbf{y}_{0} + \left(1-\gamma_{t} \right) f_{\phi}(\mathbf{x})) + 2 \gamma_{t} (1-\gamma_{t}) f_{\phi}(\mathbf{x}) \mathbf{y}_{0} \right] \right\}}d \mathbf{y}_{0}\\
    & =  \frac{c_1}{\left( \gamma_{t}^{2}+c_2 \sigma_{t}^{2} \right)^{\frac{d}{2}}} \exp{\left\{- \frac{c_2 \left\|\mathbf{y}_{t}-(1-\gamma_{t})f_{\phi}(\mathbf{x}))   \right\|^{2}}{2 (\gamma_{t}^{2}+c_2 \sigma_{t}^{2})}    \right\}}.
\end{aligned}
\end{equation*}
where the inequality is associated with the light tail hypothesis of the original conditional distribution $q(\mathbf{y}_{0}|f_{\phi}(\mathbf{x})) \leq c_{1} \exp\left\{ -c_{2} \left\| \mathbf{y}_{0} \right\|^{2}_{2} /2 \right\}$, and the final equation is obtained by means of the progressive expansion and simplification of the exponential terms.

Next, we similarly derive the lower bound of the conditional density function.
\begin{equation*}
\begin{aligned}
    q(\mathbf{y}_{t}|f_{\phi}(\mathbf{x})) & = \int_{\mathbb{R}^{d}} q(\mathbf{y}_{t},\mathbf{y}_{0}|f_{\phi}(\mathbf{x}))  d \mathbf{y}_{0} 
     = \int_{\mathbb{R}^{d}} q(\mathbf{y}_{t}|\mathbf{y}_{0},f_{\phi}(\mathbf{x})) q(\mathbf{y}_{0}|f_{\phi}(\mathbf{x})) d \mathbf{y}_{0}\\
     & = \frac{1}{\sigma_{t}^{d} (2 \pi)^{\frac{d}{2}}} \int_{\mathbb{R}^{d}}  
    q(\mathbf{y}_{0}|f_{\phi}(\mathbf{x}))      
    \exp {\left\{-\frac{\left\| \mathbf{y}_{t}-\left[ \gamma_{t} \mathbf{y}_{0} + \left(1-\gamma_{t} \right) f_{\phi}(\mathbf{x}) \right] \right\|^{2}}{2 \sigma_{t}^{2}} \right\}} d \mathbf{y}_{0}\\
    & \stackrel{(i)}{\geq} \frac{1}{\sigma_{t}^{d} (2 \pi)^{\frac{d}{2}}} \int_{\mathbb{R}^{d}}   
    q(\mathbf{y}_{0}|f_{\phi}(\mathbf{x}))      
    \exp {\left\{-\frac{2\left\| \mathbf{y}_{t}\right\|^{2} 
    +2 \left\|\gamma_{t} \mathbf{y}_{0} + \left(1-\gamma_{t} \right) f_{\phi}(\mathbf{x})  \right\|^{2}}{2 \sigma_{t}^{2}} \right\}} d \mathbf{y}_{0}\\
    & \stackrel{(ii)}{\geq} \frac{1}{\sigma_{t}^{d} (2 \pi)^{\frac{d}{2}}} \int_{\mathbb{R}^{d}}   
    q(\mathbf{y}_{0}|f_{\phi}(\mathbf{x}))      
    \exp {\left\{-\frac{2\left\| \mathbf{y}_{t}\right\|^{2} 
    + 4\gamma_{t}^{2}\left\|\mathbf{y}_{0} \right\|^{2} 
    + 4 \left(1-\gamma_{t} \right)^{2} \left\|f_{\phi}(\mathbf{x})  \right\|^{2}}{2 \sigma_{t}^{2}} \right\}} d \mathbf{y}_{0}\\
    & = \frac{1}{\sigma_{t}^{d} (2 \pi)^{\frac{d}{2}}} 
    \exp {\left\{-\frac{\left\| \mathbf{y}_{t}\right\|^{2} 
    + 2 \left(1-\gamma_{t} \right)^{2} \left\|f_{\phi}(\mathbf{x})  \right\|^{2}}{\sigma_{t}^{2}} \right\}}
    \int_{\mathbb{R}^{d}}   
    q(\mathbf{y}_{0}|f_{\phi}(\mathbf{x}))      
    \exp {\left\{-\frac{4\gamma_{t}^{2}\left\|\mathbf{y}_{0} \right\|^{2}}{2 \sigma_{t}^{2}} \right\}} d \mathbf{y}_{0}\\
    & \stackrel{(iii)}{\geq} \frac{1}{\sigma_{t}^{d} (2 \pi)^{\frac{d}{2}}} 
    \exp {\left\{-\frac{\left\| \mathbf{y}_{t}\right\|^{2} 
    + 2 \left(1-\gamma_{t} \right)^{2} \left\|f_{\phi}(\mathbf{x})  \right\|^{2}}{\sigma_{t}^{2}} \right\}}
    \int_{\left\| \mathbf{y}_{0} \right\|_{\infty} \leq R}   
    q(\mathbf{y}_{0}|f_{\phi}(\mathbf{x}))      
    \exp {\left\{-\frac{4\gamma_{t}^{2}\left\|\mathbf{y}_{0} \right\|^{2}}{2 \sigma_{t}^{2}} \right\}} d \mathbf{y}_{0}\\
    & \stackrel{(iv)}{\geq} \frac{1}{\sigma_{t}^{d} (2 \pi)^{\frac{d}{2}}} 
    \exp {\left\{-\frac{\left\| \mathbf{y}_{t}\right\|^{2} 
    + 2 \left(1-\gamma_{t} \right)^{2} \left\|f_{\phi}(\mathbf{x})  \right\|^{2}}{\sigma_{t}^{2}} \right\}}  
    \exp {\left\{-\frac{4\gamma_{t}^{2} R^{2}}{2\sigma_{t}^{2}} \right\}}
    \int_{\left\| \mathbf{y}_{0} \right\|_{\infty} \leq R}   
    q(\mathbf{y}_{0}|f_{\phi}(\mathbf{x}))  d \mathbf{y}_{0}\\
    & = \frac{c_3}{\sigma_{t}^{d}} 
    \exp {\left\{-\frac{\left\| \mathbf{y}_{t}\right\|^{2} 
    + 2 \left(1-\gamma_{t} \right)^{2} \left\|f_{\phi}(\mathbf{x})  \right\|^{2} + 2\gamma_{t}^{2} R^{2}}{\sigma_{t}^{2}} \right\}}.
\end{aligned}
\end{equation*}
where we define $c_3 = \int_{\left\| \mathbf{y}_{0} \right\|_{\infty} \leq R} q(\mathbf{y}_{0}|f_{\phi}(\mathbf{x}))  d \mathbf{y}_{0} / (2 \pi)^{\frac{d}{2}}$ in inequality (iv), and this term can be computed and regarded as a constant. Meanwhile, inequalities (i) and (ii) are obtained by repeatedly applying the Cauchy-Schwarz inequality, inequality (iii) is the use of truncation strategy that assuming $\mathbf{y}_{0} \in D_{1}$.

\begin{lemma}\label{lem-2}
Similarly, we further obtain a bound on the gradient term of the conditional density function $\nabla q(\mathbf{y}_{t}|f_{\phi}(\mathbf{x}))$ under the $L_{\infty}$ norm.
\begin{equation}\label{BB7}
\begin{aligned}
    \left\| \nabla q(\mathbf{y}_{t}|f_{\phi}(\mathbf{x})) \right\|_{\infty} 
    & \leq \frac{c_1}{\left( \gamma_{t}^{2}+c_2 \sigma_{t}^{2} \right)^{\frac{d}{2}}} \exp{\left\{- \frac{c_2 \left\|\mathbf{y}_{t}-(1-\gamma_{t})f_{\phi}(\mathbf{x}))   \right\|^{2}}{\gamma_{t}^{2}+c_2 \sigma_{t}^{2}}  \right\}} \\
    &\ \ \left( \frac{c_2 \left\| \mathbf{y}_{t} \right\|_{\infty}}{\gamma_{t}^{2}+c_2 \sigma_{t}^{2}} 
    + \frac{4 \gamma_{t}^{2} (1-\gamma_{t}) \left\| f_{\phi}(\mathbf{x}) \right\|_{\infty}}{\sigma_{t}^{2} \left( \gamma_{t}^{2}+c_2 \sigma_{t}^{2} \right)} 
    + \frac{4 \gamma_{t}}{\sigma_{t} \sqrt{\gamma_{t}^{2}+c_2 \sigma_{t}^{2}}}  \right).
\end{aligned}
\end{equation}
\end{lemma}

\noindent \textbf{Proof.} Without loss of generality, we can solve it for any dimension, and then the symmetry can be applied to the whole d dimensions. Suppose we solve for the first coordinate of $\nabla q(\mathbf{y}_{t}|f_{\phi}(\mathbf{x}))$ and denote as $\nabla q(\mathbf{y}_{t}|f_{\phi}(\mathbf{x}))_{1} $. 
\begin{equation*}
\begin{aligned}
    \nabla q(\mathbf{y}_{t}|f_{\phi}(\mathbf{x})) _{1} &= \left[ \frac{\partial q(\mathbf{y}_{t}|f_{\phi}(\mathbf{x}))}{\partial \mathbf{y}_{t}} \right]_{1}
    = \left[ \frac{\partial}{\partial \mathbf{y}_{t}}\int_{\mathbb{R}^{d}} q(\mathbf{y}_{t}|\mathbf{y}_{0},f_{\phi}(\mathbf{x})) q(\mathbf{y}_{0}|f_{\phi}(\mathbf{x})) d \mathbf{y}_{0}  \right]_{1}\\
    & = \frac{1}{\sigma_{t}^{d} (2 \pi)^{\frac{d}{2}}} \int_{\mathbb{R}^{d}}
    \frac{\mathbf{y}_{1}-[\gamma_{t}\mathbf{y}_{01}+(1-\gamma_{t})f_{\phi}(\mathbf{x})_{1}]}{\sigma_{t}^{2}} 
    q(\mathbf{y}_{0}|f_{\phi}(\mathbf{x}))     
    \exp {\left\{-\frac{\left\| \mathbf{y}_{t}-\left[ \gamma_{t} \mathbf{y}_{0} + \left(1-\gamma_{t} \right) f_{\phi}(\mathbf{x}) \right] \right\|^{2}}{2 \sigma_{t}^{2}} \right\}} d \mathbf{y}_{0},
\end{aligned}
\end{equation*}
where $\mathbf{y}_{01}$ and $f_{\phi}(\mathbf{x})_{1}$ table the first dimension of the vector $\mathbf{y}_{0}$ and $f_{\phi}(\mathbf{x})$ respectively. Simplification the absolute value of fractions in the first dimension,
\begin{equation*}
\begin{aligned}
\left| \frac{\mathbf{y}_{1}-[\gamma_{t}\mathbf{y}_{01}+(1-\gamma_{t})f_{\phi}(\mathbf{x})_{1}]}{\sigma
    _{t}^{2}} \right| \leq 
\frac{2 c_2}{\gamma_{t}^{2}
+ c_2 \sigma_{t}^{2}} |\mathbf{y}_{1}| + \frac{4 \gamma_{t}^{2} (1-\gamma_{t})}{\sigma_{t}^{2}(\gamma_{t}^{2}+c_2 \sigma_{t}^{2})}|f_{\phi}(\mathbf{x})_{1}|
+ \frac{4 \gamma_t}{\sigma_{t}^{2}} \left| \mathbf{y}_{01}-\frac{\gamma_t}{\gamma_{t}^{2}+c_2 \sigma_{t}^{2}} [\mathbf{y}_{1}-(1-\gamma_{t})f_{\phi}(\mathbf{x})_{1}] \right|.
\end{aligned}
\end{equation*}
Regarding the absolute value of the above expression, 
\begin{equation*}
\begin{aligned}
    \left| \nabla q(\mathbf{y}_{t}|f_{\phi}(\mathbf{x})) _{1} \right|
    & \leq \frac{c_1}{\sigma_{t}^{d} (2 \pi)^{\frac{d}{2}}} \int_{\mathbb{R}^{d}}
    \left| \frac{\mathbf{y}_{1}-[\gamma_{t}\mathbf{y}_{01}+(1-\gamma_{t})f_{\phi}(\mathbf{x})_{1}]}{\sigma
    _{t}^{2}} \right|
    \exp{\left\{-\frac{c_2 \left\|\mathbf{y}_{01}\right\|^{2}}{2}\right\}}    
    \exp {\left\{-\frac{\left\| \mathbf{y}_{t}-\left[ \gamma_{t} \mathbf{y}_{0} + \left(1-\gamma_{t} \right) f_{\phi}(\mathbf{x}) \right] \right\|^{2}}{2 \sigma_{t}^{2}} \right\}} d \mathbf{y}_{0}\\
    & = \frac{c_1}{\sigma_{t}^{d} (2 \pi)^{\frac{d}{2}}}
    \exp{\left\{ \frac{c_2 \left\| \mathbf{y}_{t}-\left(1-\gamma_{t} \right) f_{\phi}(\mathbf{x}) \right\|^{2}}{\gamma_{t}^{2}+c_2 \sigma_{t}^{2}}   \right\}}\\
    & \ \ \ \int_{\mathbb{R}^{d}}
    \left| \frac{\mathbf{y}_{1}-[\gamma_{t}\mathbf{y}_{01}+(1-\gamma_{t})f_{\phi}(\mathbf{x})_{1}]}{\sigma
    _{t}^{2}} \right|    
    \exp{\left\{-\frac{\left\| \mathbf{y}_{0}- \gamma_{t} \left[\mathbf{y}_{t} + \left(1-\gamma_{t} \right) f_{\phi}(\mathbf{x}) \right] / (\gamma_{t}^{2}+c_2 \sigma_{t}^{2} )\right\|^{2}} {2 \sigma_{t}^{2} / (\gamma_{t}^{2}+c_2 \sigma_{t}^{2})} \right\} } d \mathbf{y}_{0}\\
    & \leq \frac{c_1}{\left( \gamma_{t}^{2}+c_2 \sigma_{t}^{2} \right)^{\frac{d}{2}}} \exp{\left\{- \frac{c_2 \left\|\mathbf{y}_{t}-(1-\gamma_{t})f_{\phi}(\mathbf{x}))   \right\|^{2}}{\gamma_{t}^{2}+c_2 \sigma_{t}^{2}}  \right\}} \left( \frac{c_2 \left| \mathbf{y}_{1} \right|}{\gamma_{t}^{2}+c_2 \sigma_{t}^{2}} 
    + \frac{4 \gamma_{t}^{2} (1-\gamma_{t}) \left| f_{\phi}(\mathbf{x})_{1} \right|}{\sigma_{t}^{2} \left( \gamma_{t}^{2}+c_2 \sigma_{t}^{2} \right)} 
    + \frac{4 \gamma_{t}}{\sigma_{t} \sqrt{\gamma_{t}^{2}+c_2 \sigma_{t}^{2}}}  \right).
\end{aligned}
\end{equation*}
By taking all components into account based on symmetry, then the $L_{\infty}$ norm is employed to acquire an upper bound for the gradient term.

\subsubsection{Score function upper bounds}
\begin{lemma}\label{lem-3}
Under above mentioned assumptions, there exist constants $c_4,c_5>0$ such that the logarithmic condition density function $ \nabla \log q(\mathbf{y}_{t}|f_{\phi}(\mathbf{x}))$ under the $L_{\infty}$ norm can be bounded as:
\begin{equation}\label{BB8}
\left\| \nabla \log q(\mathbf{y}_{t}|f_{\phi}(\mathbf{x})) \right\|_{\infty} 
\leq \frac{c_4}{\sigma_{t}^{3}} \left( \left\|\mathbf{y}_{t} \right\|_{\infty}^{2} + 2(1-\gamma_{t})^{2} \left\| f_{\phi}(\mathbf{x})\right\|_{\infty}^{2}  \right)^{\frac{1}{2}} + c_5.
\end{equation}
\end{lemma}

\noindent \textbf{Proof.} We also solve the first coordinate of $\nabla \log q(\mathbf{y}_{t}|f_{\phi}(\mathbf{x}))$, then apply to the whole variable by symmetry. We denote first coordinate of the logarithmic gradient term is expressed as
$\nabla \log q(\mathbf{y}_{t}|f_{\phi}(\mathbf{x}))_{1} = \nabla q(\mathbf{y}_{t}|f_{\phi}(\mathbf{x}))_{1} / q(\mathbf{y}_{t}|f_{\phi}(\mathbf{x}))_{1}$.

Based on the truncation concept presented in Assumption \ref{assum8} and Assumption \ref{assum9}, we anticipate that the error related to $\nabla q(\mathbf{y}_{t}|f_{\phi}(\mathbf{x}))$ and $q(\mathbf{y}_{t}|f_{\phi}(\mathbf{x}))$ outside the truncation domain will be controllable. From the inspiration of the Lemma A.8 in XXX, we may as well obtain Assumption \ref{assum10}. First we assume that $\mathbf{y}_{0} \in D_{01}$ and $q(\mathbf{y}_{t}|f_{\phi}(\mathbf{x}))>2 \varepsilon_{low}$, i.e
\begin{equation*}
q(\mathbf{y}_{t}|f_{\phi}(\mathbf{x})) > \int_{D_{02}} q(\mathbf{y}_{0}|f_{\phi}(\mathbf{x})) q(\mathbf{y}_{t}|\mathbf{y}_{0},f_{\phi}(\mathbf{x})) d \mathbf{y}_{0} > \varepsilon_{low}.
\end{equation*}
\begin{equation*}
\begin{aligned}
\left| \nabla \log q(\mathbf{y}_{t}|f_{\phi}(\mathbf{x}))_{1} \right| &= \left| \frac{\nabla q(\mathbf{y}_{t}|f_{\phi}(\mathbf{x}))_{1}}{q(\mathbf{y}_{t}|f_{\phi}(\mathbf{x}))_{1}} \right| 
= \frac{\int_{R^{d}} \left| \frac{\mathbf{y}_{1}-[\gamma_{t}\mathbf{y}_{01}+(1-\gamma_{t})f_{\phi}(\mathbf{x})_{1}]}{\sigma_{t}^{2}} \right| q(\mathbf{y}_{0}|f_{\phi}(\mathbf{x})) q(\mathbf{y}_{t}|\mathbf{y}_{0},f_{\phi}(\mathbf{x})) d \mathbf{y}_{0}}{\int_{R^{d}} q(\mathbf{y}_{0}|f_{\phi}(\mathbf{x})) q(\mathbf{y}_{t}|\mathbf{y}_{0},f_{\phi}(\mathbf{x})) d \mathbf{y}_{0}} \\
&< \frac{\int_{D_{01}} \left| \frac{\mathbf{y}_{1}-[\gamma_{t}\mathbf{y}_{01}+(1-\gamma_{t})f_{\phi}(\mathbf{x})_{1}]}{\sigma_{t}^{2}} \right| q(\mathbf{y}_{0}|f_{\phi}(\mathbf{x})) q(\mathbf{y}_{t}|\mathbf{y}_{0},f_{\phi}(\mathbf{x})) d \mathbf{y}_{0}}{\int_{D_{01}} q(\mathbf{y}_{0}|f_{\phi}(\mathbf{x})) q(\mathbf{y}_{t}|\mathbf{y}_{0},f_{\phi}(\mathbf{x})) d \mathbf{y}_{0}} \\
& \ \ + \frac{\int_{R^{d} \backslash D_{01}} \left| \frac{\mathbf{y}_{1}-[\gamma_{t}\mathbf{y}_{01}+(1-\gamma_{t})f_{\phi}(\mathbf{x})_{1}]}{\sigma_{t}^{2}} \right| q(\mathbf{y}_{0}|f_{\phi}(\mathbf{x})) q(\mathbf{y}_{t}|\mathbf{y}_{0},f_{\phi}(\mathbf{x})) d \mathbf{y}_{0}}{\int_{D_{01}} q(\mathbf{y}_{0}|f_{\phi}(\mathbf{x})) q(\mathbf{y}_{t}|\mathbf{y}_{0},f_{\phi}(\mathbf{x})) d \mathbf{y}_{0}}\\
& \leq \frac{2\int_{D_{01}} \left| \frac{\mathbf{y}_{1}-[\gamma_{t}\mathbf{y}_{01}+(1-\gamma_{t})f_{\phi}(\mathbf{x})_{1}]}{\sigma_{t}^{2}} \right| q(\mathbf{y}_{0}|f_{\phi}(\mathbf{x})) q(\mathbf{y}_{t}|\mathbf{y}_{0},f_{\phi}(\mathbf{x})) d \mathbf{y}_{0}}{\int_{D_{01}} q(\mathbf{y}_{0}|f_{\phi}(\mathbf{x})) q(\mathbf{y}_{t}|\mathbf{y}_{0},f_{\phi}(\mathbf{x})) d \mathbf{y}_{0}}  + \frac{\varepsilon}{\varepsilon_{low}}\\
&\stackrel{(i)}{\leq} \frac{\frac{2c}{\sigma_{t}^{2}} \int_{D_{01}} \sqrt{\log \varepsilon^{-1}} q(\mathbf{y}_{0}|f_{\phi}(\mathbf{x})) q(\mathbf{y}_{t}|\mathbf{y}_{0},f_{\phi}(\mathbf{x})) d \mathbf{y}_{0}}{\int_{D_{01}} q(\mathbf{y}_{0}|f_{\phi}(\mathbf{x})) q(\mathbf{y}_{t}|\mathbf{y}_{0},f_{\phi}(\mathbf{x})) d \mathbf{y}_{0}} + 1\\
& = \frac{2}{\sigma_{t}^{2}} c\sqrt{\log \varepsilon^{-1}} +1\\
&\stackrel{(ii)}{\leq} \frac{2c}{\sigma_{t}^{2}} \sqrt{\log \left(\frac{\sigma_{t}^{d}}{c_3}\right)} + \frac{2\sqrt{2}c \gamma_{t} R}{\sigma_{t}^{3}} 
+ \frac{2c \sqrt{d+d_{x}}}{\sigma_{t}^{3}} \sqrt{\left\| \mathbf{y}_{t}\right\|^{2} + 2 \left(1-\gamma_{t} \right)^{2} \left\|f_{\phi}(\mathbf{x})  \right\|^{2}} + 1 \\
&\stackrel{(iii)}{=} \frac{c_4}{\sigma_{t}^{3}} \left(\left\| \mathbf{y}_{t}\right\|^{2}_{2} 
+ 2 \left(1-\gamma_{t} \right)^{2} \left\|f_{\phi}(\mathbf{x})  \right\|^{2}_{2}\right)^{\frac{1}{2}} + c_5.\\
\end{aligned}
\end{equation*}
For inequality (i), we assume $\varepsilon=\varepsilon_{low}$ and denote $\varepsilon = \min \left( \frac{c_3}{\sigma_{t}^{d}}  \exp {\left\{-\frac{\left\| \mathbf{y}_{t}\right\|^{2} 
+ 2 \left(1-\gamma_{t} \right)^{2} \left\|f_{\phi}(\mathbf{x})  \right\|^{2} + 2\gamma_{t}^{2} R^{2}}{\sigma_{t}^{2}} \right\}} , \frac{1}{e} \right)$ based on the lower bound of $q(\mathbf{y}_{t}|f_{\phi}(\mathbf{x}))$ in Lemma \ref{lem-1}.
Inequality (ii) holds if and only if $\varepsilon$ takes the first value above, and use the vector Cauchy-Schwarz inequality. 
In simple terms, we define a vector $\mathbf{n}$ of dimension $(d + d_x)$ as $\mathbf{n}=(\mathbf{y}_{t1}/\sigma,\cdots,\mathbf{y}_{td}/\sigma, \sqrt{2} (1-\gamma_t)f_{\phi}(\mathbf{x})_{1}/\sigma,\cdots,\sqrt{2} (1-\gamma_t)f_{\phi}(\mathbf{x})_{d_x}/\sigma)^T$, and a vector $\mathbf{m}$ of the same dimension with all elements being $1$ and we have $(\mathbf{m}\cdot\mathbf{n})^2 \leq \|\mathbf{m}\|_{2}^2\|\mathbf{n}\|_{2}^2=(d + d_x)\|\mathbf{n}\|_{2}^2$ by the Cauchy-Schwarz inequality.
For inequality (iii), we denote 
\begin{equation*}
c_4=2c \sqrt{d+d_{x}},\   c_5=\frac{2c}{\sigma_{t}^{2}} \sqrt{\log \left(\frac{\sigma_{t}^{d}}{c_3}\right)} + \frac{2\sqrt{2}c \gamma_{t} R}{\sigma_{t}^{3}}.
\end{equation*}
where $c_4$ depends on the dimensions of the two random variables $\mathbf{y}_{t},f_{\phi}(\mathbf{x})$.

Similarly, consider all dimensions under the $L_{\infty}$ norm and
get the final result.

\subsubsection{Bound analysis of integral terms}
\begin{lemma}\label{lem-4}
Suppose Assumption \ref{assum8} holds. For any $R>1,\ t>0$, we have
\begin{equation}\label{BB9}
\begin{aligned}
    \int_{\left\|\mathbf{y}_{t} \right\|_{\infty} \geq R} q(\mathbf{y}_{t}|f_{\phi}(\mathbf{x})) d \mathbf{y}_{t} 
    & \leq \frac{c_1 R}{(\gamma_{t}^{2}+c_2 \sigma_{t}^{2})^{\frac{d}{2}}} \exp{\left\{-\frac{c_2 \left\| R-(1-\gamma_{t})f_{\phi}(\mathbf{x}) \right\|_{2}^{2}}{2(\gamma_{t}^{2}+c_2 \sigma_{t}^{2})}  \right\}}.\\
    % &\lesssim R \exp{\left\{-\frac{c_2 \left\| R-(1-\gamma_{t})f_{\phi}(\mathbf{x}) \right\|_{2}^{2}}{\gamma_{t}^{2}+c_2 \sigma_{t}^{2}}  \right\}}.
\end{aligned}
\end{equation}

\begin{equation}\label{BB10}
\begin{aligned}
    &\ \ \int_{\left\|\mathbf{y}_{t} \right\|_{\infty} \geq R} 
    \left\| \nabla \log q(\mathbf{y}_{t}|f_{\phi}(\mathbf{x})) \right\|_{2}^{2}
    q(\mathbf{y}_{t}|f_{\phi}(\mathbf{x})) d \mathbf{y}_{t} \\
    & \leq \frac{d c_1 c_4^{2}}{\sigma_{t}^{6} (\gamma_{t}^{2}+c_2 \sigma_{t}^{2})^{\frac{d}{2}}}  \left[ R^{3} + 6(1-\gamma_{t})^{2} \left\|f_{\phi}(\mathbf{x})) \right\|_{2} ^{2}R  \right]
    \exp{\left\{-\frac{c_2 \left\| R-(1-\gamma_{t})f_{\phi}(\mathbf{x}) \right\|_{2}^{2}}{2(\gamma_{t}^{2}+c_2 \sigma_{t}^{2})}  \right\}}.
\end{aligned}
\end{equation}
\end{lemma}

\noindent \textbf{Proof.} Lemma \ref{lem-4} shows that when we only focus on the range of values of the variable $\mathbf{y}_{t}$, specifically, under the condition specified in Assumption \ref{assum8} when $\mathbf{y}_{t}$ is outside the given region, the truncation error of the integrals related to the conditional density and the logarithmic gradient can be effectively controlled.

By leveraging the conclusions of Lemma \ref{lem-1} and performing a straightforward algebraic manipulation, we obtain the following result.
\begin{equation*}
\begin{aligned}
\int_{\left\|\mathbf{y}_{t} \right\|_{\infty} \geq R} q(\mathbf{y}_{t}|f_{\phi}(\mathbf{x})) d \mathbf{y}_{t} 
& \leq \int_{\left\|\mathbf{y}_{t} \right\|_{\infty} \geq R} \frac{c_1}{(\gamma_{t}^{2}+c_2 \sigma_{t}^{2})^{\frac{d}{2}}} \exp{\left\{-\frac{c_2 \left\|\mathbf{y}_{t}-(1-\gamma_{t})f_{\phi}(\mathbf{x}) \right\|_{2}^{2}}{2(\gamma_{t}^{2}+c_2 \sigma_{t}^{2})}  \right\}} d \mathbf{y}_{t}\\
& \leq \frac{c_1 R}{(\gamma_{t}^{2}+c_2 \sigma_{t}^{2})^{\frac{d}{2}}} \exp{\left\{-\frac{c_2 \left\| R-(1-\gamma_{t})f_{\phi}(\mathbf{x}) \right\|_{2}^{2}}{2(\gamma_{t}^{2}+c_2 \sigma_{t}^{2})}  \right\}}.
\end{aligned}
\end{equation*}
Analogously, we apply Lemma \ref{lem-3} and have
\begin{equation*}
\begin{aligned}
    &\ \ \int_{\left\|\mathbf{y}_{t} \right\|_{\infty} \geq R} 
    \left\| \nabla \log q(\mathbf{y}_{t}|f_{\phi}(\mathbf{x})) \right\|_{2}^{2}
    q(\mathbf{y}_{t}|f_{\phi}(\mathbf{x})) d \mathbf{y}_{t} \\
    & \leq \int_{\left\|\mathbf{y}_{t} \right\|_{\infty} \geq R} 
     d \left\| \nabla \log q(\mathbf{y}_{t}|f_{\phi}(\mathbf{x})) \right\|_{\infty}^{2} q(\mathbf{y}_{t}|f_{\phi}(\mathbf{x}))\\
    & \leq \frac{d c_1 c_4^{2}}{\sigma_{t}^{6} (\gamma_{t}^{2}+c_2 \sigma_{t}^{2})^{\frac{d}{2}}}  \int_{\left\|\mathbf{y}_{t} \right\|_{\infty} \geq R}  \left(  \left\|\mathbf{y}_{t}\right\|^{2} + 2(1-\gamma_{t})^{2} \left\|f_{\phi}(\mathbf{x})\right\|^{2} \right) \exp{\left\{-\frac{c_2 \left\| \mathbf{y}_{t}-(1-\gamma_{t})f_{\phi}(\mathbf{x}) \right\|_{2}^{2}}{2(\gamma_{t}^{2}+c_2 \sigma_{t}^{2})}  \right\}}\\
    & \leq \frac{d c_1 c_4^{2}}{\sigma_{t}^{6} (\gamma_{t}^{2}+c_2 \sigma_{t}^{2})^{\frac{d}{2}}}  \left[ R^{3} + 6(1-\gamma_{t})^{2} \left\|f_{\phi}(\mathbf{x})) \right\|_{2} ^{2}R  \right]
    \exp{\left\{-\frac{c_2 \left\| R-(1-\gamma_{t})f_{\phi}(\mathbf{x}) \right\|_{2}^{2}}{2(\gamma_{t}^{2}+c_2 \sigma_{t}^{2})}  \right\}}.
\end{aligned}
\end{equation*}
The proof is complete.

\begin{lemma}\label{lem-5}
Suppose Assumption \ref{assum8} and Assumption \ref{assum9} hold. For any $R>1,\ t>0$ and $\varepsilon_{low}>0$, we have  
\begin{equation}\label{BB11}
    \int_{\left\|\mathbf{y}_{t} \right\|_{\infty} \leq R} q(\mathbf{y}_{t}|f_{\phi}(\mathbf{x})) \mathbf{1}_{\left\{ \left|q(\mathbf{y}_{t}|f_{\phi}(\mathbf{x}))\right| < \varepsilon_{low} \right\} } d \mathbf{y}_{t} \leq  2^{d} R^{d} \varepsilon_{low}.
\end{equation}

\begin{equation}\label{BB12}
\begin{aligned}
    \int_{\left\|\mathbf{y}_{t} \right\|_{\infty} \leq R} 
    \left\| \nabla \log q(\mathbf{y}_{t}|f_{\phi}(\mathbf{x})) \right\|_{2}^{2}
    q(\mathbf{y}_{t}|f_{\phi}(\mathbf{x})) \mathbf{1}_{\left\{ \left|q(\mathbf{y}_{t}|f_{\phi}(\mathbf{x}))\right| < \varepsilon_{low} \right\} }   d \mathbf{y}_{t}
    \leq \frac{2^{d} c_4^{2} d R^{d} \varepsilon_{low}}{\sigma_{t}^{6}} \left[ d R^{2} + 2(1-\gamma_{t})^{2} \left\|f_{\phi}(\mathbf{x})\right\|_{2}^{2}\right].
\end{aligned}
\end{equation}
\end{lemma}

\noindent \textbf{Proof.} Lemma \ref{lem-5} shows that based on the assumption of the truncation domains, when the value of $\mathbf{y}_{t}$ is taken in the given region, while its conditional density is small, the truncation error of the corresponding integral term can be limited.

We can do some simple algebra as follows.
\begin{equation*}
\begin{aligned}
\int_{\left\|\mathbf{y}_{t} \right\|_{\infty} \leq R} q(\mathbf{y}_{t}|f_{\phi}(\mathbf{x})) \mathbf{1}_{\left\{ \left|q(\mathbf{y}_{t}|f_{\phi}(\mathbf{x}))\right| < \varepsilon_{low} \right\} } d \mathbf{y}_{t} 
\leq \varepsilon_{low} \int_{\left\|\mathbf{y}_{t} \right\|_{\infty} \leq R} d \mathbf{y}_{t} 
= (2R)^{d} \varepsilon_{low}.
% \lesssim R^{d} \varepsilon_{low}.
\end{aligned}
\end{equation*}
When the $d$ dimensional vector $\mathbf{y}_{t}$ lies within the range $-R < \mathbf{y}_{ti} < R$ for all $i = 1,2,\cdots,d$, the integral essentially computes the volume of a $d$ dimensional cube whose side length measures $2R$.
We involve the upper bound in Lemma \ref{lem-3} and have
\begin{equation*}
\begin{aligned}
&\ \ \int_{\left\|\mathbf{y}_{t} \right\|_{\infty} \leq R} 
\left\| \nabla \log q(\mathbf{y}_{t}|f_{\phi}(\mathbf{x})) \right\|_{2}^{2}
q(\mathbf{y}_{t}|f_{\phi}(\mathbf{x})) \mathbf{1}_{\left\{ \left|q(\mathbf{y}_{t}|f_{\phi}(\mathbf{x}))\right| < \varepsilon_{low} \right\} }   d \mathbf{y}_{t}\\
% & \leq \varepsilon_{low} \int_{\left\|\mathbf{y}_{t} \right\|_{\infty} \leq R} \left\| \nabla \log q(\mathbf{y}_{t}|f_{\phi}(\mathbf{x})) \right\|_{2}^{2}   d \mathbf{y}_{t}\\
& \leq d \varepsilon_{low} \int_{\left\|\mathbf{y}_{t} \right\|_{\infty} \leq R} \left\| \nabla \log q(\mathbf{y}_{t}|f_{\phi}(\mathbf{x})) \right\|_{\infty}^{2} d \mathbf{y}_{t}\\
& \leq d \varepsilon_{low}  \left[ \frac{c_4}{\sigma_t^{3}} \left(  \left\|\mathbf{y}_{t}\right\|^{2} + 2(1-\gamma_{t})^{2} \left\|f_{\phi}(\mathbf{x})\right\|^{2} \right)^{\frac{1}{2}} + c_5\right]^{2} \int_{\left\|\mathbf{y}_{t} \right\|_{\infty} \leq R} d \mathbf{y}_{t}\\
& \leq d \varepsilon_{low} (2R)^{d} \left[ \frac{c_4}{\sigma_t^{3}} \left(  d R^{2} + 2(1-\gamma_{t})^{2} \left\|f_{\phi}(\mathbf{x})\right\|^{2} \right)^{\frac{1}{2}} \right]^{2} \\
&\leq \frac{2^{d} c_4^{2} d R^{d} \varepsilon_{low}}{\sigma_{t}^{6}} \left[ d R^{2} + 2(1-\gamma_{t})^{2} \left\|f_{\phi}(\mathbf{x})\right\|_{2}^{2}\right].
% \\&\lesssim \frac{ R^{d} \varepsilon_{low}}{\sigma_{t}^{6}} \left[ d R^{2} + 2(1-\gamma_{t})^{2} \left\|f_{\phi}(\mathbf{x})\right\|_{2}^{2}\right].
\end{aligned}
\end{equation*}
The proof is complete.

\subsection{Conditional diffused local Taylor polynomials}
\begin{lemma}\label{lem-6}
Suppose above Assumptions hold. Given any integer $N>0, t \in [0,T]$ and $R_{*}=\max (2R,2R_{f})$ for constants $R>1,R_{f}>1$, we constrain $(\mathbf{y}_{t},f_{\phi}(\mathbf{x})) \in [0,1]^{d} \times [0,1]^{d_x}$, then there exists a network $s_{\theta}(\mathbf{y}_{t},f_{\phi}(\mathbf{x}),t)$ satisfying
\begin{equation}\label{BB27}
\begin{aligned}
&\ \ \ \left\| s_{\theta}(\mathbf{y}_{t},f_{\phi}(\mathbf{x}),t)-\nabla \log q(\mathbf{y}_{t}|f_{\phi}(\mathbf{x})) \right\|_{\infty} q(\mathbf{y}_{t}|f_{\phi}(\mathbf{x})) \\
& \leq  \left[   \frac{1}{\sigma_{t}}\left( \varepsilon 
+ \frac{c BR^{s}_{*} (d+d_x)^{s} \left(\log \varepsilon^{-1}\right)^{\frac{1}{2}}}{N^{\beta} s! \gamma_{t}^{\frac{d}{2}}} 
+ \frac{c B R_{*}^{s+d} \varepsilon^{\frac{3}{e}} \left(\log \varepsilon^{-1}\right)^{\frac{1}{2}} }{N^d \sigma_{t}^{d} (2 \pi)^{\frac{d}{2}}} \left(1+\frac{ (d+d_x)^{s}}{N^{\beta} s!}  \right)  \right)\right.\\
&\ \ \ +\left.\left(\frac{c_4}{\sigma_{t}^{3}} \left( R^{2} + 2(1-\gamma_{t})^{2} R_{f}^{2}  \right)^{\frac{1}{2}} + c_5 \right)
\left( \varepsilon + \frac{BR^{s}_{*} (d+d_x)^{s}}{N^{\beta} s! \gamma_{t}^{\frac{d}{2}}} + \frac{d B R_{*}^{s+d} \varepsilon^{\frac{3}{e}}}{N^d \sigma_{t}^{d} (2 \pi)^{\frac{d}{2}}} \left(1+\frac{ (d+d_x)^{s}}{N^{\beta} s!} \right) \right) \right].
% \\& \lesssim \left[  \frac{1}{\sigma_{t}} \left(\varepsilon 
% + \frac{B \left(\log \varepsilon^{-1}\right)^{\frac{1}{2}}}{N^{\beta} \gamma_{t}^{\frac{d}{2}}} 
% + \frac{ B \varepsilon^{\frac{3}{e}} \left(\log \varepsilon^{-1}\right)^{\frac{1}{2}} (N^{\beta}+1)}{N^{d+\beta} \sigma_{t}^{d}} \right) 
% + \left(\frac{1}{\sigma_{t}^{3}} \left( R^{2} + 2(1-\gamma_{t})^{2} R_{f}^{2}  \right)^{\frac{1}{2}} + c_5 \right)   
% \left( \varepsilon + \frac{B}{N^{\beta}  \gamma_{t}^{\frac{d}{2}}} + \frac{B \varepsilon^{\frac{3}{e}} (N^{\beta}+1)}{N^{d+\beta} \sigma_{t}^{d}}\right)
% \right]
\end{aligned}
\end{equation} 
\end{lemma}

The core of the proof is the integral form of $q(\mathbf{y}_{t}|f_{\phi}(\mathbf{x}))$ and $\nabla q(\mathbf{y}_{t}|f_{\phi}(\mathbf{x}))$ approximated by a set of Taylor expansion polynomials as the basis function. Furthermore, our focus lies more on the scenario of the Taylor expansion at the highest-order terms. Within the context of CARD, for the sake of establishing a connection with the diffusion model, it could aptly be named the conditional diffused local Taylor polynomials. To motivate it, we repeat the integral form of $q(\mathbf{y}_{t}|f_{\phi}(\mathbf{x}))$ as follows,
\begin{equation*}
\begin{aligned}
q(\mathbf{y}_{t}|f_{\phi}(\mathbf{x})) 
& = \frac{1}{\sigma_{t}^{d} (2 \pi)^{\frac{d}{2}}} \int_{\mathbb{R}^{d}}  q(\mathbf{y}_{0}|f_{\phi}(\mathbf{x}))      
\exp {\left\{-\frac{\left\| \mathbf{y}_{t}-\left[ \gamma_{t} \mathbf{y}_{0} + \left(1-\gamma_{t} \right) f_{\phi}(\mathbf{x}) \right] \right\|^{2}}{2 \sigma_{t}^{2}} \right\}} d \mathbf{y}_{0}.
\end{aligned}
\end{equation*}
Then we convert into the Taylor expansions of its components, namely the original conditional density function $q(\mathbf{y}_{0}|f_{\phi}(\mathbf{x}))$ and the Gaussian function $\exp {\left\{-\left\| \mathbf{y}_{t}-\left[ \gamma_{t} \mathbf{y}_{0} + \left(1-\gamma_{t} \right) f_{\phi}(\mathbf{x}) \right] \right\|^{2} / 2 \sigma^{2} \right\}}$ respectively.

In Assumption \ref{assum6}, we have made an assumption regarding the smoothness of the original conditional density, $q(\mathbf{y}_{0}|f_{\phi}(\mathbf{x})) \in \mathscr{H}^{\beta}(\mathbb{R}^{d} \times \mathbb{R}^{d_{x}},B)$ for $H\ddot{o}lder$ index $\beta >0$ and constant $B>0$. We hope that the conditional density at any time during the diffusion process, $q(\mathbf{y}_{t}|f_{\phi}(\mathbf{x}))$ also has smoothness, so that a better approximation can be achieved using Taylor polynomials.
Based on the above discussion, we can approximately divide the process of solving the error bound of the conditional density function $q(\mathbf{y}_{t}|f_{\phi}(\mathbf{x}))$ into three steps:

\textbf{(i) Truncation domain.} The first step involves truncating the domain of integral term with $\mathbf{y}_{0} \in D_{0}=D_{01}\cap D_{02}$. Meanwhile, the input conditional variable $f_{\phi}(\mathbf{x})$ is truncated in $[-R_{f},R_{f}]^{d_{x}}$. Finally we define a specific range for the variables to limit the area of consideration.

\textbf{(ii) Taylor expansion of exponential function.} In the second step, the Gaussian kernel function is expanded using Taylor series, and the upper bound of its convergence is then obtained. This step helps in analyzing the behavior and approximation properties of the Gaussian kernel within the considered domain.

\textbf{(iii) Taylor expansion of original conditional density function.} In the third step, the original conditional density is approximated by means of Taylor series as well. 

We similar approximate $\nabla q(\mathbf{y}_{t}|f_{\phi}(\mathbf{x}))$ with the Taylor polynomials, the details of which are as follows. Finally, we obtain the score function $\nabla \log q(\mathbf{y}_{t}|f_{\phi}(\mathbf{x}))$ approximated by the conditional diffusion local Taylor polynomial. By following these steps, we can effectively determine the error bound.

\subsubsection{Truncation domain}
First, we truncate the original data $\mathbf{y}_{0}$ such as Eq.(\ref{D01}) and Eq.(\ref{D02}), and Eq.(\ref{BB4}) in Assumption \ref{assum10} shows that the integral with respect to $q(\mathbf{y}_{t}|f_{\phi}(\mathbf{x}))$ outside the truncation domain can be controlled by the error $\varepsilon \in (0,1/e)$, i.e., for all $\mathbf{y}_{t} \in R^{d}\ (t \in [0,T]), f_{\phi}(\mathbf{x}) \in R^{d_{x}}$, we have
\begin{equation*}
\begin{aligned}
\left| \int_{\overline{D}_{0}} q(\mathbf{y}_{0}|f_{\phi}(\mathbf{x})) \phi(\mathbf{y}_{t}|\mathbf{y}_{0},f_{\phi}(\mathbf{x})) d \mathbf{y}_{0} \right|  \leq \varepsilon,
\end{aligned}
\end{equation*}
If we denote $h_{2}(\mathbf{y}_{t},\mathbf{y}_{0},f_{\phi}(\mathbf{x}))=\int_{D_{0}} q(\mathbf{y}_{0}|f_{\phi}(\mathbf{x})) \phi(\mathbf{y}_{t}|\mathbf{y}_{0},f_{\phi}(\mathbf{x})) d \mathbf{y}_{0}$, then
\begin{equation}\label{BB13}
\begin{aligned}
% & \left| \int_{R^{d}} q(\mathbf{y}_{0}|f_{\phi}(\mathbf{x})) \phi(\mathbf{y}_{t}|\mathbf{y}_{0},f_{\phi}(\mathbf{x})) d \mathbf{y}_{0}
% -\int_{D_{0}} q(\mathbf{y}_{0}|f_{\phi}(\mathbf{x})) \phi(\mathbf{y}_{t}|\mathbf{y}_{0},f_{\phi}(\mathbf{x})) d \mathbf{y}_{0} \right|\\
& \left| q(\mathbf{y}_{t}|f_{\phi}(\mathbf{x}))
- h_{2}(\mathbf{y}_{t},\mathbf{y}_{0},f_{\phi}(\mathbf{x})) \right| \leq \varepsilon.
\end{aligned}
\end{equation}

Second, in order to approximate the conditional density distribution with greater precision, we take into account that the predicted conditional mean $f_{\phi}(\mathbf{x})$ of the input is truncated within the interval $[-R_{f}, R_{f}]$ by the original distribution range $f_{\phi}(\mathbf{x}) \in R^x$. Meanwhile, it is ensured that the error of the score function outside this truncation domain can be bounded by $1 / \sigma_{t}^{2} \exp{\left\{-c_{f_{2}} R_{f}^{2}/2  \right\}}$, as demonstrated in Lemma \ref{lem-6}. 
\begin{lemma}\label{lem-7}
Suppose above Assumptions hold. For all truncation radius $R,R_{f}>0$, constants $m_0>0$ and $f_{\phi}(\mathbf{x}) \in R^{d_x}$, we have
\begin{equation}\label{BB14}
\begin{aligned}
&\ \ \textup{E}_{f_{\phi}(\mathbf{x}) \sim q(f_{\phi}(\mathbf{x}))} \left\{ \textup{E}_{\mathbf{y}_{t} \sim q(\mathbf{y}_{t}|f_{\phi}(\mathbf{x}))} \left[ \mathbf{1}\left\{ \|f_{\phi}(\mathbf{x})\|_{\infty} \geq R_{f} \right\} \| \nabla \log q(\mathbf{y}_{t}|f_{\phi}(\mathbf{x}))  \|^{2}_{2}    \right] \right\}\\
& \leq d_x \exp{\left\{-\frac{c_{f_{2}} R_{f}^{2}}{2} \right\}}
\left[ c_{f_{1}}\left( \frac{2d c_4 c_6}{\sigma_{t}^{4}}+\frac{\sqrt{\pi}d c_4 c_5 c_6}{\sigma_{t}^{2}} \right) + c_f c_3^{\prime}
+ \frac{c_1 c_f R}{(\gamma_{t}^{2}+c_2 \sigma_{t}^{2})^{\frac{2}{d}}} \exp{\left\{-\frac{c_2 \| R-(1-\gamma_{t})R_{f} \|^{2}_{2}}{2(\gamma_{t}^{2}+c_2 \sigma_{t}^{2})} \right\}} \right].
\end{aligned}
\end{equation}
\end{lemma}

\noindent \textbf{Proof.} Lemma \ref{lem-7} shows that under the truncation assumption of $\|\mathbf{y}_{0}\|_{\infty} \leq R$, when we truncate the condition within $\|f_{\phi}(\mathbf{x})\|_{\infty} \leq R_f$, the error outside the truncation domain can be controlled by the upper bound as in Eq.(\ref{BB14}).

The previous Lemma \ref{lem-4} centers around the variation of the score function beyond the truncation domain of $\mathbf{y}_{t}$. In this context, our primary concern is whether the variation of $\mathbf{y}_{0}$ across the entire domain can be effectively regulated. 

For all truncation radius $R,R_{f}>0$ and $f_{\phi}(\mathbf{x}) \in R^{d_x}$, we first have
\begin{equation*}
\begin{aligned}
&\ \ \ \textup{E}_{\mathbf{y}_{t} \sim q(\mathbf{y}_{t}|f_{\phi}(\mathbf{x}))} \left[  \| \nabla \log q(\mathbf{y}_{t}|f_{\phi}(\mathbf{x}))  \|^{2}_{2}    \right]\\
&=\int_{R^{d}} \| \nabla \log q(\mathbf{y}_{t}|f_{\phi}(\mathbf{x})) \|^{2}_{2}  q(\mathbf{y}_{t}|f_{\phi}(\mathbf{x}))  d \mathbf{y}_{t}\\
% & \leq d \int_{R^{d}}  \| \nabla \log q(\mathbf{y}_{t}|f_{\phi}(\mathbf{x})) \|^{2}_{\infty}  q(\mathbf{y}_{t}|f_{\phi}(\mathbf{x}))  d \mathbf{y}_{t}\\
&\stackrel{(i)}{\leq} d \int_{R^{d}}  \left[\frac{c_4}{\sigma_{t}^{3}} \left( \left\|\mathbf{y}_{t} \right\|_{\infty}^{2} + 2(1-\gamma_{t})^{2} \left\| f_{\phi}(\mathbf{x})\right\|_{\infty}^{2}  \right)^{\frac{1}{2}} + c_5 \right]^{2}  q(\mathbf{y}_{t}|f_{\phi}(\mathbf{x}))  d \mathbf{y}_{t}\\
& = \frac{d c_4}{\sigma_{t}^{3}} \int_{R^{d}} \left\|\mathbf{y}_{t} \right\|_{\infty}^{2} q(\mathbf{y}_{t}|f_{\phi}(\mathbf{x}))  d \mathbf{y}_{t}
+ \left(\frac{2d c_4 (1-\gamma_{t})^{2}}{\sigma_{t}^{6}}\left\| f_{\phi}(\mathbf{x})\right\|_{\infty}^{2} +d c_5^{2}  \right) \int_{R^{d}} q(\mathbf{y}_{t}|f_{\phi}(\mathbf{x}))  d \mathbf{y}_{t}\\
&\ \ \ + \frac{2d c_4 c_5}{\sigma_{t}^{3}} \int_{R^{d}} \left( \left\|\mathbf{y}_{t} \right\|_{\infty}^{2} + 2(1-\gamma_{t})^{2} \left\| f_{\phi}(\mathbf{x})\right\|_{\infty}^{2}  \right)^{\frac{1}{2}} q(\mathbf{y}_{t}|f_{\phi}(\mathbf{x}))  d \mathbf{y}_{t}\\
&\stackrel{(ii)}{\leq} \frac{d c_4}{\sigma_{t}^{3}} \mathbb{E}_{\mathbf{y}_{t}}[\left\| \mathbf{y}_{t}\right\|_{\infty}^{2}]
+ \frac{2d c_4 c_5}{\sigma_{t}^{3}} \mathbb{E}_{\mathbf{y}_{t}}[\left\| \mathbf{y}_{t}\right\|_{\infty}]\\
&\ \ \ + \left(\frac{2d c_4 (1-\gamma_{t})^{2}}{\sigma_{t}^{6}}\left\| f_{\phi}(\mathbf{x})\right\|_{\infty}^{2}  + \frac{2 \sqrt{2} d c_4 c_5 (1-\gamma_t)}{\sigma_{t}^{3}}\left\| f_{\phi}(\mathbf{x})\right\|_{\infty} +d c_5^{2} \right) 
\int_{R^{d}} q(\mathbf{y}_{t}|f_{\phi}(\mathbf{x}))  d \mathbf{y}_{t}\\ 
% \left( \int_{\left\| \mathbf{y}_{t}\right\|_{\infty} \leq R} q(\mathbf{y}_{t}|f_{\phi}(\mathbf{x}))  d \mathbf{y}_{t} + \int_{\left\| \mathbf{y}_{t}\right\|_{\infty} \geq R} q(\mathbf{y}_{t}|f_{\phi}(\mathbf{x}))  d \mathbf{y}_{t} \right)\\
&\stackrel{(iii)}{\leq} 
\left[ \frac{2d c_4^2 (1-\gamma_{t})^2}{\sigma_{t}^{6}}  \left\| f_{\phi}(\mathbf{x})\right\|_{\infty}^{2}
+ \frac{2 \sqrt{2} d c_4 c_5 (1-\gamma_t)}{\sigma_{t}^{3}}  \left\| f_{\phi}(\mathbf{x})\right\|_{\infty}
+ d c_5^2 \right]
\left[c_3^{\prime} + \frac{c_1 R}{(\gamma_{t}^{2}+c_2 \sigma_{t}^{2})^{\frac{2}{d}}} \exp{\left\{-\frac{c_2 \| R-(1-\gamma_{t}) f_{\phi}(\mathbf{x})\|^{2}_{2}}{2(\gamma_{t}^{2}+c_2 \sigma_{t}^{2})} \right\}}  \right]\\
&\ \ \ +\frac{2d c_4 c_6}{\sigma_{t}^{4}}  + \frac{\sqrt{\pi} d c_4 c_5 c_6}{\sigma_{t}^{2}}.
\end{aligned}
\end{equation*}
For inequality (i), we use the upper bound of score function in Lemma \ref{lem-3}. 
For inequality (ii), from the sub Gaussian tail probability bound, we assume there exist constants $m,c_6 >0$ such that $P(\left\| \mathbf{y}_{t}\right\|_{\infty} \geq m) \leq c_6 \exp{\{-m^2/ 2 \sigma_t^2\} }$ holds.  
For inequality (iii), $\mathbf{y}_{t} \in R^d $ is divided into two parts $ R^d=\{ \left\| \mathbf{y}_{t}\right\|_{\infty} \leq R \} \cup \{ \left\| \mathbf{y}_{t}\right\|_{\infty} \geq R\} $ and we denote 
$c_3^{\prime} = \int_{\left\| \mathbf{y}_{t}\right\|_{\infty} \leq R} q(\mathbf{y}_{t}|f_{\phi}(\mathbf{x}))  d \mathbf{y}_{t}$ which treated it as a computable bounded constant. After simplification, the upper bound of each term is calculated separately. 

Thus, we obtain that
\begin{equation*}
\begin{aligned}
&\ \ \ \textup{E}_{f_{\phi}(\mathbf{x}) \sim q(f_{\phi}(\mathbf{x}))} \left\{ \textup{E}_{\mathbf{y}_{t} \sim q(\mathbf{y}_{t}|f_{\phi}(\mathbf{x}))} \left[ \mathbf{1}\left\{ \|f_{\phi}(\mathbf{x})\|_{\infty} \geq R_{f} \right\} \| \nabla \log q(\mathbf{y}_{t}|f_{\phi}(\mathbf{x}))  \|^{2}_{2}    \right] \right\}\\
& = \textup{E}_{f_{\phi}(\mathbf{x}) \sim q(f_{\phi}(\mathbf{x}))}  \left\{ \mathbf{1}\left\{ \|f_{\phi}(\mathbf{x})\|_{\infty} \geq R_{f} \right\} 
 \textup{E}_{\mathbf{y}_{t} \sim q(\mathbf{y}_{t}|f_{\phi}(\mathbf{x}))} \left[  \| \nabla \log q(\mathbf{y}_{t}|f_{\phi}(\mathbf{x}))  \|^{2}_{2}    \right] \right\}\\
% &\stackrel{(i)}{\leq} \mathbb{E}_{f_{\phi}(\mathbf{x})}  \left\{ 
% \mathbf{1}\left\{ \|f_{\phi}(\mathbf{x})\|_{\infty} \geq R_{f} \right\} \left(\frac{2d c_4 c_6}{\sigma_{t}^{4}}  + \frac{\sqrt{\pi} d c_4 c_5 c_6}{\sigma_{t}^{2}}\right) \right\}\\
% & \ \ \ + \mathbb{E}_{f_{\phi}(\mathbf{x}) }  \left\{ 
% \mathbf{1}\left\{ \|f_{\phi}(\mathbf{x})\|_{\infty} \geq R_{f} \right\}
% \left[ \frac{2d c_4^2 (1-\gamma_{t})^2}{\sigma_{t}^{6}}  \left\| f_{\phi}(\mathbf{x})\right\|_{\infty}^{2}
% + \frac{2 \sqrt{2} d c_4 c_5 (1-\gamma_t)}{\sigma_{t}^{3}}  \left\| f_{\phi}(\mathbf{x})\right\|_{\infty}
% + d c_5^2 \right] \right.\\
% &\ \ \ \left.\left[c_3^{\prime} + \frac{c_1 R}{(\gamma_{t}^{2}+c_2 \sigma_{t}^{2})^{\frac{2}{d}}} \exp{\left\{-\frac{c_2 \| R-(1-\gamma_{t}) f_{\phi}(\mathbf{x})\|^{2}_{2}}{2(\gamma_{t}^{2}+c_2 \sigma_{t}^{2})} \right\}}  \right] \right\}\\
&\stackrel{(i)}{\leq}\textup{E}_{f_{\phi}(\mathbf{x})}  \left\{ 
\mathbf{1}\left\{ \|f_{\phi}(\mathbf{x})\|_{\infty} \geq R_{f} \right\} \left(\frac{2d c_4 c_6}{\sigma_{t}^{4}}  + \frac{\sqrt{\pi} d c_4 c_5 c_6}{\sigma_{t}^{2}}\right) \right\}\\
& \ \ \ + \textup{E}_{f_{\phi}(\mathbf{x}) }  \left\{ 
\left[ \frac{2d c_4^2 (1-\gamma_{t})^2}{\sigma_{t}^{6}}  m_0^{2}
+ \frac{2 \sqrt{2} d c_4 c_5 (1-\gamma_t)}{\sigma_{t}^{3}}  m_0
+ d c_5^2 \right] 
\left[c_3^{\prime} + \frac{c_1 R}{(\gamma_{t}^{2}+c_2 \sigma_{t}^{2})^{\frac{2}{d}}} \exp{\left\{-\frac{c_2 \| R-(1-\gamma_{t}) f_{\phi}(\mathbf{x})\|^{2}_{2}}{2(\gamma_{t}^{2}+c_2 \sigma_{t}^{2})} \right\}}  \right] \right\}\\
&\stackrel{(ii)}{\leq} d_x \exp{\left\{-\frac{c_{f_{2}} R_{f}^{2}}{2} \right\}}
\left[ c_{f_{1}}\left( \frac{2d c_4 c_6}{\sigma_{t}^{4}}+\frac{\sqrt{\pi}d c_4 c_5 c_6}{\sigma_{t}^{2}} \right) + c_f c_3^{\prime}
+ \frac{c_1 c_f R}{(\gamma_{t}^{2}+c_2 \sigma_{t}^{2})^{\frac{2}{d}}} \exp{\left\{-\frac{c_2 \| R-(1-\gamma_{t})R_{f} \|^{2}_{2}}{2(\gamma_{t}^{2}+c_2 \sigma_{t}^{2})} \right\}} \right].
\end{aligned}
\end{equation*}
To control the upper bound of this function, we define a truncation function $m(f)=\min(\|f_{\phi}(\mathbf{x})\|_{\infty},m_0)$, where constant $m_0>0$. Evidently, when $R_f < m_0$, the indicator function $1\{m(f)\leq R_f\}=1$, conversely, it equals $0$. Hence, the inequality (i) holds. 
For inequality (i) and (ii), we both use Assumption \ref{assum7}. Specially, we denote $c_f = c_{f_{1}} \left( \frac{2d c_4^2 (1-\gamma_{t})^2}{\sigma_{t}^{6}}  m_0^{2}
+ \frac{2 \sqrt{2} d c_4 c_5 (1-\gamma_t)}{\sigma_{t}^{3}}  m_0
+ d c_5^2 \right)$ in inequality (ii).

\subsubsection{Taylor expansion of exponential function}
We have truncated the variable $\mathbf{y}_{t}(t\in [0,T])$ and the condition $f_{\phi}(\mathbf{x})$ respectively, concentrating the data $(\mathbf{y}_{t},f_{\phi}(\mathbf{x})) \in [-R,R]^{d} \times [-R_{f},R_{f}]^{d_x}$.
% \begin{equation*}
% \begin{aligned}
%  &\mathbf{y}_{0} \in D_{0}=D_{01}\cap D_{02}\\
% &=\left[-c\sqrt{\log \varepsilon^{-1}}, c\sqrt{\log \varepsilon^{-1}}  \right]
% \cap \left[ \frac{\mathbf{y}_{t}- \left(1-\gamma_{t} \right) f_{\phi}(\mathbf{x})-\sigma_{t} c\sqrt{\log \varepsilon^{-1}}}{\gamma_{t}},\ \frac{\mathbf{y}_{t}- \left(1-\gamma_{t} \right) f_{\phi}(\mathbf{x})+\sigma_{t} c\sqrt{\log \varepsilon^{-1}}}{\gamma_{t}}  \right].
% \end{aligned}
% \end{equation*}
Suppose we determine the first coordinate by solving the exponential function and leveraging Taylor expansion polynomials, which written as $\forall \mathbf{y}_{0} \in D_{0}$,
\begin{equation*}
\begin{aligned}
\exp {\left\{-\frac{\left| \mathbf{y}_{1}-\left[ \gamma_{t} \mathbf{y}_{01} + \left(1-\gamma_{t} \right) f_{\phi}(\mathbf{x})_{1} \right] \right|^{2}}{2 \sigma_{t}^{2}} \right\}} 
= \sum\limits_{k=0}^{p-1} \frac{1}{k!} \left( -\frac{\left| \mathbf{y}_{1}-\left[ \gamma_{t} \mathbf{y}_{01} + \left(1-\gamma_{t} \right) f_{\phi}(\mathbf{x})_{1} \right] \right|^{2}}{2 \sigma_{t}^{2}} \right)^{k}
+ R_{p}.
\end{aligned}
\end{equation*}
where the remainder of above Taylor polynomial, denoted as $R_{p}=\sum_{k=p}^{\infty} \frac{1}{k!} \left( -\frac{\left| \mathbf{y}_{1}-\left[ \gamma_{t} \mathbf{y}_{01} + \left(1-\gamma_{t} \right) f_{\phi}(\mathbf{x})_{1} \right] \right|^{2}}{2 \sigma_{t}^{2}} \right)^{k}$. 
Our objective is to ensure $R_{p}$ is bounded by a suitably small constant $\varepsilon_{p}$ within the open interval $(0,1)$. Mathematically, this is expressed as $R_{p} \leq \varepsilon_{p}$.
By virtue of Lagrange theorem, there exists a value $\xi$ belonging to the interval $[0,\left| \mathbf{y}_{1}-\left[ \gamma_{t} \mathbf{y}_{01} + \left(1-\gamma_{t} \right) f_{\phi}(\mathbf{x})_{1} \right] \right|^{2}/2 \sigma_{t}^{2}]$, such that $e^{-\xi} \in (0,1]$.
Upon simplification, we obtain the inequality $\varepsilon_{p} \geq (\log \varepsilon)^{p}/p!$. Correspondingly, the remainder can be expressed as $R_{p} \leq (\log \varepsilon)^{p}/p!$, where previously set $\varepsilon \in (0,\frac{1}{e})$.
From Stirling's approximation, the inequality $p! \geq (\frac{p}{3})^{p}$ can be deduced. Subsequently, when an appropriate order is selected such that $p = \frac{3 \log \varepsilon}{e}$, the inequality $R_{p} \leq \varepsilon^{\frac{3}{e}}$ is satisfied.

With respect to the entire $d$ dimensional vector, we have
\begin{equation}\label{BB15}
\begin{aligned}
&\left|\exp {\left\{-\frac{\left\| \mathbf{y}_{t}-\left[ \gamma_{t} \mathbf{y}_{0} + \left(1-\gamma_{t} \right) f_{\phi}(\mathbf{x})\right] \right\|^{2}}{2 \sigma_{t}^{2}} \right\}} 
% &\left| \prod\limits_{i=1}^{d} \exp{\left\{-\frac{\left| \mathbf{y}_{i}-\left[ \gamma_{t} \mathbf{y}_{0i} + \left(1-\gamma_{t} \right) f_{\phi}(\mathbf{x})_{i} \right] \right|^{2}}{2 \sigma^{2}} \right\}}
- \prod\limits_{i=1}^{d} \sum\limits_{k=0}^{p-1} \frac{1}{k!} \left( -\frac{\left| \mathbf{y}_{i}-\left[ \gamma_{t} \mathbf{y}_{0i} + \left(1-\gamma_{t} \right) f_{\phi}(\mathbf{x}) \right] \right|^{2}}{2 \sigma_{t}^{2}} \right)^{k} \right| \leq d \varepsilon^{\frac{3}{e}}.
\end{aligned}
\end{equation}
This gives us the upper bound of the Taylor approximation for the exponential term. The proof is complete.

\subsubsection{Taylor expansion of original conditional density function}\label{appendixB43}
Then we employ the Taylor polynomial approximation method for $q(\mathbf{y}_{0}|f_{\phi}(\mathbf{x}))$ within bounded truncation area in Section 4.4.1. 
To begin with, it is more precise within a relatively narrow range, thereby resulting in a minor approximation error for each dimension. Without loss of generality, assume that the Taylor approximation is carried out within the range of $[0, 1]^{d} \times [0, 1]^{d_x}$, then we denote
\begin{equation}\label{BB16}
    \tau(\mathbf{y},\mathbf{f})=q(R_{*}(\mathbf{y}_{0}-\frac{1}{2})|R_{*}(f_{\phi}(\mathbf{x})-\frac{1}{2})),\ \mathbf{y} \in [0, 1]^{d},\mathbf{f} \in [0, 1]^{d_x}.
\end{equation}
where $R_{*}=\max (2R,2R_{f})$.

In particular, similar to \cite{fu2024unveil}, we might as well introduce an additional parameter $N$, which represents the number of divisions in each dimension. That is, the length of each sub interval is $1/N$. Meanwhile, let vectors $\mathbf{v},\mathbf{w}$ denote the interval indices on variables $\mathbf{y},\mathbf{f}$ respectively, and we have $\mathbf{v}\in [N]^{d}, \mathbf{w}\in [N]^{d_x}$. 

\noindent\textbf{Step 1:} Perform a Taylor expansion of $\tau(\mathbf{y},\mathbf{f})$ at the point $\left( \frac{\mathbf{v}}{N}, \frac{\mathbf{w}}{N}\right)$ and there exist vector $\mathbf{\eta}^{1} \in [0,1]^{d}, \mathbf{\eta}^{2} \in  [0,1]^{d_x}$ of the highest $s$ order term. The form is as follows
\begin{equation}\label{BB17}
\begin{aligned}
\tau(\mathbf{y},\mathbf{f}) &=
\sum\limits_{\|\mathbf{n}\|_{1}+\|\mathbf{m}\|_{1}<s} \left.\frac{1}{\mathbf{n}! \mathbf{m}!} \frac{\partial^{\mathbf{n}+\mathbf{m}} \tau(\mathbf{y},\mathbf{f})}{\partial \mathbf{y}^{\mathbf{n}}  \partial \mathbf{f}^{\mathbf{m}}} \right|_{\mathbf{y}=\frac{\mathbf{v}}{N},\mathbf{f}=\frac{\mathbf{w}}{N}}  
\left( \mathbf{y}-\frac{\mathbf{v}}{N} \right)^{\mathbf{n}}  \left( \mathbf{f}-\frac{\mathbf{w}}{N} \right)^{\mathbf{m}}\\
&+ \sum\limits_{\|\mathbf{n}\|_{1}+\|\mathbf{m}\|_{1}=s} \left.\frac{1}{\mathbf{n}! \mathbf{m}!} \frac{\partial^{\mathbf{n}+\mathbf{m}} \tau(\mathbf{y},\mathbf{f})}{\partial \mathbf{y}^{\mathbf{n}}  \partial \mathbf{f}^{\mathbf{m}}} \right|_{\mathbf{y}=(1-\mathbf{\eta}^{1})\frac{\mathbf{v}}{N}+\mathbf{\eta}^{1}\mathbf{y},\mathbf{f}=(1-\mathbf{\eta}^{2})\frac{\mathbf{w}}{N}+\mathbf{\eta}^{2}\mathbf{f}  }
\left( \mathbf{y}-\frac{\mathbf{v}}{N} \right)^{\mathbf{n}}  \left( \mathbf{f}-\frac{\mathbf{w}}{N} \right)^{\mathbf{m}}.
\end{aligned}
\end{equation}
Among them, there is a linear change within this small interval where $\mathbf{y} \in \left(\frac{\mathbf{v}}{N},\frac{\mathbf{v}+1}{N}\right)^{d}, \mathbf{f} \in \left(\frac{\mathbf{w}}{N},\frac{\mathbf{w}+1}{N}\right)^{d_x}$.

\noindent\textbf{Step 2:} There is still a good Taylor approximation in the interval $\mathbf{y} \in \left(\frac{\mathbf{v}-1}{N},\frac{\mathbf{v}}{N}\right)^{d}, \mathbf{f} \in \left(\frac{\mathbf{w}-1}{N},\frac{\mathbf{w}}{N}\right)^{d_x}$. We denote the conditional diffusion local polynomial as
\begin{equation}\label{BB18}
    q^{1}(\mathbf{y},\mathbf{f}) = \sum\limits_{\mathbf{v} \in [N]^{d},\mathbf{w} \in [N]^{d_x}} 
    \Psi_{\mathbf{v},\mathbf{w}}(\mathbf{y},\mathbf{f}) \ \tau^{1}_{\mathbf{v},\mathbf{w}}(\mathbf{y},\mathbf{f}).
\end{equation}
where 
\begin{equation}\label{BB19}
\tau^{1}_{\mathbf{v},\mathbf{w}}(\mathbf{y},\mathbf{f}) = 
\sum\limits_{\|\mathbf{n}\|_{1}+\|\mathbf{m}\|_{1} \leq s} \left.\frac{1}{\mathbf{n}! \mathbf{m}!} \frac{\partial^{\mathbf{n}+\mathbf{m}} \tau(\mathbf{y},\mathbf{f})}{\partial \mathbf{y}^{\mathbf{n}}  \partial \mathbf{f}^{\mathbf{m}}} \right|_{\mathbf{y}=\frac{\mathbf{v}}{N},\mathbf{f}=\frac{\mathbf{w}}{N}}  
\left( \mathbf{y}-\frac{\mathbf{v}}{N} \right)^{\mathbf{n}}  \left( \mathbf{f}-\frac{\mathbf{w}}{N} \right)^{\mathbf{m}},
\end{equation}
is the $s$ order Taylor polynomial of $\tau(\mathbf{y},\mathbf{f})$ at the point $\left( \frac{\mathbf{v}}{N}, \frac{\mathbf{w}}{N}\right)$.
	
\begin{equation}\label{BB20}
\Psi_{\mathbf{v},\mathbf{w}}(\mathbf{y},\mathbf{f}) = \mathbf{1}\left\{ \mathbf{y} \in \left(\frac{\mathbf{v}-1}{N},\frac{\mathbf{v}}{N}\right)^{d} \right\}
\prod_{j=1}^{d_x} \psi \left[ 3N \left( \mathbf{f}_{j}-\frac{\mathbf{w}_{j}}{N}\right)\right],
\end{equation}
can be seen as an indicator function supported on the neighbor of the point and we use trapezoid function $\psi$ as 
\begin{equation*}
\psi(a)=\left\{\begin{array}{cc}
1 & |a|<1, \\
2-|a| & |a| \in[1,2], \\
0 & |a|>2.
\end{array}\right.
\end{equation*}

% We analyze the Taylor polynomial variation of the highest order $s$ when $\mathbf{y} \in \left(\frac{\mathbf{v}-1}{N},\frac{\mathbf{v}+1}{N}\right)^{d}$. 
Ultimately, we discover that the conditionally diffused local polynomial Eq.(\ref{BB18}) within the left domain approaches the Taylor polynomial Eq.(\ref{BB16}) in the right neighborhood. Thus, across the entire small neighborhood $\mathbf{y} \in \left(\frac{\mathbf{v}-1}{N},\frac{\mathbf{v}+1}{N}\right)^{d}$, we obtain
\begin{equation*}
\begin{aligned}
&\ \ \ \left|\tau^{1}_{\mathbf{v},\mathbf{w}}(\mathbf{y},\mathbf{f}) -\tau(\mathbf{y},\mathbf{f}) \right|\\
% & = \left| \sum\limits_{\|\mathbf{n}\|_{1}+\|\mathbf{m}\|_{1} \leq s} \left.\frac{1}{\mathbf{n}! \mathbf{m}!} \frac{\partial^{\mathbf{n}+\mathbf{m}} \tau(\mathbf{y},\mathbf{f})}{\partial \mathbf{y}^{\mathbf{n}}  \partial \mathbf{f}^{\mathbf{m}}} \right|_{\mathbf{y}=\frac{\mathbf{v}}{N},\mathbf{f}=\frac{\mathbf{w}}{N}}  
% \left( \mathbf{y}-\frac{\mathbf{v}}{N} \right)^{\mathbf{n}}  \left( \mathbf{f}-\frac{\mathbf{w}}{N} \right)^{\mathbf{m}} \right.\\
% &-\sum\limits_{\|\mathbf{n}\|_{1}+\|\mathbf{m}\|_{1}<s} \left.\frac{1}{\mathbf{n}! \mathbf{m}!} \frac{\partial^{\mathbf{n}+\mathbf{m}} \tau(\mathbf{y},\mathbf{f})}{\partial \mathbf{y}^{\mathbf{n}}  \partial \mathbf{f}^{\mathbf{m}}} \right|_{\mathbf{y}=\frac{\mathbf{v}}{N},\mathbf{f}=\frac{\mathbf{w}}{N}}  
% \left( \mathbf{y}-\frac{\mathbf{v}}{N} \right)^{\mathbf{n}}  \left( \mathbf{f}-\frac{\mathbf{w}}{N} \right)^{\mathbf{m}}\\
% &- \left.\sum\limits_{\|\mathbf{n}\|_{1}+\|\mathbf{m}\|_{1}=s} \left.\frac{1}{\mathbf{n}! \mathbf{m}!} \frac{\partial^{\mathbf{n}+\mathbf{m}} \tau(\mathbf{y},\mathbf{f})}{\partial \mathbf{y}^{\mathbf{n}}  \partial \mathbf{f}^{\mathbf{m}}} \right|_{\mathbf{y}=(1-\mathbf{\eta}^{1})\frac{\mathbf{v}}{N}+\mathbf{\eta}^{1}\mathbf{y},\mathbf{f}=(1-\mathbf{\eta}^{2})\frac{\mathbf{w}}{N}+\mathbf{\eta}^{2}\mathbf{f}  }
% \left( \mathbf{y}-\frac{\mathbf{v}}{N} \right)^{\mathbf{n}}  \left( \mathbf{f}-\frac{\mathbf{w}}{N} \right)^{\mathbf{m}} \right|\\
& =\left| \sum\limits_{\|\mathbf{n}\|_{1}+\|\mathbf{m}\|_{1} = s} \left.\frac{1}{\mathbf{n}! \mathbf{m}!} \frac{\partial^{\mathbf{n}+\mathbf{m}} \tau(\mathbf{y},\mathbf{f})}{\partial \mathbf{y}^{\mathbf{n}}  \partial \mathbf{f}^{\mathbf{m}}} \right|_{\mathbf{y}=\frac{\mathbf{v}}{N},\mathbf{f}=\frac{\mathbf{w}}{N}}  
\left( \mathbf{y}-\frac{\mathbf{v}}{N} \right)^{\mathbf{n}}  \left( \mathbf{f}-\frac{\mathbf{w}}{N} \right)^{\mathbf{m}} \right.\\
&-\left.\sum\limits_{\|\mathbf{n}\|_{1}+\|\mathbf{m}\|_{1}=s} \left.\frac{1}{\mathbf{n}! \mathbf{m}!} \frac{\partial^{\mathbf{n}+\mathbf{m}} \tau(\mathbf{y},\mathbf{f})}{\partial \mathbf{y}^{\mathbf{n}}  \partial \mathbf{f}^{\mathbf{m}}} \right|_{\mathbf{y}=(1-\mathbf{\eta}^{1})\frac{\mathbf{v}}{N}+\mathbf{\eta}^{1}\mathbf{y},\mathbf{f}=(1-\mathbf{\eta}^{2})\frac{\mathbf{w}}{N}+\mathbf{\eta}^{2}\mathbf{f}  }
\left( \mathbf{y}-\frac{\mathbf{v}}{N} \right)^{\mathbf{n}}  \left( \mathbf{f}-\frac{\mathbf{w}}{N} \right)^{\mathbf{m}} \right|\\
&\stackrel{(i)}{\leq} \sum\limits_{\|\mathbf{n}\|_{1}+\|\mathbf{m}\|_{1} = s}
\frac{1}{\mathbf{n}! \mathbf{m}!}
\left( \mathbf{y}-\frac{\mathbf{v}}{N} \right)^{\mathbf{n}}  \left( \mathbf{f}-\frac{\mathbf{w}}{N} \right)^{\mathbf{m}}  
\left\|[\mathbf{\eta}^{1}\mathbf{y},\mathbf{\eta}^{2}\mathbf{f}]-\frac{[\mathbf{\eta}^{1}\mathbf{v},\mathbf{\eta}^{2}\mathbf{w}]}{N}  \right\|^{\gamma}_{\infty} BR^{s}_{*}\\
&\stackrel{(ii)}{\leq} \sum\limits_{\|\mathbf{n}\|_{1}+\|\mathbf{m}\|_{1} = s}
\frac{BR^{s}_{*}}{\mathbf{n}! \mathbf{m}!} \left( \frac{1}{N}\right)^{\mathbf{n}} \left( \frac{1}{N}\right)^{\mathbf{m}} \left( \frac{1}{N}\right)^{r}\\
% & = \frac{BR^{s}_{*}}{N^{\beta}} \sum\limits_{\|\mathbf{n}\|_{1}+\|\mathbf{m}\|_{1} = s}
% \frac{1}{\mathbf{n}! \mathbf{m}!}\\
& \stackrel{(iii)}{\leq} \frac{BR^{s}_{*} (d+d_x)^{s}}{N^{\beta} s!}.
\end{aligned}
\end{equation*}
For inequality (i), we assume function $q(\mathbf{y}_{0}|f_{\phi}(\mathbf{x}))$ is $H\ddot{o}lder$ continuous by Assumption \ref{assum6},  which means $\|\tau\|_{\mathscr{H}^{\beta}([0, 1]^{d} \times [0, 1]^{d_x},B)} < B R_{*}^{s}$. For inequality (ii), the $L_{\infty}$ distance between $[\mathbf{y},\mathbf{f}]$ and $[\frac{\mathbf{v}}{N},\frac{\mathbf{w}}{N}]$ is at most $\frac{1}{N}$. For inequality (iii), we use binomial theorem. 

By integrating the preceding outcome with the equality $\sum_{\mathbf{v}\in [N]^{d}, \mathbf{w}\in [N]^{d_x}} \Psi_{\mathbf{v},\mathbf{w}}(\mathbf{y},\mathbf{f})=1$ for any $\mathbf{y} \in [0, 1]^{d},\mathbf{f} \in [0, 1]^{d_x}$, we assert that $q^{1}(\mathbf{y},\mathbf{f})$ serves as an approximation for $\tau(\mathbf{y},\mathbf{f})$ which satisfies
\begin{equation}\label{BB21}
\left|q^{1}(\mathbf{y},\mathbf{f}) -\tau(\mathbf{y},\mathbf{f}) \right| \leq \frac{BR^{s}_{*} (d+d_x)^{s}}{N^{\beta} s!}.
\end{equation}

\noindent\textbf{Step 3:} Approximate $q(\mathbf{y}_{0}|f_{\phi}(\mathbf{x}))$ by $q^{1}\left(\frac{\mathbf{y}_{0}}{R_{*}}+\frac{1}{2},\frac{\mathbf{f}}{R_{*}}+\frac{1}{2} \right)$. Similar to $h_{2}(\mathbf{y}_{t},\mathbf{y}_{0},f_{\phi}(\mathbf{x}))$, we denote
\begin{equation*}
h_{3}(\mathbf{y}_{t},\mathbf{y}_{0},f_{\phi}(\mathbf{x}))
= \int_{D_{0}} q^{1}\left(\frac{\mathbf{y}_{0}}{R_{*}}+\frac{1}{2},\frac{\mathbf{f}}{R_{*}}+\frac{1}{2}  \right)
\phi(\mathbf{y}_{t}|\mathbf{y}_{0},f_{\phi}(\mathbf{x}) d \mathbf{y}_{0},
\end{equation*}
then expanding it according to the definition, we have
\begin{equation*}
\begin{aligned}
&\ \ \ \ h_{3}(\mathbf{y}_{t},\mathbf{y}_{0},f_{\phi}(\mathbf{x}))\\
% & = \frac{1}{\sigma_{t}^{d} (2 \pi)^{\frac{d}{2}}} \int_{D_{0}}  \sum\limits_{\mathbf{v},\mathbf{w}} 
% \Psi_{\mathbf{v},\mathbf{w}}\left(\frac{\mathbf{y}_{0}}{R_{*}}+\frac{1}{2},\frac{\mathbf{f}}{R_{*}}+\frac{1}{2} \right) 
% \ \tau^{1}_{\mathbf{v},\mathbf{w}}\left(\frac{\mathbf{y}_{0}}{R_{*}}+\frac{1}{2},\frac{\mathbf{f}}{R_{*}}+\frac{1}{2} \right) 
% \exp {\left\{-\frac{\left\| \mathbf{y}_{t}-\left[ \gamma_{t} \mathbf{y}_{0} + \left(1-\gamma_{t} \right) f_{\phi}(\mathbf{x}) \right] \right\|^{2}}{2 \sigma^{2}} \right\}} d \mathbf{y}_{0}\\
& = \sum\limits_{\mathbf{v},\mathbf{w}} \sum\limits_{\|\mathbf{n}\|_{1}+\|\mathbf{m}\|_{1} = s} 
\left.\frac{1}{\mathbf{n}! \mathbf{m}!} \frac{\partial^{\mathbf{n}+\mathbf{m}} \tau\left(\frac{\mathbf{y}_{0}}{R_{*}}+\frac{1}{2},\frac{\mathbf{f}}{R_{*}}+\frac{1}{2} \right)}{\partial \mathbf{y}_{0}^{\mathbf{n}}  \partial \mathbf{f}^{\mathbf{m}}} \right|_{\mathbf{y}_{0}=\frac{\mathbf{v}}{N},\mathbf{f}=\frac{\mathbf{w}}{N}}  
\left(\frac{\mathbf{f}}{R_{*}}+\frac{1}{2}-\frac{\mathbf{w}}{N} \right)
\prod_{j=1}^{d_x} \psi \left[ 3N \left( \frac{\mathbf{f}_{j}}{R_{*}}+\frac{1}{2}-\frac{\mathbf{w}_{j}}{N}\right)\right]\\
&\ \ \  \cdot \prod\limits_{i=1}^d \frac{1}{\sigma_{t} \sqrt{2 \pi}} \int_{D_{0i}} 
\left( \frac{\mathbf{y}_{0}}{R_{*}}+\frac{1}{2}-\frac{\mathbf{v}_{i}}{N} \right)^{n_{i}}
\exp {\left\{-\frac{\left\| \mathbf{y}_{ti}-\left[ \gamma_{t} \mathbf{y}_{0i} + \left(1-\gamma_{t} \right) f_{\phi}(\mathbf{x}) \right] \right\|^{2}}{2 \sigma^{2}} \right\}} d \mathbf{y}_{0i}.
\end{aligned}
\end{equation*}
where 
\begin{equation*}
\begin{aligned}
D_{0i}
=\left[ \left(\frac{\mathbf{v}_{i}-1}{N}R_{*}-\frac{1}{2}\right),\left(\frac{\mathbf{v}_{i}}{N}R_{*}-\frac{1}{2}\right) \right]
\cap 
\left[ \frac{\mathbf{y}_{ti}- \left(1-\gamma_{t} \right) f_{\phi}(\mathbf{x})-\sigma_{t} c\sqrt{\log \varepsilon^{-1}}}{\gamma_{t}},\ \frac{\mathbf{y}_{ti}- \left(1-\gamma_{t} \right) f_{\phi}(\mathbf{x})+\sigma_{t} c\sqrt{\log \varepsilon^{-1}}}{\gamma_{t}}  \right].
\end{aligned}
\end{equation*}
Thus, regarding the error between true value $q(\mathbf{y}_{0}|f_{\phi}(\mathbf{x}))$ and its Taylor approximation $q^{1}\left(\frac{\mathbf{y}_{0}}{R_{*}}+\frac{1}{2},\frac{\mathbf{f}}{R_{*}}+\frac{1}{2} \right)$ in the truncated domain as
\begin{equation}\label{BB22}
\begin{aligned}
\left|h_{3}(\mathbf{y}_{t},\mathbf{y}_{0},f_{\phi}(\mathbf{x}))-h_{2}(\mathbf{y}_{t},\mathbf{y}_{0},f_{\phi}(\mathbf{x}))\right|
% \\&= \int_{D_{0}} \left| q^{1}\left(\frac{\mathbf{y}_{0}}{R_{*}}+\frac{1}{2},\frac{\mathbf{f}}{R_{*}}+\frac{1}{2}\right) -q(\mathbf{y}_{0}|f_{\phi}(\mathbf{x})) \right|
% \phi(\mathbf{y}_{t}|\mathbf{y}_{0},f_{\phi}(\mathbf{x}) d \mathbf{y}_{0}\\
% &\leq \frac{BR^{s}_{*} (d+d_x)^{s}}{N^{\beta} s! \sigma_{t}^{d} (2 \pi)^{\frac{d}{2}}}
% \int_{R_{d}} \exp {\left\{-\frac{\left\| \mathbf{y}_{t}-\left[ \gamma_{t} \mathbf{y}_{0} + \left(1-\gamma_{t} \right) f_{\phi}(\mathbf{x}) \right] \right\|^{2}}{2 \sigma_{t}^{2}} \right\}} d \mathbf{y}_{0}\\&
\leq \frac{BR^{s}_{*} (d+d_x)^{s}}{N^{\beta} s! \gamma_{t}^{\frac{d}{2}}}.
\end{aligned}
\end{equation}

\subsubsection{Taylor approximation of conditional density function $q(\mathbf{y}_{t}|f_{\phi}(\mathbf{x}))$}
We define $h_{1}(\mathbf{y}_{t},\mathbf{y}_{0},f_{\phi}(\mathbf{x}))$ as the conditional distribution $q(\mathbf{y}_{t}|f_{\phi}(\mathbf{x}))$ within the framework of applying the conditional diffusion local polynomial approximation, and its specific form is
\begin{equation*}
\begin{aligned}
&\ \ \ \ h_{1}(\mathbf{y}_{t},\mathbf{y}_{0},f_{\phi}(\mathbf{x}))\\
&=\sum\limits_{\mathbf{v},\mathbf{w}} \sum\limits_{\|\mathbf{n}\|_{1}+\|\mathbf{m}\|_{1} \leq s} 
\left.\frac{1}{\mathbf{n}! \mathbf{m}!} \frac{\partial^{\mathbf{n}+\mathbf{m}} \tau\left(\frac{\mathbf{y}_{0}}{R_{*}}+\frac{1}{2},\frac{\mathbf{f}}{R_{*}}+\frac{1}{2} \right)}{\partial \mathbf{y}_{0}^{\mathbf{n}}  \partial \mathbf{f}^{\mathbf{m}}} \right|_{\mathbf{y}_{0}=\frac{\mathbf{v}}{N},\mathbf{f}=\frac{\mathbf{w}}{N}}  
\left(\frac{\mathbf{f}}{R_{*}}+\frac{1}{2}-\frac{\mathbf{w}}{N} \right)
\prod_{j=1}^{d_x} \psi \left[ 3N \left( \frac{\mathbf{f}_{j}}{R_{*}}+\frac{1}{2}-\frac{\mathbf{w}_{j}}{N}\right)\right]\\
&\ \ \  \cdot \prod\limits_{i=1}^d \frac{1}{\sigma_{t} \sqrt{2 \pi}} \int_{D_{0i}} 
\left( \frac{\mathbf{y}_{0}}{R_{*}}+\frac{1}{2}-\frac{\mathbf{v}_{i}}{N} \right)^{n_{i}}
\sum\limits_{k=0}^{p-1} \frac{1}{k!} \left( -\frac{\left| \mathbf{y}_{i}-\left[ \gamma_{t} \mathbf{y}_{0i} + \left(1-\gamma_{t} \right) f_{\phi}(\mathbf{x})_{i} \right] \right|^{2}}{2 \sigma_{t}^{2}} \right)^{k} 
d \mathbf{y}_{0i},
\end{aligned}
\end{equation*}
So we have 
\begin{equation}\label{BB23}
\left|h_{3}(\mathbf{y}_{t},\mathbf{y}_{0},f_{\phi}(\mathbf{x}))-h_{1}(\mathbf{y}_{t},\mathbf{y}_{0},f_{\phi}(\mathbf{x}))\right|
\leq \frac{d B R_{*}^{s+d} \varepsilon^{\frac{3}{e}}}{N^d \sigma_{t}^{d} (2 \pi)^{\frac{d}{2}}} \left[1+\frac{ (d+d_x)^{s}}{N^{\beta} s!}  \right].
\end{equation}
and combine Eq.(\ref{BB13}), Eq.(\ref{BB22}) and Eq.(\ref{BB23}), then
\begin{equation}\label{BB24}
\begin{aligned}
\left|q(\mathbf{y}_{t}|f_{\phi}(\mathbf{x}))-h_{1}(\mathbf{y}_{t},\mathbf{y}_{0},f_{\phi}(\mathbf{x}))\right|
&\leq 2\left[ \varepsilon + \frac{BR^{s}_{*} (d+d_x)^{s}}{N^{\beta} s! \gamma_{t}^{\frac{d}{2}}} + \frac{d B R_{*}^{s+d} \varepsilon^{\frac{3}{e}}}{N^d \sigma_{t}^{d} (2 \pi)^{\frac{d}{2}}} \left(1+\frac{ (d+d_x)^{s}}{N^{\beta} s!}  \right)  \right]\\
& \lesssim \varepsilon + \frac{B}{N^{\beta}  \gamma_{t}^{\frac{d}{2}}} 
+ \frac{B \varepsilon^{\frac{3}{e}} (N^{\beta}+1)}{N^{d+\beta} \sigma_{t}^{d}}.
\end{aligned}  
\end{equation}

We can employ steps analogous to the preceding ones to apply the conditional diffusion local polynomial approximation to $\nabla q(\mathbf{y}_{t}|f_{\phi}(\mathbf{x}))$, which denoted as
\begin{equation*}
\begin{aligned}
&\ \ \ \ h_{4}(\mathbf{y}_{t},\mathbf{y}_{0},f_{\phi}(\mathbf{x}))\\
&=\sum\limits_{\mathbf{v},\mathbf{w}} \sum\limits_{\|\mathbf{n}\|_{1}+\|\mathbf{m}\|_{1} \leq s} 
\left.\frac{1}{\mathbf{n}! \mathbf{m}!} \frac{\partial^{\mathbf{n}+\mathbf{m}} \tau\left(\frac{\mathbf{y}_{0}}{R_{*}}+\frac{1}{2},\frac{\mathbf{f}}{R_{*}}+\frac{1}{2} \right)}{\partial \mathbf{y}_{0}^{\mathbf{n}}  \partial \mathbf{f}^{\mathbf{m}}} \right|_{\mathbf{y}_{0}=\frac{\mathbf{v}}{N},\mathbf{f}=\frac{\mathbf{w}}{N}}  
\left(\frac{\mathbf{f}}{R_{*}}+\frac{1}{2}-\frac{\mathbf{w}}{N} \right)
\prod_{j=1}^{d_x} \psi \left[ 3N \left( \frac{\mathbf{f}_{j}}{R_{*}}+\frac{1}{2}-\frac{\mathbf{w}_{j}}{N}\right)\right]\\
&\ \ \  \cdot \prod\limits_{i=1}^d \frac{1}{\sigma_{t} \sqrt{2 \pi}} \int_{D_{0i}} 
\left( \frac{\mathbf{y}_{0}}{R_{*}}+\frac{1}{2}-\frac{\mathbf{v}_{i}}{N} \right)^{n_{i}}
\sum\limits_{k=0}^{p-1} \frac{1}{k!} \left( -\frac{\left| \mathbf{y}_{ti}-\left[ \gamma_{t} \mathbf{y}_{0i} + \left(1-\gamma_{t} \right) f_{\phi}(\mathbf{x}) \right] \right|^{2}}{2 \sigma_{t}^{2}} \right)^{k} 
\frac{ \mathbf{y}_{ti}-\left[ \gamma_{t} \mathbf{y}_{0i} + \left(1-\gamma_{t} \right) f_{\phi}(\mathbf{x}) \right]}{ \sigma_{t}}
d \mathbf{y}_{0i},
\end{aligned}
\end{equation*}
and we finally obtain the error as follows
\begin{equation}\label{BB25}
\begin{aligned}
\left|\sigma_{t} \nabla q(\mathbf{y}_{t}|f_{\phi}(\mathbf{x}))-h_{4}(\mathbf{y}_{t},\mathbf{y}_{0},f_{\phi}(\mathbf{x}))\right|
&\leq 2\left[ \varepsilon 
+ \frac{c BR^{s}_{*} (d+d_x)^{s} \left(\log \varepsilon^{-1}\right)^{\frac{1}{2}}}{N^{\beta} s! \gamma_{t}^{\frac{d}{2}}} 
+ \frac{c B R_{*}^{s+d} \varepsilon^{\frac{3}{e}} \left(\log \varepsilon^{-1}\right)^{\frac{1}{2}} }{N^d \sigma_{t}^{d} (2 \pi)^{\frac{d}{2}}} \left(1+\frac{ (d+d_x)^{s}}{N^{\beta} s!}  \right)  \right]\\
& \lesssim \varepsilon 
+ \frac{B \left(\log \varepsilon^{-1}\right)^{\frac{1}{2}}}{N^{\beta} \gamma_{t}^{\frac{d}{2}}} 
+ \frac{ B \varepsilon^{\frac{3}{e}} \left(\log \varepsilon^{-1}\right)^{\frac{1}{2}} (N^{\beta}+1)}{N^{d+\beta} \sigma_{t}^{d}}.
\end{aligned}  
\end{equation}

Thereby, we can denote the conditional diffusion local polynomial of the network-estimated value S of the scoring function $s_{\theta}(\mathbf{y}_{t},f_{\phi}(\mathbf{x}),t)$ as 
\begin{equation}\label{BB26}
s_{\theta}(\mathbf{y}_{t},f_{\phi}(\mathbf{x}),t)= \min{\left( \frac{h_{4}}{\sigma_{t} h_{1,clip}}, 
\frac{c_4}{\sigma_{t}^{3}} \left( \left\|\mathbf{y}_{t} \right\|_{\infty}^{2} + 2(1-\gamma_{t})^{2} \left\| f_{\phi}(\mathbf{x})\right\|_{\infty}^{2}  \right)^{\frac{1}{2}} + c_5\right)},
\end{equation}
Here, $h_{1,clip} = \max{(h_1, \varepsilon_{low})}$ serves as a truncation of the original $h_{1}$ for the purpose of preventing instability in regions where values are relatively minute. 
Furthermore, $q(\mathbf{y}_{t}|f_{\phi}(\mathbf{x})) \geq \varepsilon_{low}$ is specified when $\varepsilon_{low} \geq \varepsilon + B/N^{\beta}  \gamma_{t}^{\frac{d}{2}}  + B \varepsilon^{\frac{3}{e}} (N^{\beta}+1)/N^{d+\beta} \sigma_{t}^{d}$.                                            

Finally, we have
\begin{equation*}
\begin{aligned}
&\ \ \ \left\| s_{\theta}(\mathbf{y}_{t},f_{\phi}(\mathbf{x}),t)-\nabla \log q(\mathbf{y}_{t}|f_{\phi}(\mathbf{x})) \right\|_{\infty}\\
&\leq \left\| \frac{h_{4}}{\sigma_{t} h_{1,clip}} - \nabla \log q(\mathbf{y}_{t}|f_{\phi}(\mathbf{x}))\right\|_{\infty}\\
% &\leq  \left| \frac{\sigma_{t} \nabla q(\mathbf{y}_{t}|f_{\phi}(\mathbf{x}))-h_{4}(\mathbf{y}_{t},\mathbf{y}_{0},f_{\phi}(\mathbf{x}))}{\sigma_{t} h_{1,clip}} \right|
% + \left\| \nabla q(\mathbf{y}_{t}|f_{\phi}(\mathbf{x})) \right\|_{\infty}   \left| \frac{1}{h_{1,clip}} - \frac{1}{q(\mathbf{y}_{t}|f_{\phi}(\mathbf{x}))} \right|\\
& \leq \left| \frac{\sigma_{t} \nabla q(\mathbf{y}_{t}|f_{\phi}(\mathbf{x}))-h_{4}(\mathbf{y}_{t},\mathbf{y}_{0},f_{\phi}(\mathbf{x}))}{\sigma_{t} h_{1,clip}} \right|
+ \left\| \nabla \log q(\mathbf{y}_{t}|f_{\phi}(\mathbf{x})) \right\|_{\infty} q(\mathbf{y}_{t}|f_{\phi}(\mathbf{x}))
\left| \frac{1}{h_{1,clip}} - \frac{1}{q(\mathbf{y}_{t}|f_{\phi}(\mathbf{x}))} \right|\\
% & \leq \left| \frac{\sigma_{t} \nabla q(\mathbf{y}_{t}|f_{\phi}(\mathbf{x}))-h_{4}(\mathbf{y}_{t},\mathbf{y}_{0},f_{\phi}(\mathbf{x}))}{\sigma_{t} h_{1,clip}} \right|
% + \frac{\left\| \nabla \log q(\mathbf{y}_{t}|f_{\phi}(\mathbf{x})) \right\|_{\infty}}{h_{1,clip}}  \left| q(\mathbf{y}_{t}|f_{\phi}(\mathbf{x}))-h_{1,clip}\right|  \\
% & = \frac{1}{h_{1,clip}} 
% \left( \left| \frac{\sigma_{t} \nabla q(\mathbf{y}_{t}|f_{\phi}(\mathbf{x}))-h_{4}(\mathbf{y}_{t},\mathbf{y}_{0},f_{\phi}(\mathbf{x}))}{\sigma_{t}} \right| + \left\| \nabla \log q(\mathbf{y}_{t}|f_{\phi}(\mathbf{x})) \right\|_{\infty}  \left| q(\mathbf{y}_{t}|f_{\phi}(\mathbf{x}))-h_{1,clip}\right| \right)\\
% &\leq \frac{2}{q(\mathbf{y}_{t}|f_{\phi}(\mathbf{x}))} 
% \left( \left| \frac{\sigma_{t} \nabla q(\mathbf{y}_{t}|f_{\phi}(\mathbf{x}))-h_{4}(\mathbf{y}_{t},\mathbf{y}_{0},f_{\phi}(\mathbf{x}))}{\sigma_{t}} \right| + \left\| \nabla \log q(\mathbf{y}_{t}|f_{\phi}(\mathbf{x})) \right\|_{\infty}  \left| q(\mathbf{y}_{t}|f_{\phi}(\mathbf{x}))-h_{1,clip}\right| \right)\\
& \leq \frac{4}{q(\mathbf{y}_{t}|f_{\phi}(\mathbf{x}))}
\left[   \frac{1}{\sigma_{t}}\left( \varepsilon 
+ \frac{c BR^{s}_{*} (d+d_x)^{s} \left(\log \varepsilon^{-1}\right)^{\frac{1}{2}}}{N^{\beta} s! \gamma_{t}^{\frac{d}{2}}} 
+ \frac{c B R_{*}^{s+d} \varepsilon^{\frac{3}{e}} \left(\log \varepsilon^{-1}\right)^{\frac{1}{2}} }{N^d \sigma_{t}^{d} (2 \pi)^{\frac{d}{2}}} \left(1+\frac{ (d+d_x)^{s}}{N^{\beta} s!}  \right)  \right)\right.\\
&\ \ \ +\left.\left(\frac{c_4}{\sigma_{t}^{3}} \left( \left\|\mathbf{y}_{t} \right\|_{\infty}^{2} + 2(1-\gamma_{t})^{2} \left\| f_{\phi}(\mathbf{x})\right\|_{\infty}^{2}  \right)^{\frac{1}{2}} + c_5 \right)
\left( \varepsilon + \frac{BR^{s}_{*} (d+d_x)^{s}}{N^{\beta} s! \gamma_{t}^{\frac{d}{2}}} + \frac{d B R_{*}^{s+d} \varepsilon^{\frac{3}{e}}}{N^d \sigma_{t}^{d} (2 \pi)^{\frac{d}{2}}} \left(1+\frac{ (d+d_x)^{s}}{N^{\beta} s!} \right) \right) \right]\\
& \leq \frac{4}{q(\mathbf{y}_{t}|f_{\phi}(\mathbf{x}))}
\left[   \frac{1}{\sigma_{t}}\left( \varepsilon 
+ \frac{c BR^{s}_{*} (d+d_x)^{s} \left(\log \varepsilon^{-1}\right)^{\frac{1}{2}}}{N^{\beta} s! \gamma_{t}^{\frac{d}{2}}} 
+ \frac{c B R_{*}^{s+d} \varepsilon^{\frac{3}{e}} \left(\log \varepsilon^{-1}\right)^{\frac{1}{2}} }{N^d \sigma_{t}^{d} (2 \pi)^{\frac{d}{2}}} \left(1+\frac{ (d+d_x)^{s}}{N^{\beta} s!}  \right)  \right)\right.\\
&\ \ \ +\left.\left(\frac{c_4}{\sigma_{t}^{3}} \left( R^{2} + 2(1-\gamma_{t})^{2} R_{f}^{2}  \right)^{\frac{1}{2}} + c_5 \right)
\left( \varepsilon + \frac{BR^{s}_{*} (d+d_x)^{s}}{N^{\beta} s! \gamma_{t}^{\frac{d}{2}}} + \frac{d B R_{*}^{s+d} \varepsilon^{\frac{3}{e}}}{N^d \sigma_{t}^{d} (2 \pi)^{\frac{d}{2}}} \left(1+\frac{ (d+d_x)^{s}}{N^{\beta} s!} \right) \right) \right]\\
& \lesssim \frac{1}{q(\mathbf{y}_{t}|f_{\phi}(\mathbf{x}))}
\left[  \frac{1}{\sigma_{t}} \left(\varepsilon 
+ \frac{B \left(\log \varepsilon^{-1}\right)^{\frac{1}{2}}}{N^{\beta} \gamma_{t}^{\frac{d}{2}}} 
+ \frac{ B \varepsilon^{\frac{3}{e}} \left(\log \varepsilon^{-1}\right)^{\frac{1}{2}} (N^{\beta}+1)}{N^{d+\beta} \sigma_{t}^{d}} \right) \right.\\
&\ \ \ \left.+ \left(\frac{1}{\sigma_{t}^{3}} \left( R^{2} + 2(1-\gamma_{t})^{2} R_{f}^{2}  \right)^{\frac{1}{2}} + c_5 \right)   
\left( \varepsilon + \frac{B}{N^{\beta}  \gamma_{t}^{\frac{d}{2}}} + \frac{B \varepsilon^{\frac{3}{e}} (N^{\beta}+1)}{N^{d+\beta} \sigma_{t}^{d}}\right)
\right].
% \\
% &\lesssim (N + 1)(1+\gamma_{t}^{d} ) +\varepsilon + ( (\log\varepsilon^{-1})^{\frac{1}{2}} +1 ) N^{-\beta} [\gamma_{t}^{d/2} + (N^{\beta}+1) N^{-d} ].
\end{aligned}
\end{equation*}

According to the given condition that $\gamma_{t} = e^{-\frac{\beta}{2} t},\  \sigma^{2}_{t} = 1-e^{- \beta t}$, we know that as $t$ increases, especially in the limit case when $t \to \infty$, $\sigma_{t} \to 1$ and $\gamma_{t} \to 0$. In the final result, $\gamma_{t}$ plays a fine-tuning role in the overall magnitude as $t$ varies. This also shows that we may actually need to choose a step $t_0$ that satisfies the conditions for this convergent upper bound to hold. 
$K_{t}$ is a constant related to certain hyperparameters, and the error $\varepsilon \in (0, 1/e)$ such that $\varepsilon^{\frac{3}{e}}$ can be neglected compared to other magnitudes of $\varepsilon$. Eventually, we obtain a non-asymptotic upper bound as 
\begin{equation*}
\begin{aligned}
&\int_{\mathbb{R}^{d}} \left\| s_{\theta}(\mathbf{y}_{t},f_{\phi}(\mathbf{x}),t)-\nabla \log q(\mathbf{y}_{t}|f_{\phi}(\mathbf{x})) \right\| ^{2}_{2} \ q(\mathbf{y}_{t}|f_{\phi}(\mathbf{x})) \ d \mathbf{y}_{t}\\
&\lesssim 
% O\left( \varepsilon + \frac{N + 1}{\sigma_{t}^{d}}(1+\gamma_{t}^{-d} ) + N^{-\beta} \left[(\log\varepsilon^{-1})^{\frac{1}{2}} +1 \right] \left[\gamma_{t}^{d/2} + \frac{(N^{\beta}+1) N^{-d}}{\sigma_{t}^{d}} \right]  \right).
O\left( \left[(\log\varepsilon^{-1})^{\frac{1}{2}} +1 \right] \left[\frac{N^{-\beta}}{\gamma_{t}^{d/2} } + \frac{N^{-d}}{\sigma_{t}^{d}} + \frac{N^{-d-\beta}}{\sigma_{t}^{d}} \right]  \right). 
\end{aligned}
\end{equation*}
The proof is complete.

\section{Simulation results}\label{appendixD}
\begin{figure}[H]
  \centering
  \subfloat[Moon]
  {
      %\label{fig121:subfig1}
      \includegraphics[width=0.35\textwidth]{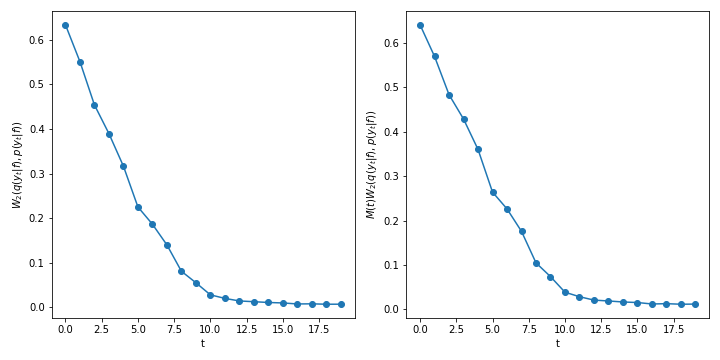}
  }
  \subfloat[Circle]
  {
      %\label{fig121:subfig2}
      \includegraphics[width=0.35\textwidth]{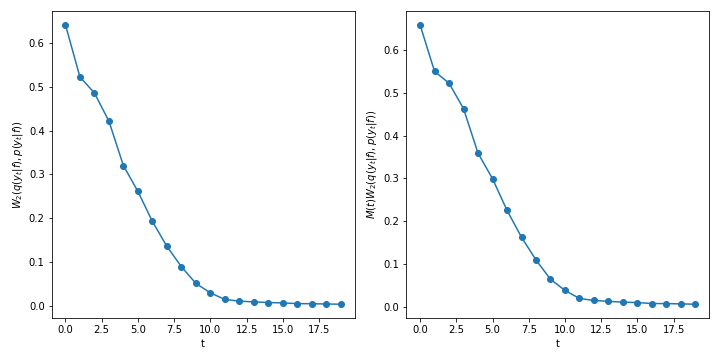}
  }
  \\
  \subfloat[$N(\mathbf{0},0.1\mathbf{I})$]
  {
      %\label{fig121:subfig3}
      \includegraphics[width=0.35\textwidth]{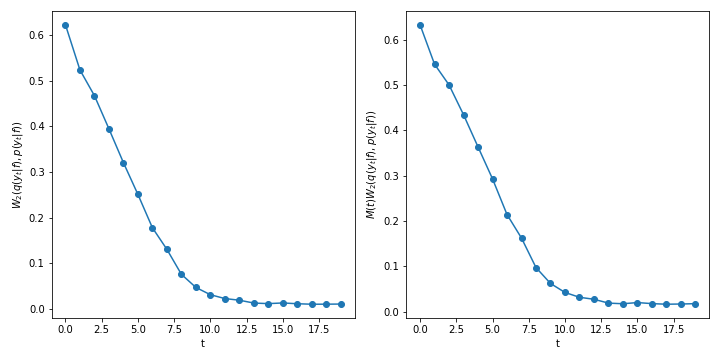}
  }
  \subfloat[$N ((\pm0.5, \pm0.5)^{T}, 0.01\mathbf{I})$]
  {
      %\label{fig121:subfig4}
      \includegraphics[width=0.35\textwidth]{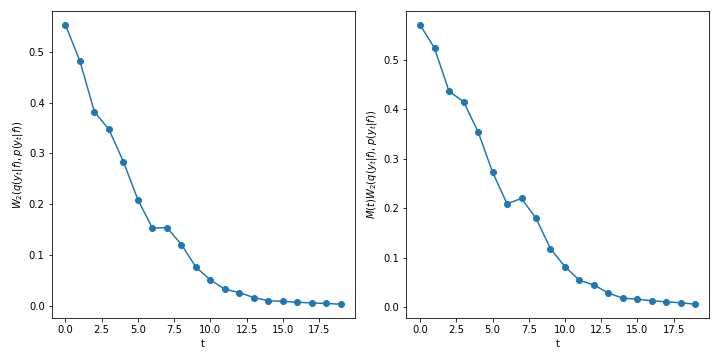}
  }
  \caption{\textbf{Trends in four datasets with respect to Wasserstein distance and its product term.} The horizontal axis shows the diffusion time step $t$, and the vertical axis illustrates the evolution of Wasserstein distance $W_{2}(q(\mathbf{y}_{t}|f_{\phi}(\mathbf{x})),p(\mathbf{y}_{t}|f_{\phi}(\mathbf{x})))$, as well as the product of Wasserstein distance with the integral factor $M(t)$, written as $ M(t)W_{2}(q(\mathbf{y}_{T}|f_{\phi}(\mathbf{x})),p(\mathbf{y}_{t}|f_{\phi}(\mathbf{x})))$.}
  \label{fig121}
\end{figure}

\begin{figure}[ht]
  \centering
  \subfloat[Moon]
  {
    %\label{fig122:subfig1}
    \includegraphics[width=0.21\textwidth]{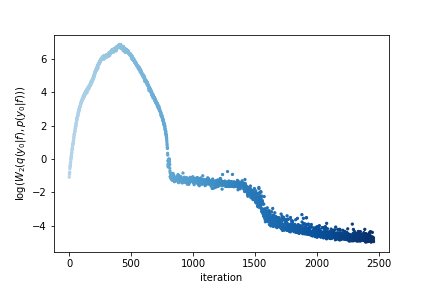}
  }
  \subfloat[Circle]
  {
      %\label{fig122:subfig2}
      \includegraphics[width=0.21\textwidth]{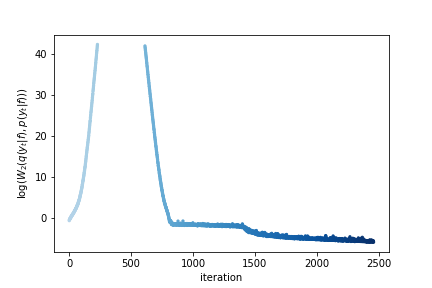}
  }
  \subfloat[$N(\mathbf{0},0.1\mathbf{I})$]
  {
      %\label{fig122:subfig3}
      \includegraphics[width=0.21\textwidth]{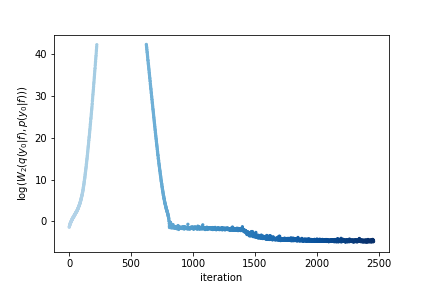}
  }
  \subfloat[$N ((\pm0.5, \pm0.5)^{T}, 0.01\mathbf{I})$]
  {
      %\label{fig122:subfig4}
      \includegraphics[width=0.21\textwidth]{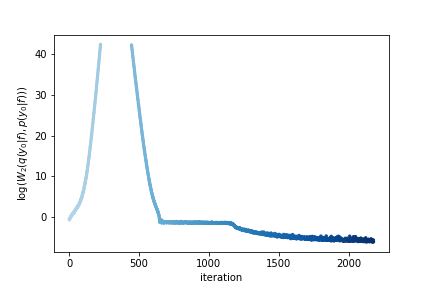}
  }
  \caption{\textbf{Log Wasserstein Distances plots for Four Datasets.} The subfigures from left to right represent Moon, Circle, zero-mean Gaussian, four-class Gaussian datasets in turn. The figures above illustrate the evolution of the logarithmic Wasserstein distance $\log W_{2}(q(\mathbf{y}_{t}|f_{\phi}(\mathbf{x})),p(\mathbf{y}_{t}|f_{\phi}(\mathbf{x})))$ as the training iterations progress for various four datasets.}
  \label{fig122}
\end{figure}

\begin{figure}[H]
  \centering
  \subfloat[Moon]
  {
      %\label{fig122:subfig1}
      \includegraphics[width=0.21\textwidth]{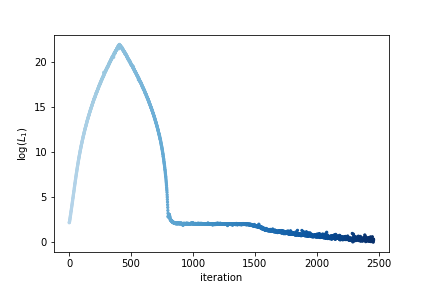}
  }
  \subfloat[Circle]
  {
      %\label{fig122:subfig2}
      \includegraphics[width=0.21\textwidth]{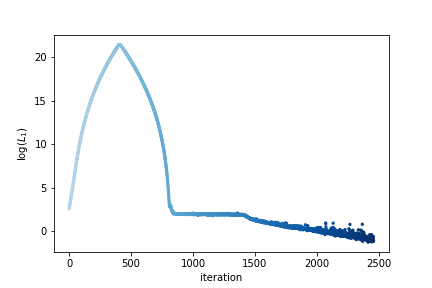}
  }
  \subfloat[$N(\mathbf{0},0.1\mathbf{I})$]
  {
      %\label{fig122:subfig3}
      \includegraphics[width=0.21\textwidth]{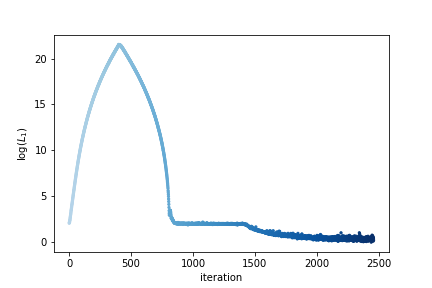}
  }
  \subfloat[$N ((\pm0.5, \pm0.5)^{T}, 0.01\mathbf{I})$]
  {
      %\label{fig122:subfig4}
      \includegraphics[width=0.21\textwidth]{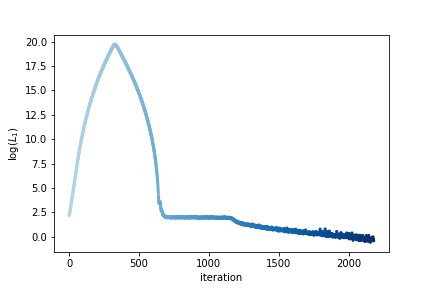}
  }
  \caption{\textbf{Log loss function plots for Four Datasets.} The subfigures from left to right represent Moon, Circle, zero-mean Gaussian, four-class Gaussian datasets in turn. The figures above illustrate the evolution of the logarithmic Wasserstein distance $\log L_{1}(\phi,\theta,\lambda)$ as the training iterations progress for various four datasets.}
  \label{fig123}
\end{figure}

\end{document}